\ifx\XeTeXversion\undefined\pdfoutput=1\fi 
\documentclass{article}
\usepackage[T1]{fontenc}
\usepackage{iclr2027_conference,times}

\usepackage{amsmath,amsfonts,bm}

\def\eqref#1{equation~\ref{#1}}

\def\1{\bm{1}}

\DeclareMathAlphabet{\mathsfit}{\encodingdefault}{\sfdefault}{m}{sl}
\SetMathAlphabet{\mathsfit}{bold}{\encodingdefault}{\sfdefault}{bx}{n}

\usepackage{hyperref}
\usepackage{url}
\usepackage{graphicx}
\usepackage{booktabs}
\IfFileExists{multirow.sty}{\usepackage{multirow}}{\newcommand{\multirow}[3]{\smash{\raisebox{\dimexpr-(\number#1-1)\ht\strutbox/2\relax}{#3}}}}
\usepackage{amsmath}
\usepackage{amssymb}
\usepackage{xcolor}
\usepackage{longtable}
\usepackage{array}
\IfFileExists{wrapfig.sty}{\usepackage{wrapfig}}{}
\IfFileExists{placeins.sty}{\usepackage{placeins}}{\newcommand{\FloatBarrier}{\clearpage}} 
\IfFileExists{adjustbox.sty}{\usepackage{adjustbox}}{%
  \newsavebox{\adjboxtmp}%
  \newenvironment{adjustbox}[1]{\begin{lrbox}{\adjboxtmp}}{\end{lrbox}\ifdim\wd\adjboxtmp>\linewidth\resizebox{\linewidth}{!}{\usebox{\adjboxtmp}}\else\usebox{\adjboxtmp}\fi}%
} 
\newcommand{\stk}[1]{\begin{tabular}[c]{@{}c@{}}#1\end{tabular}} 

\newcommand{\yFwdDriftOnedispatch}{-1.7}

\newcommand{\dimDdispatch}{56}
\newcommand{\ctxExcessdispatch}{7.8}

\newcommand{\yFwdDriftOnelocation}{-2.3}

\newcommand{\dimDlocation}{66}
\newcommand{\ctxExcesslocation}{7.6}

\newcommand{\stepGainOneOneDecorgan}{3.2}

\newcommand{\yFwdDriftOneorgan}{-3.0}

\newcommand{\dimDorgan}{78}
\newcommand{\ctxExcessorgan}{6.8}

\newcommand{\stepGainOneOneDecscene}{20.4}

\newcommand{\yFwdDriftOnescene}{-7.1}

\newcommand{\dimDscene}{1,321}

\newcommand{\gateOnevitals}{+0.40}

\newcommand{\yFwdDriftOnevitals}{-6.9}
\newcommand{\yRevDriftOnevitals}{-5.5}
\newcommand{\yRevInYearOnevitals}{51.53}
\newcommand{\yRevExtOnevitals}{46.01}

\newcommand{\dimDvitals}{1,398}

\newcommand{\gateOneimpressions}{+0.39}

\newcommand{\stepGainOneOneDecimpressions}{5.6}

\newcommand{\yFwdDriftOneimpressions}{-5.7}
\newcommand{\yRevDriftOneimpressions}{-8.2}
\newcommand{\yRevInYearOneimpressions}{59.36}
\newcommand{\yRevExtOneimpressions}{51.16}

\newcommand{\dimDimpressions}{2,073}

\newcommand{\dpcsLowTopOneSix}{42.82}
\newcommand{\dpcsLowTopThreeSix}{69.69}

\newcommand{\dpcsMedTopOneSix}{62.21}
\newcommand{\dpcsMedTopThreeSix}{82.05}

\newcommand{\dpcsHighTopOneSix}{80.61}
\newcommand{\dpcsHighTopThreeSix}{90.19}

\newcommand{\dpcsLowShareBefore}{94}
\newcommand{\dpcsHighMaxBefore}{108}
\newcommand{\dpcsMinBucket}{298}
\newcommand{\dpcsAurocThreeSix}{0.637}
\newcommand{\dpcsAurocMassThreeSix}{0.805}
\newcommand{\dpcsSupportMedian}{83.5}
\newcommand{\dpcsSupportLtThirty}{2.2}

\newcommand{\natExtMajorityPctInt}{54}

\newcommand{\natNatMajorityPctInt}{53}

\newcommand{\extEnsMacroOneimpressions}{58.18}
\newcommand{\extStepMacroOneimpressions}{60.05}

\newcommand{\stepTopOnedispatch}{20.72}
\newcommand{\stepTopThreedispatch}{38.34}

\newcommand{\dOnedispatch}{+0.30}

\newcommand{\dOnelocation}{+0.09}

\newcommand{\ensTopOnescene}{48.16}

\newcommand{\dOnescene}{+0.09}
\newcommand{\dThreescene}{+0.91}

\newcommand{\yFwdOnescene}{+0.98}
\newcommand{\yFwdThreescene}{+1.01}
\newcommand{\yRevOnescene}{+0.32}
\newcommand{\yRevThreescene}{+0.98}

\newcommand{\stepTopThreevitals}{75.56}
\newcommand{\ensTopOnevitals}{52.94}
\newcommand{\ensTopThreevitals}{74.53}
\newcommand{\dOnevitals}{+0.77}
\newcommand{\dThreevitals}{+1.02}

\newcommand{\staticOnevitals}{+0.37}
\newcommand{\staticThreevitals}{+0.82}

\newcommand{\yFwdOnevitals}{+1.11}
\newcommand{\yFwdThreevitals}{+1.24}
\newcommand{\yRevOnevitals}{+1.00}
\newcommand{\yRevThreevitals}{+1.26}
\newcommand{\stepTopOneimpressions}{59.31}
\newcommand{\stepTopThreeimpressions}{79.82}
\newcommand{\ensTopOneimpressions}{58.34}
\newcommand{\ensTopThreeimpressions}{78.56}
\newcommand{\dOneimpressions}{+0.96}
\newcommand{\dThreeimpressions}{+1.27}

\newcommand{\staticOneimpressions}{+0.57}
\newcommand{\staticThreeimpressions}{+1.18}

\newcommand{\yFwdOneimpressions}{+1.15}
\newcommand{\yFwdThreeimpressions}{+1.36}
\newcommand{\yRevOneimpressions}{+1.00}
\newcommand{\yRevThreeimpressions}{+1.24}

\newcommand{\nFamilyASurvive}{4}
\newcommand{\stageSixEnsNll}{1.696}
\newcommand{\stageSixStepNll}{1.498}
\newcommand{\stageSixVectorNll}{1.592}
\newcommand{\extEnsMicroOne}{24.42}
\newcommand{\extStepMicroOne}{23.24}
\newcommand{\extEnsMicroThree}{47.52}
\newcommand{\extStepMicroThree}{41.92}
\newcommand{\extEnsMacroOne}{58.18}
\newcommand{\extStepMacroOne}{60.05}
\newcommand{\extEnsKnownPrior}{60.81}
\newcommand{\extStepKnownPrior}{63.45}
\newcommand{\extEnsMacroF}{20.04}

\newcommand{\extNoBiasMicroOne}{25.22}
\newcommand{\extNoBiasMacroF}{20.38}
\newcommand{\extNoBiasBalOne}{+0.73}
\newcommand{\extNoBiasBalThree}{+1.43}

\newcommand{\dpcsAurocMaxPostSix}{0.809}
\newcommand{\dpcsAurocSix}{0.650}

\newcommand{\ctxGainScene}{+5.14}

\newcommand{\qwenGapOnedispatch}{5.74}

\newcommand{\qwenGapOnevitals}{26.59}

\newcommand{\qwenSdOneimpressions}{0.75}

\newcommand{\qwenGapOneimpressions}{26.01}

\newcommand{\nFamilyC}{48}

\newcommand{\medgInvalidMax}{0.21}
\newcommand{\llmCallsQwen}{196,752}
\newcommand{\llmCallsMedg}{98,376}
\newcommand{\llmCallsTotal}{295,128}
\newcommand{\llmPromptTokensMillions}{531}

\newcommand{\ensMinusNnOneorgan}{+0.24}

\newcommand{\ensTopFivevitals}{82.06}
\newcommand{\stepTopFivevitals}{82.98}
\newcommand{\dFivevitals}{+0.93}

\newcommand{\ensMinusNnOnevitals}{+2.60}

\newcommand{\ensTopFiveimpressions}{85.09}
\newcommand{\stepTopFiveimpressions}{86.36}
\newcommand{\dFiveimpressions}{+1.27}

\newcommand{\ensMinusNnOneimpressions}{+2.32}

\newcommand{\yTwentyThreeMinCount}{411}
\newcommand{\yTwentyThreeBelowThousand}{7}

\newcommand{\distExtTauStepThreeimpressions}{89.45}

\newcommand{\distExtAsDThreeimpressions}{-5.60}

\newcommand{\distExtTauDThreeimpressions}{+1.32}

\newcommand{\distNatTauStepThreeimpressions}{87.18}

\newcommand{\distNatAsDThreeimpressions}{-5.34}

\newcommand{\distNatTauDThreeimpressions}{+1.16}

\newcommand{\gbmMinusNnOneOneDecscene}{+1.4}

\newcommand{\stepMinusGbmOneOneDecscene}{+1.0}

\newcommand{\gbmMinusNnOneOneDecvitals}{+2.2}

\newcommand{\stepMinusGbmOneOneDecimpressions}{+1.2}

\newcommand{\gbmPmaxOne}{0.007}
\newcommand{\gbmPoolOnevitals}{+0.90}
\newcommand{\gbmPoolThreevitals}{+0.74}

\newcommand{\gbmPoolOneimpressions}{+0.87}
\newcommand{\gbmPoolThreeimpressions}{+0.68}

\newcommand{\gbmPoolPmax}{0.0031}
\newcommand{\archTrailMin}{0.1}
\newcommand{\archTrailMax}{1.7}

\newcommand{\harmMin}{2.7}
\newcommand{\harmMax}{5.8}
\newcommand{\harmInTopThreeMin}{71}
\newcommand{\harmInTopThreeMax}{82}

\newcommand{\hardRuleNetMax}{+0.5}
\newcommand{\hardRuleNetMin}{-2.9}
\newcommand{\conjJointMin}{29}
\newcommand{\conjJointMax}{33}
\newcommand{\conjDecisionMin}{7}
\newcommand{\conjDecisionMax}{12}

\newcommand{\synRegTwoKNoise}{+11.4}
\newcommand{\synPlainTwoKNoise}{+14.4}

\newcommand{\synRegEightKNoise}{+8.2}
\newcommand{\synPlainEightKNoise}{+8.4}

\newcommand{\synRegThirtyTwoKNoise}{+3.5}
\newcommand{\synPlainThirtyTwoKNoise}{+3.6}

\newcommand{\covUnmodMaxAbs}{0.7}
\newcommand{\covUnmodMinP}{0.2}

\newcommand{\covPadOneFiveK}{+3.9}
\newcommand{\covPadThreeFiveK}{+1.1}
\newcommand{\covPadPOneFiveK}{3.3\!\times\!10^{-6}}

\newcommand{\covHarmEnsFiveK}{10.9}
\newcommand{\covHarmGatedFiveK}{4.7}
\newcommand{\covPadOneTwentyK}{+3.9}
\newcommand{\covPadThreeTwentyK}{+0.8}
\newcommand{\covPadPOneTwentyK}{2.0\!\times\!10^{-22}}
\newcommand{\covPadStaticOneTwentyK}{+4.1}

\newcommand{\provMergedDOne}{+0.96}
\newcommand{\provMergedDThree}{+1.27}

\newcommand{\provSepDOne}{+0.74}
\newcommand{\provSepDThree}{+1.34}
\newcommand{\provSepPOne}{0.010}
\newcommand{\provSepPThree}{1.4\times10^{-7}}
\newcommand{\provSepMinusMergedEnsOne}{-0.06}
\newcommand{\provSepMinusMergedEnsThree}{-0.07}
\newcommand{\provEnsPOne}{0.88}
\newcommand{\provEnsPThree}{0.84}
\newcommand{\provSepCols}{2{,}644}
\newcommand{\expOnlyStaticscene}{44.45}
\newcommand{\expOnlyGatedscene}{44.85}

\newcommand{\expOnlyStaticvitals}{48.01}
\newcommand{\expOnlyGatedvitals}{48.83}

\newcommand{\expOnlyStaticimpressions}{54.16}
\newcommand{\expOnlyGatedimpressions}{54.94}

\newcommand{\expOnlyDeficitMin}{3.3}
\newcommand{\expOnlyDeficitMax}{4.9}

\newcommand{\selEpochMin}{9}
\newcommand{\selEpochMax}{79}
\newcommand{\seedSpreadMin}{0.10}
\newcommand{\seedSpreadMax}{0.78}
\newcommand{\selKthreeDOne}{+2.20}
\newcommand{\selKthreeDThree}{+1.80}
\newcommand{\selKthreePOne}{2.1\times10^{-24}}
\newcommand{\selKthreePThree}{3.7\times10^{-10}}
\newcommand{\selKfiveDOne}{+1.76}
\newcommand{\selKfiveDThree}{-1.21}
\newcommand{\selKfivePOne}{2.0\times10^{-14}}
\newcommand{\selKfivePThree}{9.7\times10^{-5}}
\newcommand{\selKn}{14{,}760}

\newcommand{\costLlmSeqS}{1.29}
\newcommand{\costLlmSeqMedS}{1.32}
\newcommand{\costLlmSeqPninetyS}{1.35}
\newcommand{\costLlmPromptTok}{2{,}551}
\newcommand{\costLlmComplTok}{38}
\newcommand{\costLlmBatchS}{15.4}
\newcommand{\costLlmAmortMs}{77}

\newcommand{\costLlmRunMs}{237}

\newcommand{\eoneEqBiasOneSix}{55.10}

\newcommand{\eoneGSmeanOneSix}{+0.32}
\newcommand{\eoneGSminOneSix}{+0.26}
\newcommand{\eoneGSmaxOneSix}{+0.39}

\newcommand{\eoneStaticMinusEqBiasOneSix}{+3.82}

\newcommand{\eoneGSminp}{0.033}

\newcommand{\eoneEqWorseFrom}{2}

\newcommand{\etwoNatHalf}{45{,}379}
\newcommand{\etwoNatPrevOne}{53.82}
\newcommand{\etwoNatPrevThree}{68.35}

\newcommand{\etwoNatStepDisjOneFour}{48.25}

\newcommand{\etwoNatStepStaticOneFive}{-3.33}

\newcommand{\etwoNatStepEnsOneSix}{+5.01}
\newcommand{\etwoNatStepEnsOneEeLo}{+4.64}
\newcommand{\etwoNatStepEnsOneEeHi}{+5.30}
\newcommand{\etwoNatStepEnsThreeSix}{+2.90}

\newcommand{\etwoNatStepStaticOneSix}{-0.71}

\newcommand{\etwoNatStepEnsTsOneSix}{-1.33}
\newcommand{\etwoNatStepEnsTsThreeSix}{+4.78}

\newcommand{\etwoNatStepEnsBctsOneSix}{+0.48}
\newcommand{\etwoNatStepEnsBctsThreeSix}{+0.82}

\newcommand{\etwoOldPrevOne}{52.95}
\newcommand{\etwoOldPrevThree}{66.83}

\newcommand{\etwoOldStepDisjOneFour}{41.25}

\newcommand{\etwoOldStepDisjOneFive}{49.08}

\newcommand{\etwoOldEnsDisjOneSix}{52.33}

\newcommand{\etwoOldStepEnsOneSix}{+1.82}

\newcommand{\etwoOldStepEnsThreeSix}{-0.71}

\newcommand{\etwoOldStepEnsTsOneSix}{-2.27}
\newcommand{\etwoOldStepEnsTsThreeSix}{+4.69}

\newcommand{\etwoOldStepEnsBctsOneSix}{+2.47}
\newcommand{\etwoOldStepEnsBctsThreeSix}{+2.22}

\newcommand{\etwoEnsTempMin}{1.30}
\newcommand{\etwoEnsTempMax}{1.60}
\newcommand{\etwoPoolTempMin}{0.96}
\newcommand{\etwoPoolTempMax}{1.00}
\newcommand{\llmStepFoneHigher}{81}
\newcommand{\llmStepFoneTies}{1}

\newcommand{\famStepWithin}{50}
\newcommand{\famStepPredAirway}{230}

\newcommand{\famQwenWithin}{48}
\newcommand{\famQwenPredAirway}{107}
\newcommand{\famQwenTrueAirway}{300}

\newcommand{\famGemmaWithin}{44}
\newcommand{\famGemmaPredAirway}{126}

\newcommand{\onNllSix}{-0.035}
\newcommand{\onNllSixLo}{-0.045}
\newcommand{\onNllSixHi}{-0.026}
\newcommand{\onNllBonfStages}{2--6}
\newcommand{\onNllRunsNeg}{24}
\newcommand{\onNllRunsTot}{24}
\newcommand{\onEfiveMaxAbsDll}{0.0005}
\newcommand{\onEfiveMinHolm}{0.97}
\newcommand{\onEfiveNcis}{12}
\newcommand{\onEfiveNtests}{12}
\newcommand{\onEfiveMaxDauc}{0.001}

\newcommand{\onLatEns}{0.44}

\newcommand{\onLatStepCold}{4.31}
\newcommand{\onLatStepColdP}{4.37}

\newcommand{\onLatStepIncr}{0.96}

\newcommand{\onLatStepPool}{0.10}

\newcommand{\onSessEns}{1.9}

\newcommand{\onSessStep}{5.4}
\newcommand{\onPrepMs}{0.03}

\newcommand{\onFitsPooled}{252}
\newcommand{\onGpuHoursPooled}{4.8}

\newcommand{\onGpuMinMasked}{7.9}
\newcommand{\onParamsPoolM}{7.6}

\newcommand{\onMlpsPool}{39}
\newcommand{\onSpeedupSeq}{299}
\newcommand{\onSpeedupAmort}{18}
\newcommand{\onStepOverEnsCold}{10}
\newcommand{\onEfourLandmarkPct}{92.8}
\newcommand{\onEfourReadKeptPct}{24}
\newcommand{\onEfourRecKeptPct}{60.6}

\newcommand{\onEfourHrFull}{95.2}
\newcommand{\onEfourHrPre}{52.3}
\newcommand{\onEfourStepEnsOneFive}{+1.56}
\newcommand{\onEfourStepEnsThreeFive}{+1.77}

\newcommand{\onEfourStepEnsOneSix}{+1.41}
\newcommand{\onEfourStepEnsThreeSix}{+1.13}

\newcommand{\onEfourEnsDropOneSix}{-2.62}
\newcommand{\onEfourEnsDropThreeSix}{-1.94}

\newcommand{\onEfourPmax}{1.8\!\times\!10^{-5}}

\newcommand{\onEfourMainStepEnsOneFive}{+0.77}
\newcommand{\onEfourMainStepEnsThreeFive}{+1.02}
\newcommand{\onEfourMainStepEnsOneSix}{+0.96}
\newcommand{\onEfourMainStepEnsThreeSix}{+1.27}

\newcommand{\onEsixThreeIncMin}{+0.81}
\newcommand{\onEsixThreeIncMax}{+1.22}
\newcommand{\onEsixThreePmax}{7.5\!\times\!10^{-4}}
\newcommand{\onEsixEarlyMaxAbs}{0.48}
\newcommand{\onEsixSixMaxAbs}{0.06}
\newcommand{\onEsixSixPmin}{0.5}
\newcommand{\onEsixATwoOne}{+0.72}
\newcommand{\onEsixATwoThree}{+2.09}
\newcommand{\onEsixBFourOne}{-4.29}
\newcommand{\onEsixBFourThree}{-3.89}
\newcommand{\onEsixAMeanOne}{+0.12}
\newcommand{\onEsixAMeanThree}{+0.37}
\newcommand{\onEsixBMeanOne}{-0.77}
\newcommand{\onEsixBMeanThree}{-0.68}
\newcommand{\onEfourLmN}{7,607}
\newcommand{\onEfourLmNoN}{591}
\newcommand{\onEfourNoLmPct}{7.2}
\newcommand{\onEfourLmOneFive}{+1.59}
\newcommand{\onEfourLmThreeFive}{+1.87}
\newcommand{\onEfourLmOneSix}{+1.56}
\newcommand{\onEfourLmThreeSix}{+1.21}
\newcommand{\onEfourLmPmax}{1.2\!\times\!10^{-5}}
\newcommand{\onEfourNoLmEnsOneSix}{61.6}
\newcommand{\onEfourLmEnsOneSix}{55.3}
\newcommand{\onEfourNoLmFullEnsOneSix}{61.8}
\newcommand{\onEfourLmFullEnsOneSix}{58.1}
\newcommand{\onCOnePOneFive}{0.065}
\newcommand{\onTsNatEnsThreeSix}{83.55}
\newcommand{\onRawNatEnsThreeSix}{85.46}
\newcommand{\onTsOldEnsThreeSix}{75.66}
\newcommand{\onRawOldEnsThreeSix}{80.93}
\newcommand{\onTsNatStepThreeSix}{88.33}
\newcommand{\onRawNatStepThreeSix}{88.37}
\newcommand{\onTsOldStepOneFive}{49.22}
\newcommand{\onTsOldStaticOneSix}{52.77}

\newcommand{\parStepEnsTwelveOneSix}{+0.01}
\newcommand{\parStepEnsTwelveThreeSix}{+0.06}
\newcommand{\parEnsTwelveNllSix}{1.620}
\newcommand{\parStepNllSix}{1.498}
\newcommand{\parStepFourThreeMin}{+0.54}
\newcommand{\parStepFourThreeMax}{+0.62}
\newcommand{\parStepFourThreePmin}{0.015}
\newcommand{\parStepFourThreePmax}{0.026}
\newcommand{\parStepFourOnePmin}{0.21}
\newcommand{\parStepEnsTwelvePmin}{0.13}
\newcommand{\parStepEnsTwelveOneLoSix}{-0.51}
\newcommand{\parStepEnsTwelveOneHiSix}{+0.57}
\newcommand{\parStepEnsTwelveThreeLoSix}{-0.41}
\newcommand{\parStepEnsTwelveThreeHiSix}{+0.55}
\newcommand{\parStaticNllSix}{1.534}
\newcommand{\parStaticNllFive}{1.760}
\newcommand{\parStaticEnsTwelveOneFive}{-0.84}
\newcommand{\parStaticEnsTwelvePOneFive}{0.0022}

\newcommand{\tsEnsTwelveAucFour}{0.769}
\newcommand{\tsStepAucFour}{0.789}

\newcommand{\tsEnsTwelveNllFive}{1.747}

\newcommand{\tsStepEnsTwelveNllSix}{-0.036}
\newcommand{\tsStepEnsTwelveNllLoSix}{-0.047}
\newcommand{\tsStepEnsTwelveNllHiSix}{-0.025}
\newcommand{\tsEnsTwelveNllSix}{1.534}

\newcommand{\tsEnsTwelveAucSix}{0.809}
\newcommand{\tsStepAucSix}{0.809}
\newcommand{\tsNllAbsMin}{0.027}
\newcommand{\tsNllAbsMax}{0.073}

\title{Adaptive Sequence Modeling for Progressive Emergency Medical Service Protocol Prediction Under Streaming Information}

\iclrfinalcopy
\author{%
  \begin{tabular}[b]{@{}l@{\qquad}l@{}}
    \textbf{Pasan D. Gunawardena}\textsuperscript{1} & \textbf{Liuyi Jin}\textsuperscript{1} \\
    \texttt{pgunawardena@tamu.edu} & \texttt{liuyi@tamu.edu} \\[0.6ex]
    \textbf{Radu Stoleru}\textsuperscript{1} & \textbf{Michael Middleton}\textsuperscript{2} \\
    \texttt{stoleru@cse.tamu.edu} & \texttt{mmiddleton@tamu.edu}
  \end{tabular}\\[0.6ex]
  \textsuperscript{1}Texas A\&M University, College Station, Texas, USA\\
  \textsuperscript{2}Texas A\&M University EMS, College Station, Texas, USA%
}
\hypersetup{
  pdftitle={Adaptive Sequence Modeling for Progressive Emergency Medical Service Protocol Prediction Under Streaming Information},
  pdfauthor={Pasan D. Gunawardena, Liuyi Jin, Radu Stoleru, Michael Middleton}
}

\newcommand{\ngates}{238}
\newcommand{\ngatesselected}{84}

\newcommand{\nstepfamily}{12}
\newcommand{\nllmfamily}{48}

\begin{document}
\maketitle
\pagestyle{plain} 

\begin{abstract}
Emergency Medical Technicians (EMTs) must select treatment protocols under severe time and cognitive constraints while patient information is still being collected. Existing Emergency Medical Service (EMS) prediction systems typically evaluate completed records, leaving early-stage protocol discrimination and progressive evidence accumulation underexplored. We present a framework for progressive EMS protocol prediction, matched-evidence evaluation, and controlled empirical analysis as clinical information accumulates from dispatch through provider impressions. Progressive Clinical Feature Learning (PCFL) formulates protocol ranking over six cumulative, workflow-informed stages with nested feature availability and matched encounter evaluation. Stage-wise Ensembles for Temporal Pooling (STEP) combines current-stage, earlier-stage, and modality-specific predictions through cross-fitted logarithmic pooling and confidence-conditioned Temporal Feature Aggregation (TFA). Staged-Evidence Prompting (SEP) enforces identical stage-specific observability for large language models (LLMs), enabling record-level comparisons with structured predictors. The Dispatch--Protocol Confidence Score (DPCS) defines stage-dependent reliability tiers from validation protocol recall and dispatch-conditioned accuracy to examine how contextual reliability summaries relate to model confidence. Across 82 protocols in the 2024 National Emergency Medical Services Information System (NEMSIS) dataset, protocol-ranking accuracy rises from 20.72\% top-1 (38.34\% top-3) at dispatch to 59.31\% top-1 (79.82\% top-3) after provider impressions, with structured predictors outperforming evaluated zero-shot and few-shot LLMs at every stage. Controlled ablations show that learned pooling improves late-stage accuracy over a three-seed ensemble across seed replications, cross-year transfer, and timing-restricted vital signs, while a twelve-seed joint ensemble attains comparable accuracy and adaptive gating provides a consistent log-likelihood refinement. DPCS tiers stratify observed accuracy as evidence accumulates, whereas posterior confidence provides stronger error detection, clarifying design principles for progressive clinical decision support.
\end{abstract}

\section{Introduction}
\label{sec:intro}

Emergency medical services (EMS) clinicians must act before the full clinical picture is available, resulting in consequential decisions under severe time pressure and incomplete information. An incorrect diagnosis can direct care toward potentially harmful treatment (e.g. prehospital furosemide administration in patients whose subsequent diagnoses made its use inappropriate \citep{jaronik2006furosemide}). However, waiting for diagnostic certainty is often not an option. Crews must decide what to prepare, assess, and communicate using the evidence available at that moment, then revise those decisions as the encounter unfolds.

This process is inherently progressive. Field assessment is structured and selective, with successive assessments guiding further evaluation and care \citep{hagiwara2014cdss,sasser2012field}. Information arrives through dispatch, patient contact, scene observations, vital signs, and clinical examination, creating opportunities to update the working assessment and the protocols under consideration. Consider a dispatch report of an unresponsive person: subsequent information may raise suspicion of opioid overdose, prompting the crew to ready airway support and naloxone where clinically and operationally appropriate. Further findings may reinforce that assessment or redirect attention to another condition. Each revision can change which equipment and medications are prepared, which measurements are prioritized, which resources are requested, and what is communicated to the receiving hospital. Protocol assessment therefore helps shape an evolving workflow from before patient contact through transport.

Clinical evidence underscores why the timing of these updates matters. Prehospital ECGs were associated with faster reperfusion \citep{karagounis1990field,canto1997prehospital}, while stroke prenotification was associated with faster door-to-imaging and door-to-needle times \citep{lin2012prenotification}. Updating an assessment can also improve discrimination: risk models and early-warning scores perform better as ambulance and later observations become available \citep{spangler2019validation,martinrodriguez2023tracking}. These findings, largely observational, motivate decision support that becomes useful before the encounter is complete and remains responsive as new evidence arrives.

Despite this evidence for progressive decision-making, a corresponding form of AI assistance remains missing: a system that assesses plausible EMS protocols early, revises its shortlist as information arrives, and supports the resulting changes in workflow across stages. Existing EMS decision-support systems have explored concept extraction, multimodal sensing, and interactive assistance \citep{jin2023emsassist,EMSContExt,weerasinghe2024realtimemultimodalcognitiveassistant,jin2025smartglassesemergencymedicalservices,jin2025realtimemobilevideoanalytics}, while instruction-tuned large language models (LLMs) have demonstrated broad clinical capabilities \citep{llmclinicalsurvey,llm_zero_shot_clinical}. Yet these advances have not established how to provide and refine protocol assessments throughout the evolving encounter under a common, stage-specific evidence constraint. Evaluating predictions only after the clinical record is complete leaves the earlier decisions that shape preparation, assessment, and coordination unsupported. The missing capability is therefore to make protocol predictions useful while those decisions are still being made and to improve them as the encounter unfolds.

We address this gap by proposing a progressive EMS protocol-ranking framework intended to support dynamic workflow adaptation through assessments updated at successive information stages. We study 82 recorded protocol classes from the 2024 National Emergency Medical Services Information System (NEMSIS) public-release data \citep{nemsis2024rds,gregor2026nemsis} across six cumulative stages: dispatch, complaint location, complaint organ system, scene assessment, vital signs, and provider impressions. Developed with an EMS domain expert, these stages define a retrospective, workflow-informed feature-revelation schedule rather than a verified real-time sequence. Because NEMSIS lacks per-observation decision timestamps, we additionally examine timestamp-restricted vital signs and alternative reveal orders (App.~\ref{app:timing}).

A central observation further motivates our approach: early stage features contain unexploited gains which later stages miss. Although aggregate top-1 accuracy increases at every stage, 2.5--5.7\% of test encounters leave rank~1 at each transition (Figure~\ref{fig:rankflow}). On development data, \harmMin--\harmMax\% of encounters undergo such harmful transitions, and in \harmInTopThreeMin--\harmInTopThreeMax\% of these the earlier top-1 prediction remains in the later top-3. Earlier predictions can therefore retain useful signals even after a later model changes its leading assessment. We empirically investigate how pooling earlier-stage and current-stage posteriors can improve later protocol rankings. 

\begin{figure}[t]
\centering
\includegraphics[width=0.92\linewidth]{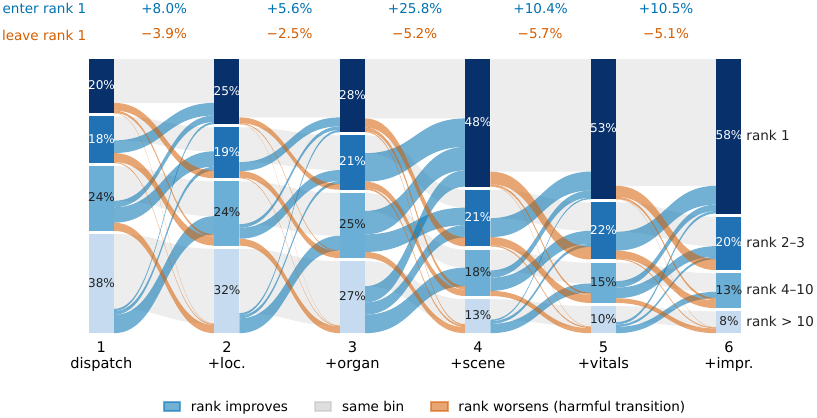}
\vspace{-1mm}
\caption{More evidence improves aggregate ranking but not every encounter. Rank of the recorded protocol under the unpooled three-seed ensemble flows between consecutive stages (blue: rank improves; vermillion: rank worsens, a harmful transition; grey: same bin). Printed shares are encounters entering and leaving rank~1 at each transition (aggregates in the repository's; see App.~\ref{app:diag}.}
\label{fig:rankflow}
\vspace{-4mm}
\end{figure}

In summary, our paper establishes the following contributions:

\begin{itemize}\itemsep0pt

\item Progressive Clinical Feature Learning (PCFL) formulates EMS protocol ranking through audited, nested feature blocks that creates predictions with stage-available evidence, using matched records and hit definitions across predictors. To our knowledge, machine learning literature has not established such a system in pre-hospital care. 

\item Stage-wise Ensembles for Temporal Pooling (STEP) learns to combine current-stage, earlier-stage, and modality-specific predictions to improve progressive protocol ranking. Motivated by harmful prediction transitions as features accumulate, controlled ablations and evaluations across datasets and experimental conditions characterize when pooling helps and distinguish its gains from seed ensembling, calibration, and adaptive gating.

\item SEP extends PCFL's information contract to LLMs by serializing only stage-permitted evidence, enabling comparisons with structured predictors on identical records under matched observability. We evaluate locally served Qwen3-32B-AWQ and MedGemma-27B-text in zero-shot and few-shot settings.

\item Retrospective analysis of deployment considerations evaluates sensitivity to natural class distributions, prior adaptation and calibration, alternative stage orders, and vital-sign timing. An exploratory Dispatch--Protocol Confidence Score (DPCS) further examines stage-dependent, dispatch-conditioned reliability annotations against posterior confidence.

\end{itemize}

\section{Related Work}
\label{sec:related}
\vspace{-4mm}
\textbf{EMS decision support.} EMSAssist pairs on-device speech recognition with protocol assistance and EMSConText builds an EMS concept lexicon \citep{jin2023emsassist,EMSContExt}; multimodal assistants integrate scene signals \citep{weerasinghe2024realtimemultimodalcognitiveassistant}, and smart-glasses and mobile-video systems extend sensing to wearable and remote platforms \citep{jin2025smartglassesemergencymedicalservices,jin2025realtimemobilevideoanalytics}. Whereas that work addresses signal capture and extraction, we study progressive inference as structured clinical evidence accumulates.

\textbf{Early prediction, stacking, and opinion pools.} Early time-series classification trades accuracy against earliness \citep{schaefer2020teaser}, anytime classifiers emit predictions as computation proceeds \citep{jazbec2023anytime}, and active feature acquisition selects which variable to query next \citep{li2021active}; here the acquisition order follows the clinical workflow, so the quantity of interest is the incremental value of each feature block as it arrives. Recurrent and attention models exploit repeated measurements in a fixed representation \citep{harutyunyan2019multitask,choi2016retain}, whereas an encounter's representation expands across heterogeneous block types, where tuned MLPs and tree ensembles remain strong baselines \citep{grinsztajn2022tree,gorishniy2021revisiting,kadra2021welltuned,borisov2022survey}. STEP builds on established ensemble and stacking principles: its stage and expert predictors are geometric-mean deep ensembles \citep{lakshminarayanan2017deepensembles}, and its pooling module TFA instantiates stacked generalization \citep{wolpert1992stacked,ting1999stacking} as a weighted logarithmic opinion pool or product of experts \citep{genest1986combining,hinton2002poe,kittler1998combining} with input-dependent gating \citep{jacobs1991mixtures}, fitted on out-of-fold posteriors to prevent in-sample overconfidence \citep{chernozhukov2018double}. Temporal recommenders re-rank as history grows \citep{tgsrec,recsys_temporal_dynamics} and confidence-aware ranking separates relevance from reliability \citep{recsys_confidence_beta}; because neural probabilities can be miscalibrated \citep{guo2017calibration}, we evaluate how empirical dispatch-conditioned reliability tiers compare alongside posterior confidence (\S\ref{sec:dpcsres}).

\textbf{LLMs on clinical tasks.} Instruction-tuned LLMs show zero- and few-shot competence across clinical benchmarks \citep{llmclinicalsurvey,llm_zero_shot_clinical}, and EMS-specific work explores prompting and retrieval augmentation \citep{expert_guided_rag_ems}. Ranking a closed set of 82 fine-grained protocols from stage-limited coded evidence differs from open-ended clinical question answering: the label set is fixed, the evidence is structured, and the binding constraint is which fields are eligible at each stage; SEP (\S\ref{sec:sep}) enforces that contract for instruction-tuned language models.

\begin{figure}[t]
\centering
\includegraphics[width=0.76\textwidth]{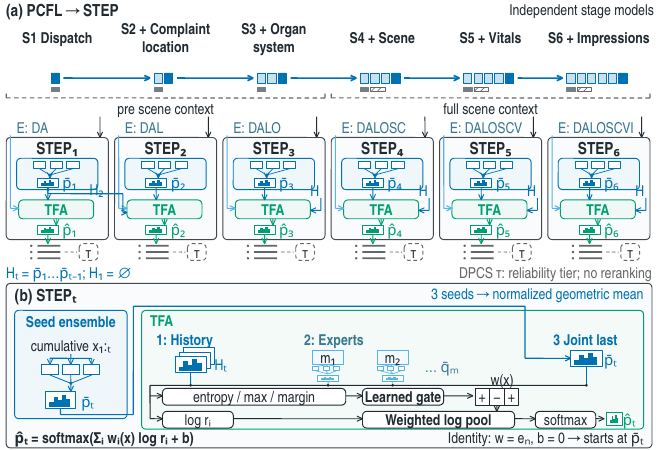}
\caption{Progressive EMS protocol prediction with STEP.
(a) Six cumulative stages S1--S6; each stage has an independently trained three-seed ensemble ($\bar p_t$), and TFA pools it with earlier unpooled stage posteriors ($H_t=\bar p_1,\ldots,\bar p_{t-1}$) and stage-eligible single-block experts into $\hat p_t$, whose top-3 shortlist ($\top$) is displayed with a DPCS reliability tier ($\tau$). ``E:'' lists active expert blocks: D dispatch, A age, L complaint location, O organ system, S symptoms, C scene context, V vitals, and I impressions (from S4 onward one context expert covers A and C). Pools are not chained.
(b) Inside stage $t$: the seed ensemble, the three input groups (history, experts, current joint posterior last), and the confidence-gated log-linear pool (Eqs.~\ref{eq:pool}--\ref{eq:gate}).}
\vspace{-5mm}
\label{fig:pipeline}
\end{figure}

\section{Progressive EMS Protocol Prediction}
\label{sec:method}
Our staged-evidence framework is shown in Figure~\ref{fig:pipeline}. Throughout, $t\in\{1,\ldots,6\}$ indexes the evidence stage, $e$ an encounter, $\mathcal{P}$ the candidate protocol set ($|\mathcal{P}|=82$), $y_e\in\mathcal{P}$ the recorded protocol, $I_e$ the dispatch complaint code, $\bar p_t$ the unpooled three-seed stage ensemble posterior, $\hat p_t$ the STEP pooled stage-$t$ posterior, and $\mathcal{K}_e^{(t)}$ the displayed top-$K$ shortlist ($K=3$); gate refers specifically to the pool's learned weighting network.

\subsection{Progressive Clinical Feature Learning (PCFL)}
\label{sec:pcfl}
\textbf{Task and stages.} Let $B_{e,1},\ldots,B_{e,6}$ denote the six ordered feature blocks of an encounter: dispatch, complaint location, complaint organ system, scene assessment and symptoms, vital signs, and provider impressions. At stage $t$,
\begin{equation}
 x_e^{(t)}=\operatorname{concat}\big(B_{e,1},\ldots,B_{e,t}\big)\in\mathbb{R}^{D_t},\qquad
 p_t\big(\cdot\mid x_e^{(t)}\big)=f_t\big(x_e^{(t)}\big)\in\Delta^{|\mathcal{P}|},
 \label{eq:stage}
\end{equation}
where $D_t$ is the cumulative feature dimension at stage $t$, $f_t$ is the stage-$t$ predictor, $p_t(y\mid x_e^{(t)})$ is its posterior probability of protocol $y$, and $\Delta^{|\mathcal{P}|}$ is the probability simplex over the $|\mathcal{P}|$ protocols. Ranking the posterior probabilities yields the shortlist $\mathcal{K}_e^{(t)}$ of the $K=3$ most probable protocols; using the indicator function $\mathbb{I}[\cdot]$, a hit at depth $K$ is $h_K(e,t)=\mathbb{I}[y_e\in\mathcal{K}_e^{(t)}]$, from which all accuracy metrics and paired tests are computed. Across stages 1--6, $D_t$ grows as \dimDdispatch, \dimDlocation, \dimDorgan, \dimDscene, \dimDvitals{}, and \dimDimpressions{} columns (App.~\ref{app:pcfl}, Table~\ref{tab:features}). Writing $X^{(t)}$ for the set of stage-$t$ feature columns, strict nesting $X^{(1)}\subset\cdots\subset X^{(6)}$ is verified by set comparison (App.~\ref{app:features}). Feature blocks and their order were defined with an EMS domain expert. Because stage membership follows audited NEMSIS workflow definitions (App.~\ref{app:pcfl}, Table~\ref{tab:capture}) rather than per-observation timestamps and primary vital signs are summarized over the call, PCFL defines a workflow-informed retrospective revelation protocol rather than verified pre-decision availability; each stage is a tabular classification problem, and App.~\ref{app:timing} evaluates sensitivity to timestamp-restricted vitals (Table~\ref{tab:e4timing}) and alternative reveal schedules (Table~\ref{tab:e6order}).

\subsection{Stage-wise Ensembles for Temporal Pooling (STEP)}
\label{sec:step}
At each stage, STEP forms a three-seed geometric-mean ensemble of the joint stage predictor and combines it with earlier-stage and single-modality posteriors through a confidence-dependent Temporal Feature Aggregation (TFA) log-linear pool (App.~\ref{app:step}).

\textbf{Stage predictor and seed ensemble.} Each stage predictor is an MLP with hidden widths $[256,128]$, ReLU activations, dropout $0.2$, and 82 output logits, trained with softmax cross-entropy (Adam, learning rate $10^{-3}$, batch size 256, 200 epochs, last-epoch weights; Table~\ref{tab:hparams}). Posteriors from fixed seeds $s\in\{0,1,2\}$ are combined via a normalized geometric mean,
\begin{equation}
 \bar p_t(y\mid x)=\frac{\prod_{s=0}^{2}p_t^{(s)}(y\mid x)^{1/3}}{\sum_{y'}\prod_{s=0}^{2}p_t^{(s)}(y'\mid x)^{1/3}},
 \label{eq:ens}
\end{equation}
equivalent to averaging and renormalizing log-softmax outputs. Single-modality experts use the same recipe on individual feature blocks; across stages, single-seed test top-1 accuracy varies by up to \seedSpreadMax{} percentage points (App.~\ref{app:stepens}, Tables~\ref{tab:steptopk}--\ref{tab:experts}).

\textbf{Temporal pooling (TFA).} At stage $t$, TFA combines two input axes with the current joint posterior $\bar p_t$ (App.~\ref{app:steppool}): stage history, comprising unpooled earlier-stage ensembles $\bar p_1,\ldots,\bar p_{t-1}$, and modality experts, comprising ensembled MLPs $q_{m,t}$ for each stage-eligible block (dispatch, location, organ system, symptoms, vitals, impressions, and context, where the context expert uses dispatch age at stages 1--3 and full scene context from stage 4 onward; Tables~\ref{tab:capture} and~\ref{tab:experts}). Pools are not chained. Placing $\bar p_t$ last yields $n=2t+1$ ordered inputs $r_1,\ldots,r_n$ (3 at stage 1 to 13 at stage 6):
\begin{align}
 \hat p_t(y\mid x)&=\operatorname{softmax}_y\Big(\textstyle\sum_{s=1}^{n} w_s(x)\,\log r_s(y\mid x)+b_y\Big),
 \label{eq:pool}\\
 w(x)&=w_0+W_2\tanh\big(W_1c(x)+b_1\big)+b_2.
 \label{eq:gate}
\end{align}
Here $c(x)\in\mathbb{R}^{3n}$ collects each input's Shannon entropy, maximum probability, and top-1/top-2 margin. $W_1,b_1$ and $W_2,b_2$ are the hidden (width 32) and output parameters of the gate, producing unconstrained class-tied weights $w(x)\in\mathbb{R}^n$ alongside per-class bias $b\in\mathbb{R}^{82}$ (App.~\ref{app:steppool}).

\textbf{Fitting and selection.} Identity initialization sets $w_0=\mathbf{e}_n$ (the $n$-th standard basis vector placing unit weight on $r_n=\bar p_t$) with $W_2=0$, $b_2=0$, and $b=0$, yielding $\hat p_t=\bar p_t$ at epoch 0. Under a single clean data boundary (App.~\ref{app:stepfit}), the training split is partitioned once into 66,402 fit rows and 7,378 inner-validation rows. Every base model is trained exclusively on the 66,402 fit rows, and the stage-$t$ pool is trained on five-fold out-of-fold (OOF) posteriors within those fit rows so that it never observes overconfident in-sample predictions. Pool checkpoints (including epoch 0) are selected by NLL on the 7,378 inner-validation rows (Table~\ref{tab:epochsel}), and the 8,198 test rows are evaluated once per frozen configuration.

\textbf{Scope of pooling.} Every pooled input respects stage-$t$ feature eligibility. At stages 1--5, $x_e^{(t)}$ contains all features used by earlier-stage and expert models, so pooling provides estimation benefits without additional patient information. At stage 6, the standard clinical block sums shared symptom and impression ICD-10-CM indicators; retraining the stage-6 joint model with separated symptom and impression columns (\provSepCols{} columns) changes ensemble accuracy by $\provSepMinusMergedEnsOne$ top-1 and $\provSepMinusMergedEnsThree$ top-3 points ($p=\provEnsPOne, \provEnsPThree$, not significant) while preserving the top-3 pooling gain ($\provSepDThree$, $p=\provSepPThree$; App.~\ref{app:stepscope}, Table~\ref{tab:provenance}). Because a class-tied log-linear pool over single-block experts is additive in log-odds, the joint predictor $\bar p_t$ is required to capture cross-modality interactions (App.~\ref{app:additivity}); empirically, pooling experts without $\bar p_t$ trails the ensemble by \expOnlyDeficitMin{} to \expOnlyDeficitMax{} top-1 points at stages 4--6 (App.~\ref{app:stepscope}).

\subsection{Dispatch--Protocol Confidence Score (DPCS)}
\label{sec:dpcs}
\begin{figure}[t]
\centering
\includegraphics[width=0.8\textwidth]{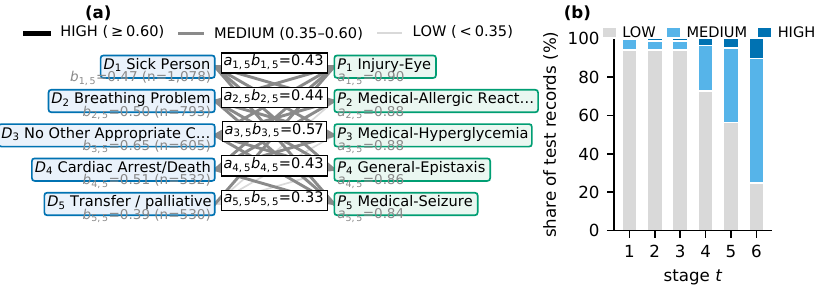}
\vspace{-8mm}
\caption{DPCS on the STEP shortlist. (a) Stage-5 dispatch--protocol reliability graph: $D_1$--$D_5$ are the five most frequent test dispatch codes with validation accuracy $b_{i,5}$, $P_1$--$P_5$ the five protocols with highest stage-5 validation recall $a_{j,5}$, and edges are shaded by the tier of $a_{j,5}b_{i,5}$. (b) Share of test records ($n=8{,}198$) in each tier across stages 1--6 (counts and accuracies in App.~\ref{app:dpcs}, Table~\ref{tab:dpcs}).}
\label{fig:dpcsmethod}
\vspace{-4mm}
\end{figure}

DPCS investigates stage-specific, dispatch-conditioned reliability annotation by combining validation protocol recall and dispatch-conditioned prediction accuracy (App.~\ref{app:dpcs}). For the $N_j$ inner-validation encounters with true protocol $j$ and the $M_i$ with dispatch complaint code $i$,
\begin{equation}
 a_{j,t}=\frac{1}{N_j}\sum_{r:\,y_r=j}\mathbb{I}[\hat y_{r,t}=j],\qquad
 b_{i,t}=\frac{1}{M_i}\sum_{r:\,I_r=i}\mathbb{I}[\hat y_{r,t}=y_r],
 \label{eq:ab}
\end{equation}
where $\hat y_{r,t}$ is STEP's stage-$t$ top-1 prediction on inner-validation encounter $r$ computed from the fit-rows base models. Using only the 7,378 inner-validation rows, the shortlist score and reliability tier $\ell_e^{(t)}$ are defined as:
\begin{equation}
 \operatorname{DPCS}_e^{(t)}=b_{I_e,t}\max_{j\in\mathcal{K}_e^{(t)}}a_{j,t},\qquad
 \ell_e^{(t)}=\begin{cases}
 \textsc{low}&\operatorname{DPCS}_e^{(t)}<0.35\\
 \textsc{medium}&0.35\le\operatorname{DPCS}_e^{(t)}<0.60\\
 \textsc{high}&\operatorname{DPCS}_e^{(t)}\ge0.60.
 \end{cases}
 \label{eq:dpcs}
\end{equation}
Figure~\ref{fig:dpcsmethod} shows (a) the stage-5 reliability graph with edge weights $a_{j,5}b_{i,5}$ and (b) test-set tier proportions across stages 1--6 (Table~\ref{tab:dpcs}). DPCS is a predictor-agnostic, descriptive summary of empirical validation hit rates conditioned on dispatch code and protocol; it does not rerank candidates, measure ground-truth dispatch correctness, or provide a calibrated probability or statistical confidence bound, and we evaluate its stratification and error-detection properties relative to posterior confidence in \S\ref{sec:dpcsres} and App.~\ref{app:dpcs}.

\subsection{Staged-Evidence Prompting (SEP) of Existing LLMs}
\label{sec:sep}
\textbf{Interface and prompt assembly.} SEP shares the supervised arm's stage definitions and per-field eligibility rules through a common module (App.~\ref{app:llm}): scene-time fields enter only from Stage 4 onward, unmeasured vitals render as \texttt{[not measured]}, and post-decision medications, procedures, and identifiers are excluded. Each prompt contains a byte-identical cached prefix (task, output contract, field legend, and 82 protocols as ordered \texttt{id<TAB>name} pairs), $k$ stage-matched demonstrations $\mathcal{D}_k^{(t)}$ sampled deterministically from distinct protocols in the fit rows, and the serialized stage-$t$ query $x_e^{(t)}$ (Figure~\ref{fig:prompt}; list-order ablations in App.~\ref{app:llmsubsets}).

\textbf{Output contract and instantiation.} The language model $g$ returns a JSON object ranking ten protocol identifiers, from which the top-$K$ shortlist is parsed:
\begin{equation}
 \rho_e^{(t)}=\operatorname{parse}\Big(g\big(\operatorname{prompt}_t\big(x_e^{(t)};\mathcal{D}_k^{(t)}\big)\big)\Big),\qquad
 \mathcal{K}_e^{(t)}=\rho_e^{(t)}[1{:}K],
 \label{eq:sep}
\end{equation}
where $\operatorname{parse}$ strips markdown fences or reasoning traces and parses the ranked IDs (scoring any unparseable output as a miss; invalid rate is $0.00\%$ for Qwen3 and $\le 0.21\%$ for MedGemma). Qwen3-32B-AWQ \citep{qwen25} is served locally via vLLM \citep{vllm} on two A30 GPUs (Table~\ref{tab:decoding}). Prompt format and $k\in\{0,3,5\}$ were selected on \texttt{val820} (an 820-row inner-validation subset; App.~\ref{app:stats}): $k=3$ improved validation top-1 and top-3 over $k=0$ and was selected, whereas $k=5$ harmed top-3 (App.~\ref{app:llmsubsets}). MedGemma-27B-text \citep{medgemma2025} (bf16 on three A30 GPUs) is evaluated at Qwen3's frozen prompt with $k\in\{0,3\}$ at seed 0 (App.~\ref{app:llmprotocol}). SEP matches stage-available input evidence without task-specific fine-tuning.
\section{Experimental Setup}
\label{sec:setup}
\textbf{Data and splits.} The 2024 NEMSIS public-release dataset \citep{nemsis2024rds,gregor2026nemsis} provides 81,978 single-protocol encounters across 82 protocols (approximately 1,000 encounters per protocol; App.~\ref{app:pcfl}). This balanced design weights all protocols equally so the stage curve measures protocol identifiability rather than class prevalence; deployment-like skew is evaluated separately on naturally distributed 2024 (90,759 records; 54\% in one protocol) and 2023 (100,000 records; 53\% in one protocol) cohorts alongside a balanced 2023 cohort (80,284 records; \S\ref{sec:dist}, App.~\ref{app:diag}). A 90/10 stratified outer split (seed 42) defines 73,780 training and 8,198 test encounters, and an inner 10\% split (seed 1234) defines 66,402 fit rows and 7,378 inner-validation rows that govern all selection, calibration, and DPCS rate estimation (App.~\ref{app:stepfit}).

\textbf{Comparators and statistical protocol.} We evaluate STEP against the unpooled three-seed stage MLP ensemble, a single-seed stage MLP (NN, seed 0), a twelve-seed joint stage ensemble (Tables~\ref{tab:steptopk} and~\ref{tab:parity}), equal-weight and static log-linear pools, Product Anytime (PA), and recalibrated ensembles on identical OOF inputs (Tables~\ref{tab:controls}--\ref{tab:e1pool}), tuned LSTM, Transformer, and cascaded neural baselines (Table~\ref{tab:architectures}), tuned and pooled XGBoost (Table~\ref{tab:gbm}), a single block-masked MLP ensemble serving all six stages (Tables~\ref{tab:e3cost} and~\ref{tab:flips}), and SEP with Qwen3-32B-AWQ and MedGemma-27B-text (Tables~\ref{tab:llm}--\ref{tab:perprotocol}). Top-1 and top-3 accuracy are co-primary endpoints, tested via exact two-sided paired McNemar tests \citep{mcnemar1947note,dietterich1998approximate} on per-record hit indicators $h_1$ and $h_3$ with bootstrap 95\% confidence intervals. The $\nstepfamily$ primary STEP-versus-ensemble tests and $\nllmfamily$ STEP-versus-LLM tests are Bonferroni-corrected at $0.05/\nstepfamily$ and $0.05/\nllmfamily$, respectively, while post-review ablations and diagnostics on the test split are reported with exact $p$-values and intervals (App.~\ref{app:stats}).

\section{Results}
\label{sec:results}

\begin{table}[t]
\centering\scriptsize\setlength{\tabcolsep}{3pt}
\caption{Balanced test split ($n=8{,}198$), repaired family, clean boundary: NN (seed-0 stage MLP), ens.\ (3-seed geometric mean) and STEP (ens.\ + gated log-linear pool), top-1 / top-3 (\%); $\Delta$: STEP $-$ ens., paired, bootstrap 95\% CI and exact McNemar $p$ (the corrected family of 12 tests, Bonferroni $0.05/12=0.0042$).}
\label{tab:main}
\vspace{4pt}
\resizebox{\textwidth}{!}{\begin{tabular}{lrrrll}
\toprule
stage & NN top-1 / top-3 & ens.\ top-1 / top-3 & STEP top-1 / top-3 & $\Delta_1$ [95\% CI] ($p_1$) & $\Delta_3$ [95\% CI] ($p_3$)\\
\midrule
1 dispatch & 20.36 / 38.09 & 20.42 / 38.02 & \textbf{20.72} / \textbf{38.34} & $+0.30$ [0.00, 0.61] (0.06) & $+0.32$ [-0.07, 0.72] (0.13)\\
2 +location & 24.54 / 43.38 & 24.54 / 43.79 & \textbf{24.63} / \textbf{43.91} & $+0.09$ [-0.18, 0.35] (0.59) & $+0.12$ [-0.29, 0.55] (0.6)\\
3 +organ system & 27.31 / 47.79 & 27.56 / 48.16 & \textbf{27.85} / \textbf{48.39} & $+0.29$ [-0.04, 0.61] (0.098) & $+0.23$ [-0.20, 0.65] (0.3)\\
4 +symptoms, scene & 45.84 / 67.32 & 48.16 / 69.60 & \textbf{48.24} / \textbf{70.52} & $+0.09$ [-0.51, 0.67] (0.8) & $+0.91$ [0.39, 1.43] ($6.2\!\times\!10^{-4}$)\\
5 +vitals & 50.34 / 72.13 & 52.94 / 74.53 & \textbf{53.71} / \textbf{75.56} & $+0.77$ [0.24, 1.31] (0.0051) & $+1.02$ [0.54, 1.51] ($3.7\!\times\!10^{-5}$)\\
6 +impressions & 56.03 / 75.98 & 58.34 / 78.56 & \textbf{59.31} / \textbf{79.82} & $+0.96$ [0.38, 1.51] ($7.7\!\times\!10^{-4}$) & $+1.27$ [0.78, 1.77] ($6.8\!\times\!10^{-7}$)\\
\bottomrule
\end{tabular}}
\end{table}

\begin{figure}[t]
\centering
\includegraphics[width=0.8\textwidth]{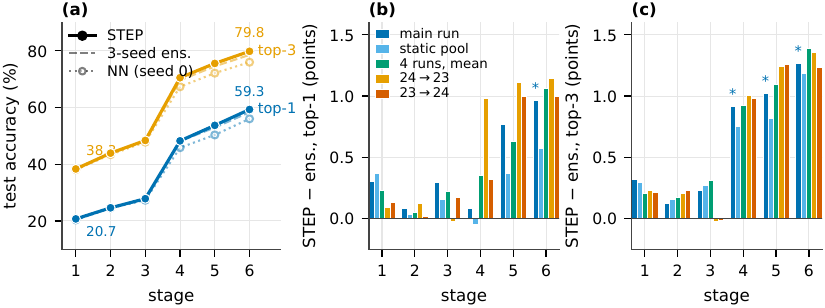}
\caption{Balanced test split ($n=8{,}198$; values in Tables~\ref{tab:main} and~\ref{tab:robust}). (a) Stage-wise top-1 and top-3 accuracy for the single-seed MLP (NN, seed 0), three-seed ensemble, and STEP. (b, c) Paired STEP minus ensemble increments (percentage points) across five evaluation settings (stars: main-run comparisons clearing $0.05/\nstepfamily$).}
\label{fig:main}
\vspace{-5mm}
\end{figure}

\subsection{How do stages and pooling affect ranking accuracy and robustness?}
\label{sec:stepres}
\label{sec:robust}
Protocol-ranking accuracy increases across all stages, from \stepTopOnedispatch\% to \stepTopOneimpressions\% top-1 and \stepTopThreedispatch\% to \stepTopThreeimpressions\% top-3, with Stage 4 contributing the largest gain (+\stepGainOneOneDecscene{} top-1 points; Table~\ref{tab:main}, Figures~\ref{fig:main}--\ref{fig:timing}). Three-seed ensembling accounts for most late-stage improvements over single-seed predictions, while STEP further improves top-3 accuracy at Stages 4--6 and top-1 at Stage 6 after Bonferroni correction. Pooling also consistently reduces test NLL (Stage 6: \stageSixEnsNll{} to \stageSixStepNll; Table~\ref{tab:controls}). These gains persist across four disjoint-seed replications, bidirectional cross-year transfer, and restrictions on vital-sign timing (Tables~\ref{tab:reps}, \ref{tab:year}, \ref{tab:e4timing}). Alternative reveal orders affect intermediate-stage accuracy but preserve Stage 4--6 top-3 pooling gains of $\onEsixThreeIncMin$ to $\onEsixThreeIncMax$ points ($p\le\onEsixThreePmax$; Table~\ref{tab:e6order}).

Controlled comparisons clarify the source of pooling gains. Equal-weight pools underperform the three-seed ensemble from Stage \eoneEqWorseFrom{} onward, whereas learned static log-linear pooling \citep{wolczyk2021ztw,heskes1998selecting} achieves 58.92\% / 79.74\% top-1/top-3 at Stage 6, recovering most top-3 and roughly half to three-fifths of top-1 improvements (Table~\ref{tab:e1pool}). Confidence-dependent TFA gating provides modest additional accuracy gains that do not survive multiple-comparison correction but consistently reduces NLL (Stage 6: $\onNllSix$ nats, adjusted CI $[\onNllSixLo,\onNllSixHi]$). A twelve-seed joint ensemble achieves comparable Stage 6 accuracy (59.29\% / 79.76\%), while increasing STEP to twelve seeds per input yields marginal additional accuracy and lower NLL (Table~\ref{tab:parity}). Overall, learned weighting is essential for effective pooling of heterogeneous stage predictions, larger joint ensembles offer an alternative for top-$k$ accuracy, and adaptive gating primarily improves probabilistic predictions.

\subsection{What changes under natural and multi-year distributions?}
\label{sec:dist}
\label{sec:prior}
\begin{figure}[t]
\centering
\includegraphics[width=\textwidth]{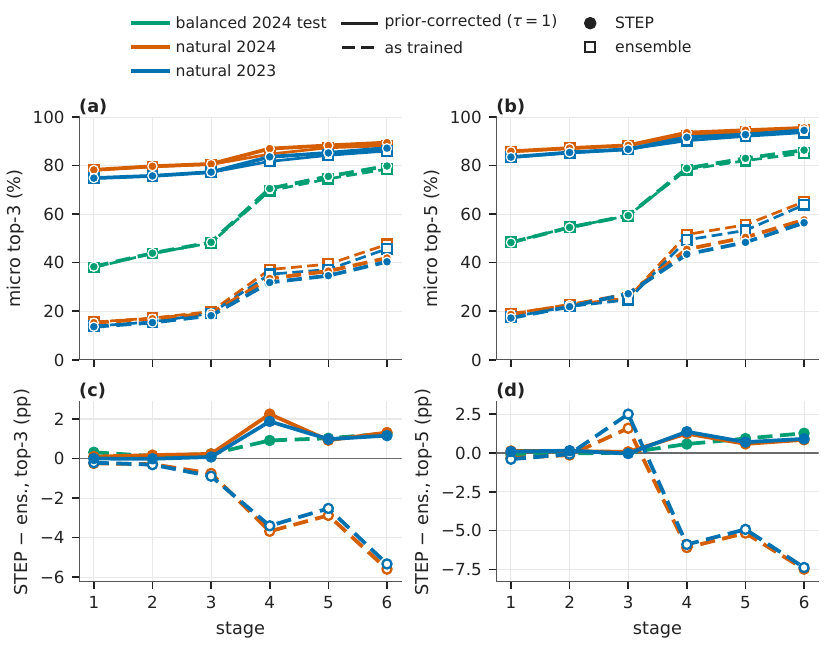}
\caption{Micro (a) top-3 and (b) top-5 accuracy on balanced 2024 ($n=8{,}198$), natural 2024 ($90{,}759$), and natural 2023 ($100{,}000$) cohorts, as trained (dashed) and with oracle prior correction at $\tau{=}1$ (solid). (c, d) Paired STEP (filled circles) minus ensemble (open squares) differences in percentage points.}
\label{fig:dist}
\vspace{-6mm}
\end{figure}
Unlike balanced evaluation, natural cohorts exhibit severe class imbalance, with one protocol comprising \natExtMajorityPctInt\% and \natNatMajorityPctInt\% of the 2024 and 2023 cohorts, respectively. We evaluate balanced-trained models as trained, with oracle prior correction \citep{saerens2002adjusting} $p_\tau(y\mid x)\propto p(y\mid x)(\pi_{\text{target}}(y)/\pi_{\text{train}}(y))^\tau$ at $\tau{=}1$, and with label-free expectation--maximization (EM) adaptation (Figure~\ref{fig:dist}, Tables~\ref{tab:dist}--\ref{tab:e2em}).

\emph{As trained.} STEP improves 2024 Stage 6 macro top-1 from \extEnsMacroOneimpressions\% to \extStepMacroOneimpressions\% but reduces micro top-3 by $\distExtAsDThreeimpressions$ and $\distNatAsDThreeimpressions$ points (signed differences) on 2024 and 2023, respectively, because its balanced-fitted class bias shifts probability away from the dominant protocol.
\emph{Oracle prior correction.} STEP outperforms the ensemble on every metric, reaching Stage 6 micro top-3 of \distExtTauStepThreeimpressions\% and \distNatTauStepThreeimpressions\% on 2024 and 2023, respectively, with gains over the ensemble of $\distExtTauDThreeimpressions$ and $\distNatTauDThreeimpressions$ points.
\emph{Label-free EM.} Disjoint-half EM adaptation yields Stage 6 top-1/top-3 STEP--ensemble differences of $\etwoNatStepEnsOneSix$ / $\etwoNatStepEnsThreeSix$ on 2024 and $\etwoOldStepEnsOneSix$ / $\etwoOldStepEnsThreeSix$ on 2023. Given EM's calibration sensitivity \citep{alexandari2020maximum}, source temperature scaling shifts these differences to $\etwoNatStepEnsTsOneSix$ / $\etwoNatStepEnsTsThreeSix$ and $\etwoOldStepEnsTsOneSix$ / $\etwoOldStepEnsTsThreeSix$, respectively (Table~\ref{tab:e2em}). Thus, protocol-level gains transfer across years, whereas micro-accuracy gains depend on target priors and calibration.

\subsection{Why does stage pooling help, and how do alternative architectures compare?}
\label{sec:controls}
\vspace{-2mm}
Under a unified validation sweep, no LSTM, Transformer, or cascaded network significantly beats the per-stage MLP ensemble at any stage, trailing by \archTrailMin{} to \archTrailMax{} top-1 points at Stages 5--6 (Table~\ref{tab:architectures}). A tuned XGBoost baseline beats the single-seed MLP by \gbmMinusNnOneOneDecscene{} to \gbmMinusNnOneOneDecvitals{} top-1 points at Stages 4--6 and comes within one point of the MLP ensemble; STEP leads tuned XGBoost by $\stepMinusGbmOneOneDecscene$ to $\stepMinusGbmOneOneDecimpressions$ ($p\le\gbmPmaxOne$; Table~\ref{tab:gbm}), and applying the same pool to cross-fitted XGBoost posteriors adds $\gbmPoolOnevitals$ / $\gbmPoolThreevitals$ and $\gbmPoolOneimpressions$ / $\gbmPoolThreeimpressions$ at Stages 5--6 over the single boosted model ($p\le\gbmPoolPmax$). A single block-masked MLP ensemble serving all six stages without pooling (3 network fits vs.\ \onFitsPooled{}; Tables~\ref{tab:e3cost},~\ref{tab:flips}) reaches 58.82\% Stage 6 top-1 accuracy and cuts $5\rightarrow 6$ harmful flips from 5.07\% to 2.94\%.

\textbf{Mechanism and generality} (App.~\ref{app:diag}; Tables~\ref{tab:cong},~\ref{tab:flips},~\ref{tab:synthetic}--\ref{tab:covtype}). At each transition, \harmMin\% to \harmMax\% of records flip from correct top-1 at stage $t{-}1$ to incorrect at stage $t$, and for \harmInTopThreeMin\% to \harmInTopThreeMax\% of them the true protocol remains in the stage-$t$ top-3 (Figure~\ref{fig:rankflow}). Soft log-linear pooling recovers these records whereas hard reversion rules and stage-congruent distillation \citep{yan2021positive} do not, and in a conjunction test the joint model recovers \conjJointMin--\conjJointMax\% of multi-expert agreements missed by an experts-only pool versus \conjDecisionMin--\conjDecisionMax\% for decision-level combiners. Regularizing the joint model in synthetic experiments removes at most 21\% of the pooling gain (Table~\ref{tab:synreg}), and on UCI Covertype (Table~\ref{tab:covtype}) pooling is inert across low-dimensional blocks (within $\pm\covUnmodMaxAbs$, $p>\covUnmodMinP$) but adds $\covPadOneFiveK$ / $\covPadThreeFiveK$ at $n=5{,}000$ ($p=\covPadPOneFiveK$) and $\covPadOneTwentyK$ / $\covPadThreeTwentyK$ at $n=20{,}000$ ($p=\covPadPOneTwentyK$) and halves harmful transitions (\covHarmEnsFiveK\% to \covHarmGatedFiveK\%) once a 500-column weakly informative block is introduced, consistent with finite-sample variance reduction.

\begin{figure}[t]
\centering
\includegraphics[width=\textwidth]{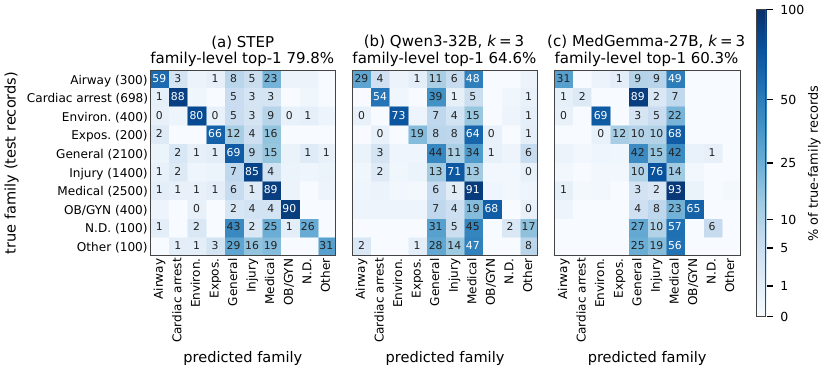}
\caption{Stage-6 top-1 predictions collapsed to the ten coarse protocol families, balanced test split (printed when $\ge 0.5\%$; square-root colour scale): (a) STEP, (b) Qwen3-32B-AWQ $k{=}3$ seed 0, (c) MedGemma-27B-text $k{=}3$ seed 0.}
\label{fig:conffamily}
\end{figure}

\subsection{How do prompted LLMs compare under the same stage contract?}
\label{sec:sepres}
Figure~\ref{fig:conffamily} and Figure~\ref{fig:sep} (App.~\ref{app:llm}, Tables~\ref{tab:llm}--\ref{tab:perprotocol}) compare STEP record-for-record with Qwen3-32B-AWQ and MedGemma-27B-text. Both prompted LLMs reproduce the progressive stage curve, including the Stage 4 jump, but concentrate predictions heavily into the General and Medical families and under-predict specific families such as Airway (\famQwenPredAirway{} and \famGemmaPredAirway{} predictions vs.\ \famStepPredAirway{} for STEP out of \famQwenTrueAirway{} records). About half of each model's errors remain within the true family (STEP \famStepWithin\%, Qwen3 \famQwenWithin\%, MedGemma \famGemmaWithin\%). Under matched evidence, both prompted LLMs trail STEP at every stage ($p<0.05/\nllmfamily$ across all $\nllmfamily$ tests) and the gap widens as high-dimensional coded features accumulate (Stage 6 top-1: 59.31\% STEP vs.\ 33.29\% Qwen3 and 28.84\% MedGemma at $k{=}3$). STEP achieves higher per-protocol F1 on \llmStepFoneHigher{} of 82 protocols (\llmStepFoneTies{} tie; Table~\ref{tab:perprotocol}) and runs in \onLatStepCold~ms cold (\onLatStepIncr~ms cached) on one CPU thread versus \costLlmSeqS~s sequential (\costLlmAmortMs~ms amortized) on GPU for Qwen3 (Table~\ref{tab:e3cost}).

\subsection{What does DPCS reveal about contextual reliability versus model confidence?}
\label{sec:dpcsres}
\vspace{-2mm}
Under fixed thresholds (0.35 and 0.60), DPCS conservatively assigns at least \dpcsLowShareBefore\% of test records to \textsc{low} at Stages 1--3. As clinical evidence accumulates, the distribution shifts upward, with 64.7\% of records classified as \textsc{medium} by Stage 6 (Table~\ref{tab:dpcs}). From Stage 4 onward ($n\ge\dpcsMinBucket$ per tier), accuracy increases monotonically across tiers (Stage 6 top-1: \dpcsLowTopOneSix\%, \dpcsMedTopOneSix\%, and \dpcsHighTopOneSix\% for \textsc{low}, \textsc{medium}, and \textsc{high}, respectively; Figure~\ref{fig:dpcs}). However, STEP's maximum posterior provides stronger error detection (Stage 6 AUROC: \dpcsAurocMaxPostSix{} vs.\ \dpcsAurocSix; Table~\ref{tab:dpcsscores}). Adding unsmoothed DPCS to posterior confidence yields no significant improvement in held-out log loss ($|\Delta\mathrm{LL}|\le\onEfiveMaxAbsDll{}$ nats; all 12 CIs include zero; Holm $p\ge\onEfiveMinHolm$; Table~\ref{tab:e5dpcs}). Thus, DPCS remains an exploratory, interpretable measure of contextual reliability, while posterior confidence is preferable for offline error detection. Its utility for clinician reliance requires prospective human-factors evaluation (App.~\ref{app:dpcs}).

\section{Conclusion}
\label{sec:conclusion}
This work formulates prehospital protocol ranking as progressive prediction under accumulating clinical evidence. PCFL defines six auditable, workflow-informed evidence stages, SEP enables matched-observability comparisons with language models, STEP investigates stage-wise and modality-specific opinion pooling, and DPCS examines dispatch-conditioned reliability tiers. Across 82 NEMSIS protocols, ranking accuracy rises from \stepTopOnedispatch\% top-1 (\stepTopThreedispatch\% top-3) at dispatch to \stepTopOneimpressions\% top-1 (\stepTopThreeimpressions\% top-3) after provider impressions, consistently outperforming the evaluated zero-shot and few-shot prompted Qwen3-32B-AWQ and MedGemma-27B-text configurations under matched evidence constraints. Controlled ablations demonstrate that learned pooling improves late-stage accuracy over a three-seed ensemble across seed replications, bidirectional cross-year transfer, and timestamp-restricted vital signs, while establishing that a twelve-seed joint ensemble attains comparable accuracy at lower training cost, TFA gating refines held-out log-likelihood, and posterior confidence detects errors more effectively than unsmoothed dispatch--protocol hit rates. Because our stages evaluate workflow-defined evidence availability retrospectively against historically documented protocols, translating progressive protocol ranking to prospective clinical decision support requires three validation steps: silent logging against verified real-time field-entry timestamps, independent clinical adjudication of protocol appropriateness, and human-factors studies with practicing EMTs evaluating cognitive workload, interface presentation, and appropriate reliance.

\section*{Ethics Statement}
This work concerns sensitive prehospital records and a safety-critical decision context. The system predicts historical protocol labels for research and assistive ranking; it is not an autonomous treatment system. No patient-level record, identifier or example appears in this paper or its figures; all qualitative material is class-level aggregate. A hosted-LLM arm (commercial API models such as GPT, Claude or Gemini) was deliberately not run pending a data-use determination, since sending prehospital records to a third-party API raises privacy and data-governance questions that the local arms avoid, and the local arm never places an identifier, a medication or a procedure in a prompt. Prospective use would require data governance, institutional oversight, fairness and distribution-shift evaluation, validation with EMS professionals, and an interface that preserves clinician authority.

\section*{Reproducibility Statement}
Our code is available for reproducing results at \url{https://anonymous.4open.science/r/step-ems-protocol-F025/}. Splits derive from seeds 42 (outer, 73,780 / 8,198) and 1234 (inner, 66,402 / 7,378) and the loader asserts row alignment across stages; the LLM subsets are derived at load time and SHA-256 hashed (App.~\ref{app:llm}). App.~\ref{app:pcfl} gives the cohort construction, capture-time audit, defect measurements, the exact stage-model hyperparameters and splits; App.~\ref{app:llm} the prompt template, decoding and serving configuration; App.~\ref{app:step} the pool hyperparameters, experts, the all-stage STEP decomposition and the pool-reproducibility experiments; App.~\ref{app:stats} the historical acceptance test, the multiplicity policy and the per-wave summary of every tested comparison, whose full machine-readable ledger ships with the code. Every table and figure is generated from the saved logits and result files by scripts in the accompanying repository, which record the track, training protocol, ensemble, split and $n$ behind each number.

\section*{AI Use Statement}
Generative AI assisted literature discovery, manuscript restructuring, language editing, and the scripts that regenerate tables and figures from saved artifacts. It did not generate experimental observations or alter results. The authors verified the assisted material and take responsibility for the final content.

\bibliography{references}
\bibliographystyle{iclr2027_conference}

\clearpage
\appendix
\renewcommand{\topfraction}{0.9}\renewcommand{\bottomfraction}{0.6}\renewcommand{\textfraction}{0.08}
\renewcommand{\floatpagefraction}{0.75}\setcounter{topnumber}{3}\setcounter{totalnumber}{5}

\paragraph{Abbreviations used in the appendix.} PCFL: Progressive Clinical Feature Learning (\S\ref{sec:pcfl}); STEP: Stage-wise Ensembles for Temporal Pooling (\S\ref{sec:step}); TFA: Temporal Feature Aggregation, STEP's pooling step; DPCS: Dispatch--Protocol Confidence Score (\S\ref{sec:dpcs}); SEP: Staged-Evidence Prompting (\S\ref{sec:sep}); NN: the matched single-seed stage network (seed 0); ens.: the three-seed ensemble without pooling; MLP: multilayer perceptron; LSTM: long short-term memory network; OOF: out-of-fold; NLL: negative log-likelihood; LLM: large language model; EMS: Emergency Medical Services; NEMSIS: National Emergency Medical Services Information System; ICD-10-CM: International Classification of Diseases, Tenth Revision, Clinical Modification; AUROC: area under the receiver operating characteristic curve; AURC: area under the risk--coverage curve; EM: expectation--maximization; PA: Product Anytime, the fixed-weight cumulative product of \citet{jazbec2023anytime}; KL: Kullback--Leibler divergence; ECG: electrocardiogram; GCS: Glasgow Coma Scale; AVPU: alert / voice / pain / unresponsive responsiveness scale; ETCO$_2$: end-tidal carbon dioxide; CAD: computer-aided dispatch; AWQ: activation-aware weight quantization; CPU / GPU: central / graphics processing unit; JSON: JavaScript Object Notation; UCI: University of California, Irvine (machine-learning repository). ``Repaired family'' denotes the feature matrices after the data repairs of App.~\ref{app:pcfl}; every reported number is from it unless a passage is marked \emph{historical (pre-repair)}.

\section{PCFL: Cohort, Features, Splits and Architectures}
\label{app:pcfl}

\begin{table}[h]
\centering\footnotesize
\begin{tabular}{p{4.6cm}p{8.3cm}}
\toprule
stage model (\texttt{direct\_nn}) & value \\
\midrule
hidden widths / activation / dropout & $[256,128]$ / ReLU / 0.2 after each hidden layer \\
normalisation / feature scaling & none / none (17 continuous columns enter raw: five vitals as mean and standard deviation, GCS mean and minimum, pain mean and maximum, ETCO$_2$ mean and standard deviation, age) \\
output / loss & 82 logits / unweighted softmax cross-entropy \\
optimiser / learning rate / schedule / weight decay & Adam / $10^{-3}$ / none / 0 \\
batch size / epochs / checkpoint & 256 / 200 / last epoch (best-validation epoch tested and rejected) \\
weight initialisation & PyTorch \texttt{nn.Linear} default (Kaiming-uniform, $a=\sqrt5$) \\
seeds and splits & model seeds $\{0,1,2\}$ per dataset; outer split seed 42, inner split seed 1234; cross-fitting: stratified five-fold split of the fit rows (split seed = model seed), fold-model initialisation seed $100\cdot\text{seed}+\text{fold}$ \\
hardware & 64 CPU cores, 3$\times$ NVIDIA A30 (24 GB); stage models and experts run concurrently on CPU/GPU, the LLM arm on two GPUs \\
\bottomrule
\end{tabular}
\caption{Stage-model hyperparameters and hardware, read from the code (\texttt{scripts/paper/review/train\_base.py}, \texttt{src/nemsis\_ems/temporal\_refinement.py}); the feature set is called config D in the generated tables.}
\label{tab:hparams}
\end{table}

\paragraph{Cohort.} NEMSIS 2024 public release; \texttt{Pub\_PCRevents.txt} (dispatch, complaint location and organ system, scene/context and age), \texttt{FACTPCRVITAL.txt} (timestamped vitals), \texttt{PCRVITALECGGROUP.txt} (ECG rhythm), with medications and procedures read only to be excluded and, in the timing check of App.~\ref{app:timing}, for their timestamps. About 1,000 encounters were sampled per protocol for the 82 protocols with enough volume (978--1,000 retained per class). Invalid and ``not recorded'' sentinels are stripped before code features are built. Procedure coverage is 69.27\% of the cohort (683 distinct codes), and 94.9\% of procedures with a recorded answer to ``performed prior to this unit's care'' were performed by the documenting unit, and hence plausibly after the protocol decision. The feature matrices were rebuilt after three data repairs (a vitals join that had dropped readings, clinical codes attached to the wrong rows, and a field-delimiter defect in \texttt{Pub\_PCRevents.txt}); every number in this paper is from the repaired matrices, except the historical acceptance tests of App.~\ref{app:stats}.

\paragraph{Features.}\label{app:features} Every feature is an indicator, a small count, or a raw value carrying its own missing flag; nothing is scaled, normalised or imputed. Vitals are re-joined from the raw tables so that every value column gets its own missing indicator and an unmeasured vital never collapses onto a measured zero; glucose is retained, and ECG rhythm (multi-hot), GCS mean and minimum, AVPU, pain and ETCO$_2$ are added on the same terms. Age enters as a raw value, a missing flag and eight fixed clinical bands at 1, 5, 13, 18, 40, 65 and 80 years. Table~\ref{tab:features} gives the block breakdown per stage; the stage column lists $X^{(t)}$ are nested, which the build verifies by set comparison. The information in the context block is carried by incident location type (169 codes) and age, the other three scene-time fields being near-constant. The merged \texttt{clinical\_*} block sums four source indicators: the symptom vocabulary has 1,063 codes, the impression vocabulary 1,246, and 571 codes are in both, so the union is the 1,738-column stage-6 block, in which 43.7\% of records carry a value of 2 (usually one ICD-10-CM code documented both as a symptom and as an impression) and three cells reach 3. This is not a label leak, since both fields are recorded before the protocol decision, but it does discard which field a shared code came from; the provenance-preserving alternative (the stage-5 matrix with the impression block kept separate, \provSepCols{} columns) is trained and compared in Table~\ref{tab:provenance} (App.~\ref{app:stepscope}).

\begin{table}[t]
\centering\footnotesize\setlength{\tabcolsep}{4pt}
\caption{Cumulative feature stages of the repaired family. $D_t$ is counted from the built datasets; every transition adds columns and drops none (the stage-$(t{-}1)$ columns are a subset of the stage-$t$ columns with identical values at stages 1--5; at stage 6 the merged clinical vocabulary changes the values of the shared symptom columns, App.~\ref{app:step}). The context block is 10 columns (age) before symptoms arrive and 190 after, the capture-time audit of App.~\ref{app:pcfl} (Table~\ref{tab:capture}) made visible in the data; the clinical block is one merged symptom/impression vocabulary (1,063 codes at stages 4--5; 1,738 at stage 6).}
\label{tab:features}
\vspace{4pt}
\begin{adjustbox}{max width=\linewidth}
\begin{tabular}{clrrl}
\toprule
stage & block revealed & $D_t$ & new columns & composition\\
\midrule
1 & Dispatch & 56 & 56 & disp.\ 46; context 10\\
2 & +Complaint location & 66 & 10 & disp.\ 46; loc.\ 10; context 10\\
3 & +Organ system & 78 & 12 & disp.\ 46; loc.\ 10; organ 12; context 10\\
4 & +Symptoms, scene context & 1,321 & 1,243 & disp.\ 46; loc.\ 10; organ 12; clinical 1,063; context 190\\
5 & +Vitals & 1,398 & 77 & disp.\ 46; loc.\ 10; organ 12; clinical 1,063; vitals 77; context 190\\
6 & +Impressions & 2,073 & 675 & disp.\ 46; loc.\ 10; organ 12; clinical 1,738; vitals 77; context 190\\
\bottomrule
\end{tabular}
\end{adjustbox}
\end{table}

\begin{table}[h]
\centering\scriptsize\setlength{\tabcolsep}{3pt}
\begin{tabular}{llp{1.3cm}p{7.5cm}}
\toprule
field & NEMSIS element & capture time & evidence from the element's own definition \\
\midrule
\texttt{ctx\_first\_ems} & eScene.01 & scene & ``first EMS unit \ldots\ \emph{on the scene}'': an arrival fact \\
\texttt{ctx\_num\_patients} & eScene.06 & scene & ``how many patients were \emph{at the scene}''; NONE = ``crew were unable to locate anyone'' \\
\texttt{ctx\_mci} & eScene.07 & scene & ``\emph{overwhelmed} \ldots\ by the number and severity of victims at an emergency scene'' \\
\texttt{ctx\_location\_type} & eScene.09 & scene & a two-tier pick list that \emph{EMS personnel} choose from, not an address derivation \\
\texttt{ctx\_age\_years} & ePatient.15/16 & dispatch & a demographic with no scene-observation dependency; a CAD element in its own right (eCad.66) \\
\bottomrule
\end{tabular}
\caption{Per-field capture timing from the NEMSIS element definitions. Scene-time fields may not appear before stage 4; age is the one dispatch-time field, and the one whose placement rests on a judgement rather than on the element's own wording.}
\label{tab:capture}
\end{table}

\begin{figure}[h]
\centering
\includegraphics[width=0.49\textwidth]{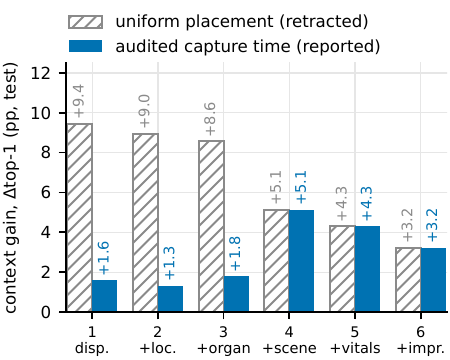}\hfill
\includegraphics[width=0.49\textwidth]{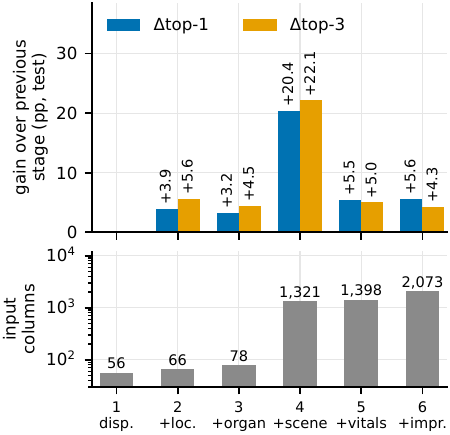}
\caption{Left: the gain from the scene/context block under the audited capture timing against the uniform placement the audit replaced. Right: per-transition gain of STEP (top) and the input dimension (bottom; log axis).}
\label{fig:timing}
\end{figure}

\paragraph{The capture-time audit in the data.} Figure~\ref{fig:timing} contrasts the gain of the context block under the audited per-field timing with the uniform placement of all five fields at every stage that the audit replaced: uniform placement credited stages 1--3 with \ctxExcessdispatch, \ctxExcesslocation{} and \ctxExcessorgan{} top-1 points that the fields cannot legitimately supply, and the two placements coincide from the scene stage on. Table~\ref{tab:context} gives the gain of the audited block at every stage; it peaks at the scene stage ($\ctxGainScene$ top-1), where the four scene-time fields legitimately arrive, and the right panel of Figure~\ref{fig:timing} shows the same pivot as the per-transition gain of STEP against the input dimension.

\begin{table}[h]
\centering
\scriptsize
\setlength{\tabcolsep}{2.5pt}
\caption{The scene/context + age block at its audited capture time, repaired family: three-seed ensembles trained on the fit rows without context versus with context placed when each field legitimately exists (age from stage 1, the four scene-time fields from stage 4). Accepted at every stage on both metrics. The gain peaks at stage 4 (where the four scene-time fields legitimately arrive), not at dispatch. Validation $p$: the historical acceptance test on the inner-validation rows; test deltas and exact McNemar $p$ on the test split, read once.}
\label{tab:context}
\vspace{4pt}
\begin{tabular}{crrrrrrrr}
\toprule
stage & no ctx.\ top-1 & timed top-1 & $\Delta$top-1 & $\Delta$top-3 & val $p_1$ & val $p_3$ & test $p_1$ & test $p_3$\\
\midrule
1 & 18.83 & \textbf{20.42} & $+1.59$ & $+2.09$ & $5.3\!\times\!10^{-9}$ & $1.1\!\times\!10^{-11}$ & $4.0\!\times\!10^{-9}$ & $2.3\!\times\!10^{-12}$\\
2 & 23.24 & \textbf{24.54} & $+1.31$ & $+1.87$ & $6.1\!\times\!10^{-7}$ & $2.7\!\times\!10^{-6}$ & $1.0\!\times\!10^{-5}$ & $2.4\!\times\!10^{-8}$\\
3 & 25.76 & \textbf{27.56} & $+1.79$ & $+2.16$ & $1.4\!\times\!10^{-6}$ & $3.6\!\times\!10^{-8}$ & $1.5\!\times\!10^{-9}$ & $1.1\!\times\!10^{-10}$\\
4 & 43.02 & \textbf{48.16} & $+5.14$ & $+5.76$ & $5.1\!\times\!10^{-34}$ & $9.0\!\times\!10^{-29}$ & $3.3\!\times\!10^{-32}$ & $2.8\!\times\!10^{-43}$\\
5 & 48.63 & \textbf{52.94} & $+4.31$ & $+4.07$ & $5.5\!\times\!10^{-16}$ & $2.8\!\times\!10^{-15}$ & $3.5\!\times\!10^{-25}$ & $1.2\!\times\!10^{-28}$\\
6 & 55.14 & \textbf{58.34} & $+3.21$ & $+2.65$ & $3.0\!\times\!10^{-18}$ & $1.4\!\times\!10^{-10}$ & $1.0\!\times\!10^{-16}$ & $1.3\!\times\!10^{-14}$\\
\bottomrule
\end{tabular}
\end{table}

\paragraph{Splits and training protocol.} Outer split: 90/10 stratified on protocol, seed 42 (73,780 training / 8,198 test). Inner split: 10\% of the training split stratified on protocol, seed 1234 (66,402 fit rows / 7,378 inner-validation rows). The six stage datasets are built from the same cohort in the same row order, so the split is identical across stage directories; the loader raises unless the label and identifier vectors match exactly. One data boundary (fit / inner-validation / test) underlies every reported number; App.~\ref{app:stepfit} states it in full. In short: every base model is trained on the 66,402 fit rows only; its out-of-fold posteriors (five folds within the fit rows) are the pool's training inputs; its predictions on the inner-validation rows are the selection quantities; and its predictions on the 8,198 test rows, read once, are the reported quantities. No model trained on the whole training split is used.

\paragraph{Alternative architectures.} Table~\ref{tab:architectures} compares the per-stage MLP with an LSTM and a Transformer over the stage blocks and with two cascades that train all six stages jointly and pass the previous stage's output forward, on the identical records, features, splits and seeds, under one selection policy that also applies to the MLP control (its caption gives the policy and the validation sweep). No alternative significantly beats the MLP ensemble at any stage (small non-significant leads occur at stages 1--2); at stages 5--6 all trail it by \archTrailMin{} to \archTrailMax{} top-1 points. The single seeds, the three-seed ensemble and STEP at every depth are in Table~\ref{tab:steptopk}.

\begin{table}[t]
\centering\scriptsize\setlength{\tabcolsep}{3pt}
\caption{Matched architecture comparison on the repaired feature contract, canonical partition (fit rows 66,402 / inner validation 7,378 / test 8,198), identical features at every stage (age from stage 1, scene/context block from stage 4), seeds $\{0,1,2\}$, one policy for every architecture: train on the fit rows, keep the best-validation-NLL epoch (early stop after 25 epochs without improvement), report the 3-seed geometric-mean ensemble on test; the single-seed mean is given for comparison and is not the ensemble. Each architecture is shown at the configuration a validation sweep over learning rate ($10^{-3}$, $3{\cdot}10^{-4}$) and width (128 / [256,128] against 256 / [512,256]) selected on seed-0 validation top-1 at stages 4--6 (default: lr $10^{-3}$, narrow). $\Delta$: ensemble minus the selected MLP ensemble, exact paired McNemar $p$ on the 8,198 test records (descriptive; 48 comparisons). Cascades train all six stages jointly (sum of per-stage cross-entropies). Parameters are for one seed at that stage.}
\label{tab:architectures}
\vspace{4pt}
\begin{adjustbox}{max width=\linewidth}
\begin{tabular}{clrrrrrrrr}
\toprule
st. & architecture (selected config) & params & seed top-1 (mean) & ens.\ top-1 & top-3 & top-5 & top-10 & NLL & $\Delta_1$ / $\Delta_3$ vs MLP ($p$)\\
\midrule
1 & mlp (lr $3{\cdot}10^{-4}$, wide) & 182k & 20.44 & 20.57 & 38.39 & \textbf{48.32} & \textbf{63.06} & \textbf{3.213} & --\\
 & lstm (lr $3{\cdot}10^{-4}$, wide) & 585k & 20.38 & 20.53 & 38.22 & 48.08 & 62.75 & 3.216 & $-0.04$ (0.91) / $-0.17$ (0.42)\\
 & transformer (lr $3{\cdot}10^{-4}$, wide) & 1113k & 20.48 & 20.40 & 38.38 & 48.19 & 62.71 & 3.225 & $-0.17$ (0.45) / $-0.01$ (1)\\
 & cascaded\_mlp (lr $3{\cdot}10^{-4}$, wide) & 3683k & \textbf{20.55} & 20.58 & \textbf{38.45} & 48.17 & 62.91 & 3.215 & $+0.01$ (1) / {\bfseries\boldmath $+0.06$} (0.79)\\
 & cascaded\_transformer (default) & 1253k & 20.20 & \textbf{20.60} & 37.90 & 47.90 & 62.61 & 3.242 & {\bfseries\boldmath $+0.04$} (0.92) / $-0.49$ (0.034)\\
\addlinespace[1pt]
2 & mlp (lr $3{\cdot}10^{-4}$, wide) & 187k & \textbf{24.64} & 24.66 & 44.18 & 54.59 & \textbf{68.46} & 2.998 & --\\
 & lstm (lr $3{\cdot}10^{-4}$, wide) & 588k & 24.44 & 24.74 & 43.99 & 54.62 & 68.33 & 3.004 & $+0.07$ (0.73) / $-0.20$ (0.41)\\
 & transformer (lr $3{\cdot}10^{-4}$, wide) & 1117k & 24.36 & 24.54 & 43.94 & 54.51 & 68.25 & 3.013 & $-0.12$ (0.53) / $-0.24$ (0.32)\\
 & cascaded\_mlp (lr $3{\cdot}10^{-4}$, wide) & 3683k & \textbf{24.64} & \textbf{24.81} & \textbf{44.40} & \textbf{54.71} & 68.43 & \textbf{2.997} & {\bfseries\boldmath $+0.15$} (0.36) / {\bfseries\boldmath $+0.22$} (0.26)\\
 & cascaded\_transformer (default) & 1253k & 24.19 & 24.54 & 43.62 & 54.00 & 68.11 & 3.030 & $-0.12$ (0.61) / $-0.56$ (0.027)\\
\addlinespace[1pt]
3 & mlp (lr $3{\cdot}10^{-4}$, wide) & 193k & \textbf{27.82} & \textbf{28.03} & \textbf{48.95} & \textbf{59.60} & 72.93 & \textbf{2.821} & --\\
 & lstm (lr $3{\cdot}10^{-4}$, wide) & 592k & 27.51 & 27.53 & 48.87 & \textbf{59.60} & 72.98 & 2.828 & $-0.50$ (0.018) / {\bfseries\boldmath $-0.09$} (0.76)\\
 & transformer (lr $3{\cdot}10^{-4}$, wide) & 1121k & 27.48 & 27.88 & 48.66 & 59.36 & \textbf{73.16} & 2.840 & $-0.15$ (0.48) / $-0.29$ (0.22)\\
 & cascaded\_mlp (lr $3{\cdot}10^{-4}$, wide) & 3683k & 27.68 & 27.99 & 48.78 & 59.59 & 72.93 & 2.827 & {\bfseries\boldmath $-0.04$} (0.89) / $-0.17$ (0.45)\\
 & cascaded\_transformer (default) & 1253k & 26.77 & 27.26 & 48.23 & 59.22 & 72.82 & 2.864 & $-0.77$ (0.0014) / $-0.72$ (0.0044)\\
\addlinespace[1pt]
4 & mlp (lr $3{\cdot}10^{-4}$, wide) & 829k & \textbf{47.49} & \textbf{48.17} & \textbf{70.05} & \textbf{78.56} & \textbf{87.62} & \textbf{1.972} & --\\
 & lstm (lr $3{\cdot}10^{-4}$, wide) & 911k & 46.97 & 47.94 & 69.94 & 78.35 & 87.34 & 2.009 & $-0.23$ (0.44) / {\bfseries\boldmath $-0.11$} (0.72)\\
 & transformer (lr $3{\cdot}10^{-4}$, wide) & 1441k & 45.95 & 47.28 & 69.18 & 78.13 & 87.23 & 2.042 & $-0.89$ (0.0031) / $-0.88$ (0.0036)\\
 & cascaded\_mlp (lr $3{\cdot}10^{-4}$, wide) & 3683k & 47.46 & 48.06 & 69.52 & 78.26 & 87.40 & 1.992 & {\bfseries\boldmath $-0.11$} (0.69) / $-0.54$ (0.026)\\
 & cascaded\_transformer (default) & 1253k & 46.19 & 47.98 & 69.36 & 78.04 & 86.80 & 2.060 & $-0.20$ (0.54) / $-0.70$ (0.022)\\
\addlinespace[1pt]
5 & mlp (lr $3{\cdot}10^{-4}$, wide) & 869k & 51.48 & \textbf{52.48} & \textbf{74.69} & \textbf{82.58} & \textbf{89.99} & \textbf{1.813} & --\\
 & lstm (lr $3{\cdot}10^{-4}$, wide) & 932k & 49.82 & 50.95 & 73.36 & 81.24 & 89.14 & 1.889 & $-1.52$ (7.6e-05) / $-1.33$ (6.9e-05)\\
 & transformer (lr $3{\cdot}10^{-4}$, wide) & 1462k & 49.41 & 50.74 & 73.33 & 81.29 & 89.45 & 1.918 & $-1.73$ (9.6e-06) / $-1.35$ (6.4e-05)\\
 & cascaded\_mlp (lr $3{\cdot}10^{-4}$, wide) & 3683k & \textbf{51.62} & 52.33 & 73.70 & 81.82 & 89.66 & 1.833 & {\bfseries\boldmath $-0.15$} (0.64) / $-0.99$ (0.00024)\\
 & cascaded\_transformer (default) & 1253k & 49.91 & 52.07 & 73.79 & 81.57 & 89.17 & 1.901 & $-0.40$ (0.27) / {\bfseries\boldmath $-0.90$} (0.0056)\\
\addlinespace[1pt]
6 & mlp (lr $3{\cdot}10^{-4}$, wide) & 1214k & \textbf{57.32} & \textbf{58.45} & \textbf{78.60} & \textbf{85.59} & \textbf{91.99} & \textbf{1.591} & --\\
 & lstm (lr $3{\cdot}10^{-4}$, wide) & 1105k & 55.70 & 56.92 & 77.86 & 84.79 & 91.35 & 1.695 & $-1.54$ (4.7e-05) / $-0.74$ (0.021)\\
 & transformer (lr $3{\cdot}10^{-4}$, wide) & 1636k & 55.20 & 57.00 & 78.02 & 85.00 & 91.41 & 1.715 & $-1.45$ (0.00023) / $-0.59$ (0.072)\\
 & cascaded\_mlp (lr $3{\cdot}10^{-4}$, wide) & 3683k & 56.66 & 57.26 & 78.08 & 85.25 & 91.73 & 1.610 & {\bfseries\boldmath $-1.20$} (1.4e-05) / {\bfseries\boldmath $-0.52$} (0.03)\\
 & cascaded\_transformer (default) & 1253k & 55.02 & 56.75 & 77.46 & 84.35 & 91.03 & 1.700 & $-1.71$ (1.2e-06) / $-1.15$ (0.00022)\\
\addlinespace[1pt]
\bottomrule
\end{tabular}
\end{adjustbox}
\end{table}

\section{SEP: Serving, Prompt, Subsets and Per-Stage Results}
\label{app:llm}
This appendix details Staged-Evidence Prompting (SEP, \S\ref{sec:sep}), the large-language-model (LLM) arm. Every LLM number below is from the rerun on the repaired feature family (run \texttt{t3\_repaired}) with the identical prompts, decoding and subsets, except the \texttt{val820} selection runs, which were made before the repair and fixed the configuration.

\paragraph{Serving and decoding.} Qwen3-32B-AWQ \citep{qwen25} is served locally by vLLM \citep{vllm} with decoding pinned for reproducibility (Table~\ref{tab:decoding}). The 2,050-record checkpoint subset reproduces as an exact prefix of the full test run under greedy decoding.

\begin{table}[h]
\centering\footnotesize
\begin{tabular}{p{2.2cm}p{10.8cm}}
\toprule
model & Qwen3-32B-AWQ (4-bit AWQ, \texttt{awq\_marlin}) \\
serving & vLLM, max model length 8,192, prefix caching, eager execution, GPU memory fraction 0.88 \\
parallelism & tensor parallel 1, pipeline parallel 2 (tensor parallelism hung in the first forward pass on this host's PCIe-only GPU pair) \\
decoding & temperature 0 (greedy), top-$p$ 1.0, max 80 output tokens, seed 0, \texttt{chat\_template\_kwargs=\{enable\_thinking: false\}} \\
retries & transport failures only; a parse failure is a miss (a retry is a no-op under greedy decoding) \\
\bottomrule
\end{tabular}
\caption{Serving and decoding configuration of the LLM arm.}
\label{tab:decoding}
\end{table}

\paragraph{Subsets, metrics and selection.}\label{app:llmsubsets} Three evaluation subsets are derived from the split at load time, never read from a file, and identified by a SHA-256 over their sorted record identifiers plus an order hash, checked at load: \texttt{val820} (ten inner-validation records per protocol; every selection decision), \texttt{test2050} (25 test records per protocol; a checkpoint) and \texttt{test8198} (all test rows; the reported numbers). Two accuracies are defined: valid-only over parsed responses, and \emph{overall}, which counts an unparseable response as a miss. Overall is the primary metric because it is what a deployment sees; the measured invalid rate is 0.00\% at every stage, so the two coincide. \emph{Number of demonstrations.} Pooled over the six stages and three demonstration seeds on \texttt{val820} ($n=\selKn$ paired predictions), three-shot prompting changed top-1 / top-3 by $\selKthreeDOne$ / $\selKthreeDThree$ points against zero-shot ($p=\selKthreePOne$ / $\selKthreePThree$) and was accepted; five-shot changed them by $\selKfiveDOne$ / $\selKfiveDThree$ ($p=\selKfivePOne$ / $\selKfivePThree$) and was rejected because it harmed top-3 (\texttt{artifacts/llm/runs/t2\_run/t3\_decision.json}). \emph{Other variants}, explored on \texttt{val820} only and never read on test: a second fixed candidate ordering changed zero-shot top-1 by $+0.7$, $-1.5$, $-0.9$, $+3.1$, $+2.1$ and $+1.0$ points at stages 1--6 and cleared the acceptance threshold at stage 4 only; an alternative prefix was not carried into the test protocol; and thinking mode improved top-3 at five stages but ran at 13.8 to 66.2 times the wall-clock of the adopted configuration (mean 34.8) and was rejected on cost. The candidate order is fixed because shuffling it per query would destroy the server's prefix cache. Table~\ref{tab:llm} gives the arm at every stage with the seed spread of the $k=3$ configuration and the within-arm verdicts.

\begin{table}[t]
\centering
\scriptsize
\setlength{\tabcolsep}{3pt}
\caption{The prompted LLM arm (Qwen3-32B-AWQ, greedy, thinking disabled) on the repaired feature family, the same 8,198 test records and the same six stages as Table~\ref{tab:main}. Zero-shot and $k{=}3$ demonstrations (mean $\pm$ std over three demonstration seeds); the invalid rate is the share of unparseable responses, counted as misses. Right: the within-arm zero-shot$\to k{=}3$ dual gate on test (seed 0; accept if top-1 or top-3 improves at $p<0.025$ with no significant harm at $p<0.05$).}
\label{tab:llm}
\vspace{4pt}
\begin{tabular}{crrlrrl}
\toprule
stage & ZS top-1 & ZS top-3 & $k{=}3$ top-1 (mean $\pm$ sd) & $k{=}3$ top-3 & invalid\% & ZS$\to k{=}3$ test (seed 0)\\
\midrule
1 & \textbf{14.99} & \textbf{23.38} & 14.98 $\pm$ 0.73 & 22.77 & 0.00 & selected (top-1)\\
2 & 15.48 & 23.52 & {\bfseries\boldmath 15.93 $\pm$ 0.40} & \textbf{24.60} & 0.00 & rejected\\
3 & 15.52 & 24.41 & {\bfseries\boldmath 16.34 $\pm$ 0.66} & \textbf{26.26} & 0.00 & selected (top-3)\\
4 & 22.71 & 33.35 & {\bfseries\boldmath 25.74 $\pm$ 0.54} & \textbf{36.00} & 0.00 & selected (top-1, top-3)\\
5 & 24.10 & 34.00 & {\bfseries\boldmath 27.12 $\pm$ 0.69} & \textbf{36.76} & 0.00 & selected (top-1, top-3)\\
6 & 29.81 & 40.00 & {\bfseries\boldmath 33.29 $\pm$ 0.75} & \textbf{42.72} & 0.00 & selected (top-1, top-3)\\
\bottomrule
\end{tabular}
\end{table}

\paragraph{The prompt.} Figure~\ref{fig:prompt} gives the template. The fixed prefix is byte-identical for a whole run and placed first, so the server's prefix cache covers it; the $k$ demonstrations and the query record follow. Only the template is shown: a rendered prompt contains a patient record.

\begin{figure}[h]
\centering
\begin{minipage}{0.96\textwidth}
\hrule\vspace{3pt}
\scriptsize
\begin{verbatim}
You are an experienced EMS protocol classifier for a US NEMSIS 2024 patient care record.

A dispatcher or paramedic will act on a SHORTLIST, not on a single answer, so rank the
candidates: put the protocol you consider most likely first, then the next most likely,
and so on.

Information about a call arrives progressively. You are shown only what is documented by
the current point in the workflow. Fields that were not documented are shown explicitly
as [not recorded] or [not measured]; treat that as genuine absence of a measurement, and
do NOT assume a normal value in its place.

Answer with ONE compact JSON object and nothing else:
{"ranked":[id1,id2,id3,id4,id5,id6,id7,id8,id9,id10]}
- exactly 10 protocol IDs, most likely first
- IDs only, as integers, drawn from the candidate list below
- no duplicates, no names, no explanation, no markdown fence

HOW TO READ A RECORD
- Dispatch complaint: the call-taker's coded reason for the call.
- Complaint location / organ system: where the patient's complaint is, and which system.
- Symptoms and impressions: ICD-10-CM terms documented by the crew.
- Vitals are summarised over the call as mean+-sd; a single reading shows one number.
- GCS is 3-15 (15 = normal). AVPU is Alert / Verbal / Painful / Unresponsive.
- Pain is a 0-10 score. ETCO2 is end-tidal CO2 in mmHg. SpO2 is a percentage.

CANDIDATE PROTOCOLS (82 classes, id<TAB>name):
<82 lines, one per protocol, in one fixed ordering>
                             --- end of the cached prefix ---
<k demonstrations: a record rendered at the query's stage, then its gold ranked answer>
<the query record, rendered at the current stage>
\end{verbatim}
\vspace{-3pt}\hrule
\end{minipage}
\caption{The prompt template, reproduced from the builder. Everything above the marker is the fixed, cacheable prefix; the marker itself is not part of the prompt. Angle brackets stand for content that varies: the candidate block is 82 lines of \texttt{id<TAB>name}, and the demonstrations and query are records rendered by the same stage-aware code that produces the supervised arm's features. No rendered prompt is shown, because it would contain a patient record.}
\label{fig:prompt}
\end{figure}

\begin{figure}[t]
\centering
\includegraphics[width=\textwidth]{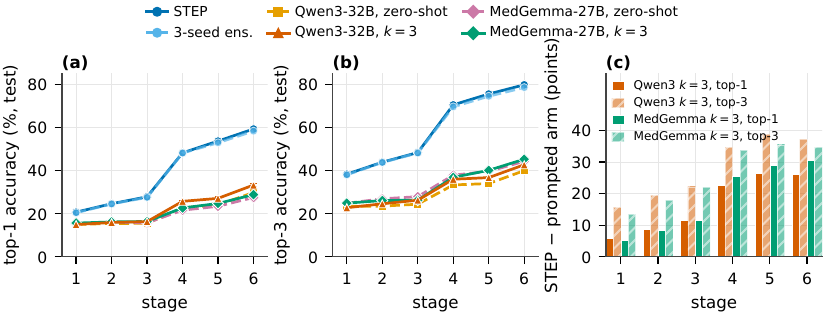}
\caption{SEP on the same 8,198 test records: (a) top-1 and (b) top-3 by stage for STEP, the ensemble, Qwen3-32B-AWQ and MedGemma-27B-text; (c) STEP minus each arm at $k{=}3$. Values in Tables~\ref{tab:stepllm}, \ref{tab:medgemma} and~\ref{tab:stepvsllm}.}
\label{fig:sep}
\end{figure}

\paragraph{Against STEP.} Figure~\ref{fig:sep} and Table~\ref{tab:stepllm} pair every prompted run with STEP on the same 8,198 test records. The prompted model reproduces the shape of the stage curve, including the jump at the scene stage, but the distance grows with the evidence: with three demonstrations it is \qwenGapOnedispatch{} top-1 points behind STEP at dispatch, \qwenGapOnevitals{} after vital signs and \qwenGapOneimpressions{} after impressions, and all \nllmfamily\ paired tests (four prompted runs, six stages, two metrics) favour STEP at $p<0.05/\nllmfamily$. Within the prompted arm, the seed-0 $k=3$ run is accepted over zero-shot at every stage but the second (Table~\ref{tab:llm}). A stage-6 record is \dimDimpressions{} mostly binary columns, which the supervised model consumes directly and the prompted model reads as a serialised field list, and it falls furthest behind where that list is longest. One hypothesis, which this study does not test, is that the widening gap reflects the evidence format rather than model scale; no scale sweep, fine-tuning or alternative serialisation was run. The last two columns of Table~\ref{tab:stepllm} give the gaps to the three-seed ensemble without pooling; against STEP every gap widens slightly, by the pool's own gain.

\begin{table}[h]
\centering
\scriptsize
\setlength{\tabcolsep}{3.5pt}
\caption{The prompted arm against STEP on the repaired family, same 8,198 test records, in percentage points. Each of the four prompted runs per stage (zero-shot; $k{=}3$ with demonstration seeds 0--2) is paired record for record with STEP and with the 3-seed ensemble; ``largest $p$'' is the largest exact McNemar $p$ over the eight STEP comparisons of the stage (all far below $0.05/24$).}
\label{tab:stepllm}
\vspace{4pt}
\begin{tabular}{crrrrrrr}
\toprule
& \multicolumn{2}{c}{zero-shot $-$ STEP} & \multicolumn{2}{c}{$k{=}3$ $-$ STEP (mean)} & & \multicolumn{2}{c}{top-1 gap to the ensemble}\\
\cmidrule(lr){2-3}\cmidrule(lr){4-5}\cmidrule(lr){7-8}
stage & top-1 & top-3 & top-1 & top-3 & largest $p$ & zero-shot & $k{=}3$\\
\midrule
1 & {\bfseries\boldmath $-5.73$} & {\bfseries\boldmath $-14.95$} & $-5.74$ & $-15.57$ & $6.9\!\times\!10^{-32}$ & {\bfseries\boldmath $-5.43$} & $-5.44$\\
2 & $-9.15$ & $-20.40$ & {\bfseries\boldmath $-8.70$} & {\bfseries\boldmath $-19.31$} & $1.5\!\times\!10^{-69}$ & $-9.06$ & {\bfseries\boldmath $-8.62$}\\
3 & $-12.33$ & $-23.98$ & {\bfseries\boldmath $-11.51$} & {\bfseries\boldmath $-22.13$} & $3.2\!\times\!10^{-103}$ & $-12.04$ & {\bfseries\boldmath $-11.22$}\\
4 & $-25.53$ & $-37.17$ & {\bfseries\boldmath $-22.50$} & {\bfseries\boldmath $-34.52$} & $5.2\!\times\!10^{-285}$ & $-25.45$ & {\bfseries\boldmath $-22.42$}\\
5 & $-29.60$ & $-41.56$ & {\bfseries\boldmath $-26.59$} & {\bfseries\boldmath $-38.80$} & $<\!10^{-300}$ & $-28.84$ & {\bfseries\boldmath $-25.82$}\\
6 & $-29.49$ & $-39.83$ & {\bfseries\boldmath $-26.01$} & {\bfseries\boldmath $-37.11$} & $<\!10^{-300}$ & $-28.53$ & {\bfseries\boldmath $-25.05$}\\
\bottomrule
\end{tabular}
\end{table}

\begin{figure}[h]
\centering
\includegraphics[width=\textwidth]{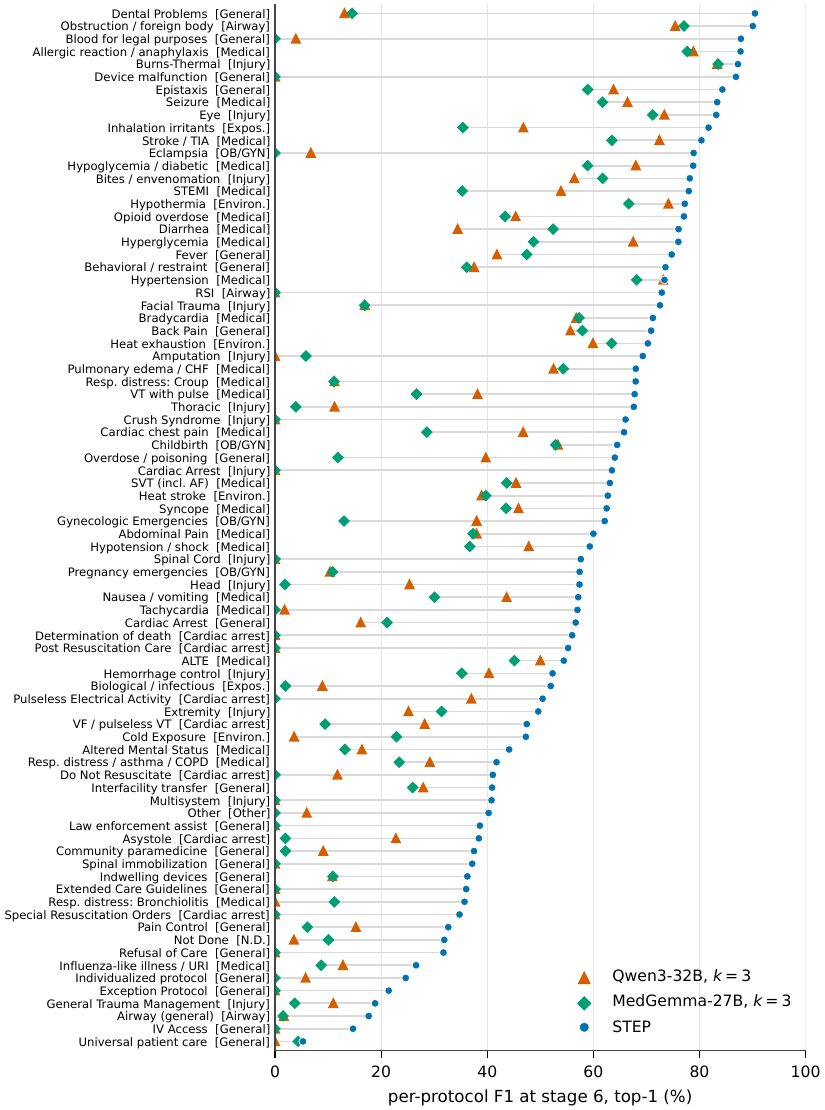}
\caption{Per-protocol stage-6 F1 of STEP, Qwen3-32B-AWQ and MedGemma-27B-text ($k{=}3$) on the balanced test split; values in Table~\ref{tab:perprotocol}.}
\label{fig:perprotf1}
\end{figure}

\begin{figure}[p]
\centering
\includegraphics[width=0.96\textwidth]{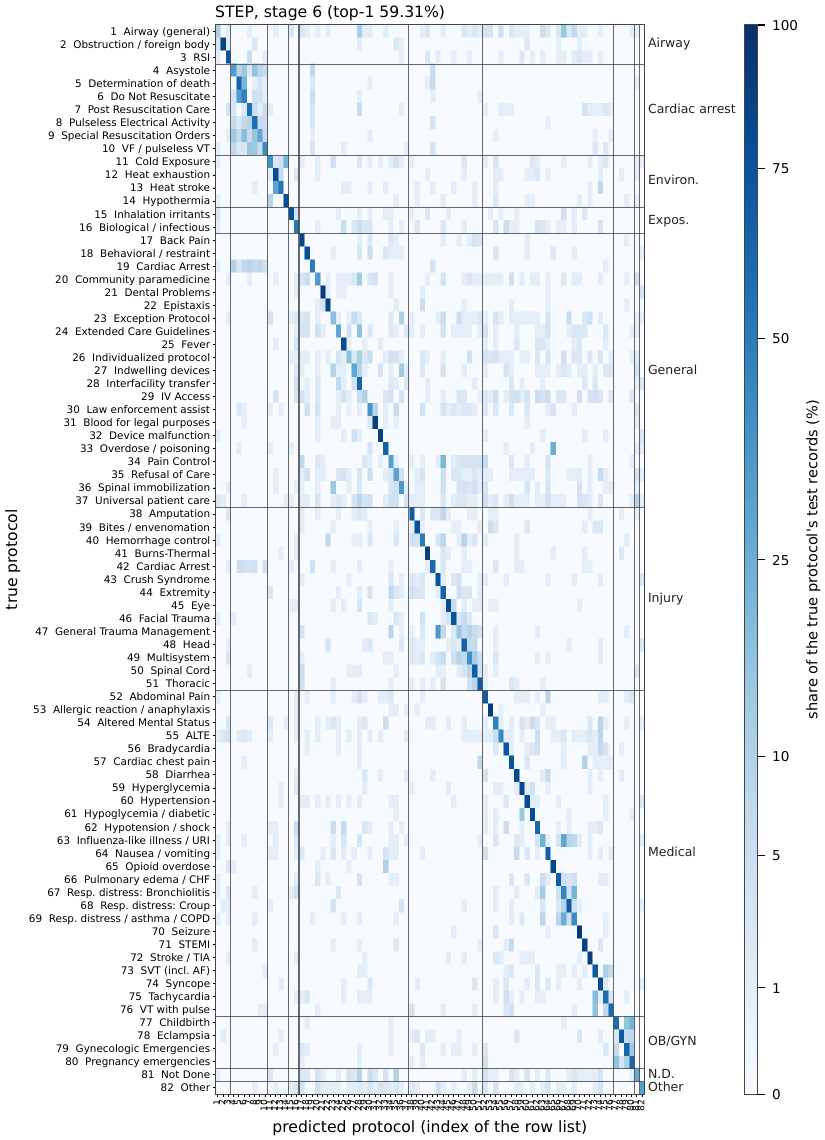}
\caption{Complete stage-6 confusion matrix of STEP over the 82 protocols on the balanced test split (rows: recorded protocol; columns: top-1 prediction).}
\label{fig:confstep}
\end{figure}

\begin{figure}[p]
\centering
\includegraphics[width=0.96\textwidth]{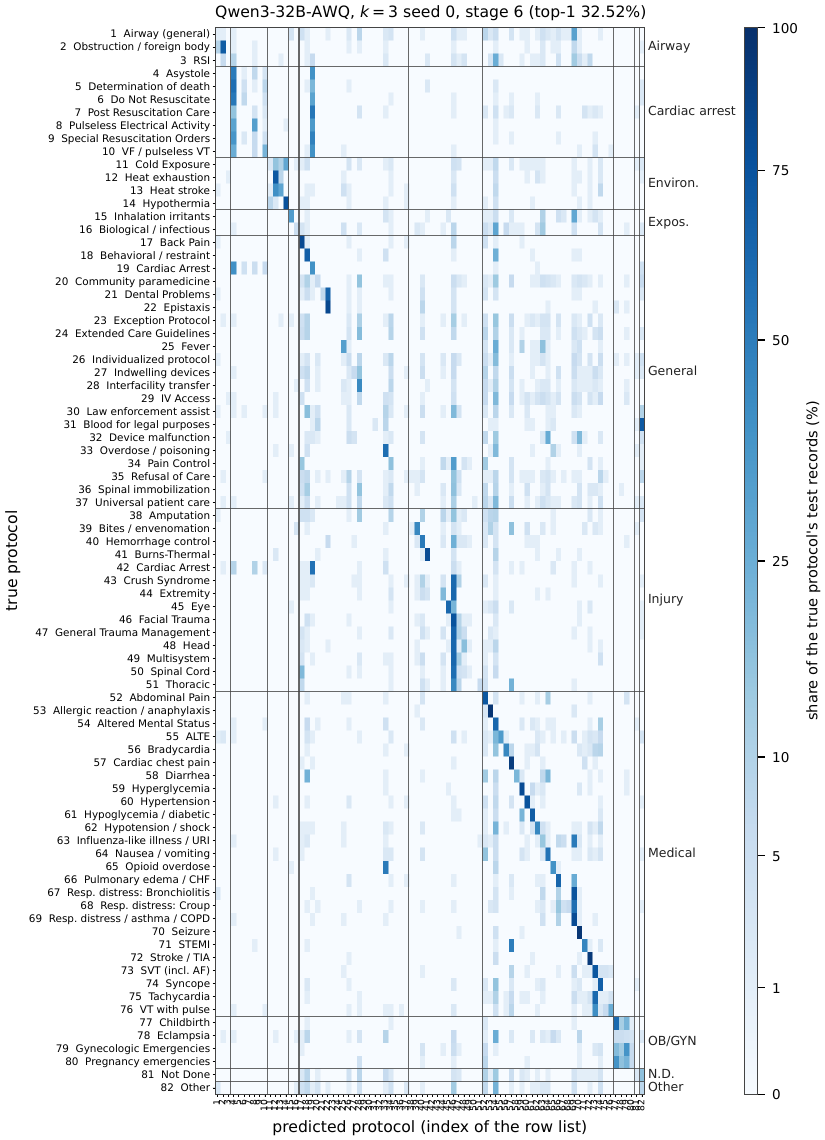}
\caption{As Figure~\ref{fig:confstep} for Qwen3-32B-AWQ ($k{=}3$, seed 0). Takeaway: mass collapses onto a few general and medical columns.}
\label{fig:confqwen3}
\end{figure}

\begin{figure}[p]
\centering
\includegraphics[width=0.96\textwidth]{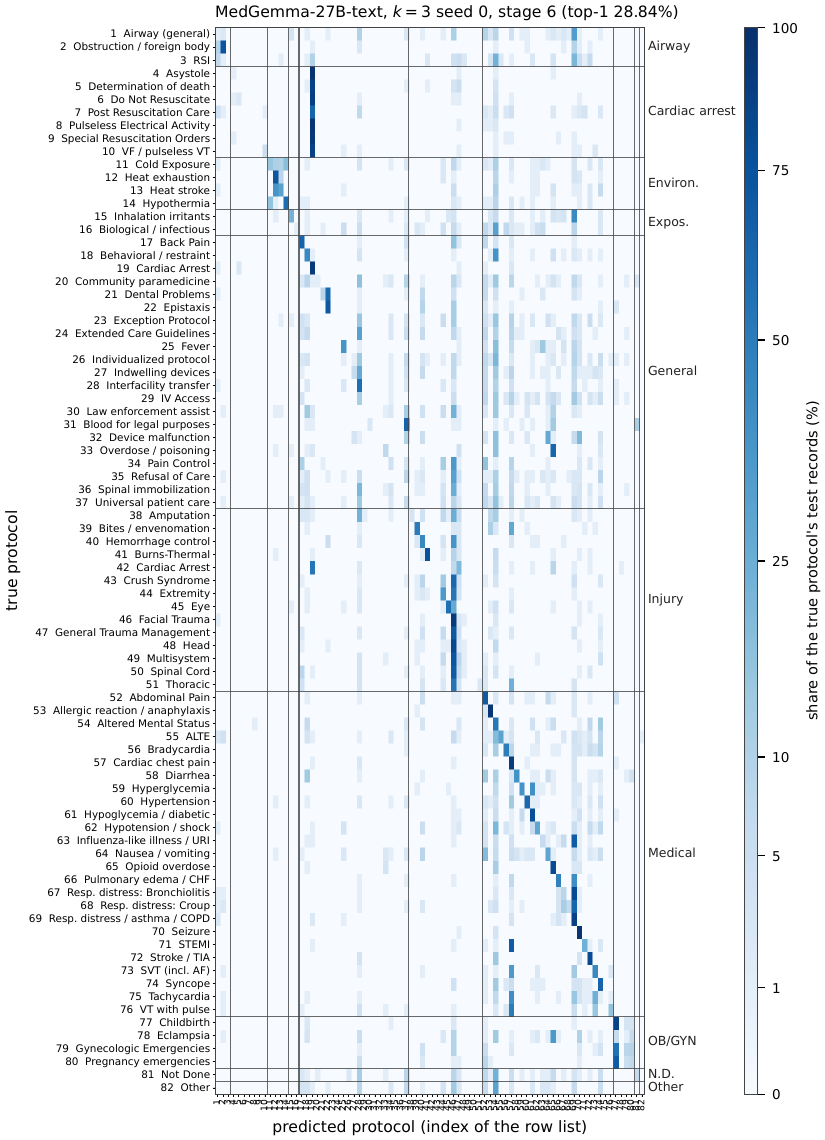}
\caption{As Figure~\ref{fig:confstep} for MedGemma-27B-text ($k{=}3$, seed 0). Takeaway: the same collapse as Qwen3, with a stronger pull towards the medical columns.}
\label{fig:confmedgemma}
\end{figure}

\paragraph{MedGemma-27B-text: protocol and caveats.}\label{app:llmprotocol}
The second prompted arm is MedGemma-27B-text \citep{medgemma2025}, a medically tuned instruction model, served locally by vLLM in bfloat16 (pipeline parallel over three A30 GPUs, maximum model length 4,096, greedy decoding, seed 0, at most 160 output tokens because its tokenizer splits digits individually, no thinking mode exists for it) under the identical staged-evidence prompts, subsets and metric code as Qwen3, with the record identifier, medications and procedures never rendered. It was run at Qwen3's selected configuration (the same prefix, candidate order, $k\in\{0,3\}$ and demonstration seed 0), fixed before any MedGemma test row was read, so it answers ``same prompt, different model'' and not ``MedGemma at its own best''. Every MedGemma row is matched to the same PcrKey in the Qwen3 run and in the supervised arm; an unmatched row is an error, not a warning. Table~\ref{tab:medgemma} gives the paired results at every stage; Table~\ref{tab:stepvsllm} the three models at depths 1, 3, 5 and 10; Figure~\ref{fig:conffamily} and Table~\ref{tab:prf} the family-level confusions and macro precision, recall and F1; Figures~\ref{fig:perprotf1} and~\ref{fig:confstep}--\ref{fig:confmedgemma} and Table~\ref{tab:perprotocol} the per-protocol view at stage 6. Five caveats qualify the arm. (i) \emph{Precision differs between the arms}: Qwen3 ran as a 4-bit AWQ checkpoint, MedGemma in bf16, because no usable quantised checkpoint of MedGemma-27B-text existed; bf16 is the more favourable setting, so it cannot explain a MedGemma deficit, but it is a real asymmetry. (ii) \emph{Sliding-window asymmetry}: MedGemma's architecture uses a 1,024-token local attention window on five of every six layers, and the fixed candidate list alone is 1,290 tokens under its tokenizer, so on those layers the model attends to the query record without the candidate list in its local window, whereas Qwen3 used full attention throughout; this is reported, not compensated for, because changing the prompt would break the frozen-prompt rule. (iii) \emph{Seed 0 only}: Qwen3's $k{=}3$ result is a mean over three demonstration seeds with a stage-6 top-1 spread of \qwenSdOneimpressions{} \%, MedGemma has one seed, so the paired MedGemma-vs-Qwen3 tests compare seed 0 with seed 0 and the cross-arm differences are of the order of Qwen3's own seed spread. (iv) \emph{The prompt configuration was selected on Qwen3}, not on MedGemma, so the arm may understate what MedGemma would reach with its own selection. (v) \emph{Reproducibility is not exact across batch composition}: on a 2,050-record checkpoint subset, 17.0\% of rows received a different ranked list when scored standalone and inside the full 8,198-row pass (batched floating-point reduction order), with single-digit row counts changing a top-1 or top-3 hit, whereas the Qwen3 arm reproduced exactly; the invalid-response rate is at most \medgInvalidMax\% (Table~\ref{tab:medgemma}). The reported arms comprise \llmCallsTotal{} prompted calls (\llmCallsQwen{} for Qwen3 over four conditions per stage, \llmCallsMedg{} for MedGemma over two) and \llmPromptTokensMillions{} million prompt tokens in all.

\begin{table}[t]
\centering\scriptsize\setlength{\tabcolsep}{2.6pt}
\caption{MedGemma-27B-text (bf16, seed 0, zero-shot and $k{=}3$, Qwen3's selected prompt configuration) on the same 8,198 test records of the balanced cohort, paired record for record with Qwen3-32B-AWQ seed 0 of the repaired-family rerun and with STEP (Table~\ref{tab:main}). $\Delta$: MedGemma minus the comparator in points, exact two-sided McNemar $p$; $^{*}$ marks the comparisons that clear the Bonferroni level of family C ($0.05/\nFamilyC$: the 24 MedGemma-vs-STEP and 24 MedGemma-vs-Qwen3 tests). inv.\%: unparseable responses, counted as misses. Right: the within-arm zero-shot$\to k{=}3$ dual gate on test (accept if top-1 or top-3 improves at $p<0.025$ with no significant harm at $p<0.05$), applied post hoc for symmetry with Table~\ref{tab:llm}; no MedGemma configuration was selected on validation.}
\label{tab:medgemma}
\vspace{4pt}
\begin{adjustbox}{max width=\linewidth}
\begin{tabular}{clrrrlrlrr}
\toprule
stage & config & \multicolumn{2}{c}{MedGemma top-1 / top-3} & \multicolumn{2}{c}{$-$ Qwen3 seed 0 (p)} & \multicolumn{2}{c}{$-$ STEP (p)} & inv.\% & ZS$\to k{=}3$\\
\cmidrule(lr){3-4}\cmidrule(lr){5-6}\cmidrule(lr){7-8}
 & & top-1 & top-3 & $\Delta_1$ & $\Delta_3$ & $\Delta_1$ & $\Delta_3$ & & test\\
\midrule
1 & zero-shot & 15.30 & 24.85 & $+0.30$ ($0.18$) & $+1.46$ ($4.7\!\times\!10^{-5}$)$^{*}$ & $-5.43$ ($1.8\!\times\!10^{-40}$)$^{*}$ & $-13.49$ ($7.7\!\times\!10^{-143}$)$^{*}$ & \textbf{0.00} & \\
 & $k{=}3$ & \textbf{15.64} & \textbf{24.90} & $-0.01$ ($1$) & $+1.22$ ($7.5\!\times\!10^{-6}$)$^{*}$ & {\bfseries\boldmath $-5.09$} ($6.6\!\times\!10^{-34}$)$^{*}$ & {\bfseries\boldmath $-13.44$} ($1.6\!\times\!10^{-140}$)$^{*}$ & \textbf{0.00} & rejected\\
\addlinespace[1pt]
2 & zero-shot & 16.08 & \textbf{26.92} & $+0.60$ ($0.022$) & $+3.40$ ($1.9\!\times\!10^{-19}$)$^{*}$ & $-8.55$ ($7.1\!\times\!10^{-77}$)$^{*}$ & {\bfseries\boldmath $-16.99$} ($3.8\!\times\!10^{-200}$)$^{*}$ & \textbf{0.00} & \\
 & $k{=}3$ & \textbf{16.19} & 26.12 & $+0.37$ ($0.099$) & $+2.01$ ($2.8\!\times\!10^{-13}$)$^{*}$ & {\bfseries\boldmath $-8.44$} ($1.5\!\times\!10^{-72}$)$^{*}$ & $-17.80$ ($1.8\!\times\!10^{-216}$)$^{*}$ & \textbf{0.00} & rejected\\
\addlinespace[1pt]
3 & zero-shot & 15.76 & \textbf{27.91} & $+0.24$ ($0.4$) & $+3.50$ ($4.8\!\times\!10^{-20}$)$^{*}$ & $-12.09$ ($1.2\!\times\!10^{-125}$)$^{*}$ & {\bfseries\boldmath $-20.48$} ($5.6\!\times\!10^{-247}$)$^{*}$ & \textbf{0.00} & \\
 & $k{=}3$ & \textbf{16.39} & 26.30 & $+0.81$ ($0.001$)$^{*}$ & $-0.07$ ($0.83$) & {\bfseries\boldmath $-11.45$} ($9.1\!\times\!10^{-116}$)$^{*}$ & $-22.09$ ($2.0\!\times\!10^{-289}$)$^{*}$ & \textbf{0.00} & rejected\\
\addlinespace[1pt]
4 & zero-shot & 21.63 & \textbf{37.89} & $-1.09$ ($0.002$) & $+4.54$ ($3.9\!\times\!10^{-28}$)$^{*}$ & $-26.62$ ($<10^{-300}$)$^{*}$ & {\bfseries\boldmath $-32.63$} ($<10^{-300}$)$^{*}$ & \textbf{0.00} & \\
 & $k{=}3$ & \textbf{22.66} & 36.87 & $-2.56$ ($1.9\!\times\!10^{-16}$)$^{*}$ & $+1.09$ ($0.004$) & {\bfseries\boldmath $-25.58$} ($<10^{-300}$)$^{*}$ & $-33.64$ ($<10^{-300}$)$^{*}$ & \textbf{0.00} & rejected\\
\addlinespace[1pt]
5 & zero-shot & 23.43 & 39.66 & $-0.67$ ($0.094$) & $+5.66$ ($2.0\!\times\!10^{-41}$)$^{*}$ & $-30.28$ ($<10^{-300}$)$^{*}$ & $-35.90$ ($<10^{-300}$)$^{*}$ & \textbf{0.00} & \\
 & $k{=}3$ & \textbf{24.76} & \textbf{40.14} & $-1.61$ ($9.3\!\times\!10^{-6}$)$^{*}$ & $+3.05$ ($2.6\!\times\!10^{-15}$)$^{*}$ & {\bfseries\boldmath $-28.95$} ($<10^{-300}$)$^{*}$ & {\bfseries\boldmath $-35.41$} ($<10^{-300}$)$^{*}$ & 0.21 & selected (top-1)\\
\addlinespace[1pt]
6 & zero-shot & 27.52 & 44.30 & $-2.29$ ($2.1\!\times\!10^{-8}$)$^{*}$ & $+4.31$ ($3.5\!\times\!10^{-24}$)$^{*}$ & $-31.79$ ($<10^{-300}$)$^{*}$ & $-35.52$ ($<10^{-300}$)$^{*}$ & \textbf{0.00} & \\
 & $k{=}3$ & \textbf{28.84} & \textbf{45.28} & $-3.68$ ($2.2\!\times\!10^{-21}$)$^{*}$ & $+1.76$ ($1.1\!\times\!10^{-5}$)$^{*}$ & {\bfseries\boldmath $-30.47$} ($<10^{-300}$)$^{*}$ & {\bfseries\boldmath $-34.55$} ($<10^{-300}$)$^{*}$ & 0.13 & selected (top-1, top-3)\\
\bottomrule
\end{tabular}
\end{adjustbox}
\end{table}

\begin{table}[t]
\centering\footnotesize\setlength{\tabcolsep}{3pt}\renewcommand{\arraystretch}{1.02}
\caption{STEP against the two prompted arms on the same 8,198 test records of the balanced 82-protocol cohort at depths 1, 3, 5 and 10 (test accuracy, \%; best per stage and depth in bold). STEP is the repaired-family, clean-boundary model of Table~\ref{tab:main}. Both language models were served locally with greedy decoding under the staged-evidence prompt of App.~\ref{app:llm}, on the repaired feature family: Qwen3-32B-AWQ (4-bit, thinking disabled; run \texttt{t3\_repaired}; $k{=}3$ is the mean $\pm$ sd over demonstration seeds 0--2) and MedGemma-27B-text (bf16, seed 0 only; App.~\ref{app:llmprotocol}). Top-5 and top-10 never gated any decision.}
\label{tab:stepvsllm}
\vspace{4pt}
\begin{adjustbox}{max width=\linewidth}
\begin{tabular}{llrrrr}
\toprule
stage & model & top-1 & top-3 & top-5 & top-10\\
\midrule
\multirow{5}{*}{1 dispatch} & STEP (Table~\ref{tab:main}) & \textbf{20.72} & \textbf{38.34} & \textbf{48.29} & \textbf{62.98}\\
 & Qwen3-32B, zero-shot & 14.99 & 23.38 & 28.78 & 36.28\\
 & Qwen3-32B, $k{=}3$, 3 seeds & 14.98\,{\scriptsize$\pm$0.73} & 22.77\,{\scriptsize$\pm$1.02} & 28.28\,{\scriptsize$\pm$1.54} & 35.01\,{\scriptsize$\pm$1.19}\\
 & MedGemma-27B, zero-shot & 15.30 & 24.85 & 30.92 & 40.74\\
 & MedGemma-27B, $k{=}3$, seed 0 & 15.64 & 24.90 & 31.25 & 40.52\\
\midrule
\multirow{5}{*}{2 +location} & STEP (Table~\ref{tab:main}) & \textbf{24.63} & \textbf{43.91} & \textbf{54.53} & \textbf{68.04}\\
 & Qwen3-32B, zero-shot & 15.48 & 23.52 & 28.92 & 34.76\\
 & Qwen3-32B, $k{=}3$, 3 seeds & 15.93\,{\scriptsize$\pm$0.40} & 24.60\,{\scriptsize$\pm$0.69} & 29.87\,{\scriptsize$\pm$1.00} & 35.97\,{\scriptsize$\pm$0.50}\\
 & MedGemma-27B, zero-shot & 16.08 & 26.92 & 33.24 & 42.64\\
 & MedGemma-27B, $k{=}3$, seed 0 & 16.19 & 26.12 & 31.28 & 39.74\\
\midrule
\multirow{5}{*}{3 +organ system} & STEP (Table~\ref{tab:main}) & \textbf{27.85} & \textbf{48.39} & \textbf{59.40} & \textbf{72.90}\\
 & Qwen3-32B, zero-shot & 15.52 & 24.41 & 30.25 & 37.53\\
 & Qwen3-32B, $k{=}3$, 3 seeds & 16.34\,{\scriptsize$\pm$0.66} & 26.26\,{\scriptsize$\pm$0.48} & 31.91\,{\scriptsize$\pm$0.46} & 38.72\,{\scriptsize$\pm$0.90}\\
 & MedGemma-27B, zero-shot & 15.76 & 27.91 & 34.39 & 43.86\\
 & MedGemma-27B, $k{=}3$, seed 0 & 16.39 & 26.30 & 31.79 & 40.72\\
\midrule
\multirow{5}{*}{4 +symptoms, scene} & STEP (Table~\ref{tab:main}) & \textbf{48.24} & \textbf{70.52} & \textbf{78.89} & \textbf{87.70}\\
 & Qwen3-32B, zero-shot & 22.71 & 33.35 & 37.90 & 43.71\\
 & Qwen3-32B, $k{=}3$, 3 seeds & 25.74\,{\scriptsize$\pm$0.54} & 36.00\,{\scriptsize$\pm$0.48} & 40.25\,{\scriptsize$\pm$0.53} & 44.81\,{\scriptsize$\pm$0.59}\\
 & MedGemma-27B, zero-shot & 21.63 & 37.89 & 44.36 & 52.61\\
 & MedGemma-27B, $k{=}3$, seed 0 & 22.66 & 36.87 & 42.45 & 50.66\\
\midrule
\multirow{5}{*}{5 +vitals} & STEP (Table~\ref{tab:main}) & \textbf{53.71} & \textbf{75.56} & \textbf{82.98} & \textbf{90.36}\\
 & Qwen3-32B, zero-shot & 24.10 & 34.00 & 37.89 & 43.71\\
 & Qwen3-32B, $k{=}3$, 3 seeds & 27.12\,{\scriptsize$\pm$0.69} & 36.76\,{\scriptsize$\pm$0.55} & 40.66\,{\scriptsize$\pm$0.40} & 45.23\,{\scriptsize$\pm$1.00}\\
 & MedGemma-27B, zero-shot & 23.43 & 39.66 & 45.54 & 54.28\\
 & MedGemma-27B, $k{=}3$, seed 0 & 24.76 & 40.14 & 45.29 & 53.24\\
\midrule
\multirow{5}{*}{6 +impressions} & STEP (Table~\ref{tab:main}) & \textbf{59.31} & \textbf{79.82} & \textbf{86.36} & \textbf{92.45}\\
 & Qwen3-32B, zero-shot & 29.81 & 40.00 & 44.11 & 49.09\\
 & Qwen3-32B, $k{=}3$, 3 seeds & 33.29\,{\scriptsize$\pm$0.75} & 42.72\,{\scriptsize$\pm$0.70} & 45.49\,{\scriptsize$\pm$1.07} & 49.10\,{\scriptsize$\pm$0.58}\\
 & MedGemma-27B, zero-shot & 27.52 & 44.30 & 50.63 & 59.21\\
 & MedGemma-27B, $k{=}3$, seed 0 & 28.84 & 45.28 & 50.10 & 56.77\\
\bottomrule
\end{tabular}
\end{adjustbox}
\end{table}

\paragraph{Serving cost.} STEP's latency is measured on the trained, verified models (Table~\ref{tab:e3cost}; App.~\ref{app:steprobust}): on one CPU thread a stage-6 request runs \onMlpsPool{} MLPs (stages 1--5, the joint stage-6 model and seven experts, three seeds each; \onParamsPoolM\,M parameters) and the pool in \onLatStepCold~ms (95th percentile \onLatStepColdP~ms), or \onLatStepIncr~ms when the posteriors from the encounter's earlier refreshes are cached; the gate and pool alone take \onLatStepPool~ms and the three-seed stage-6 ensemble \onLatEns~ms. This replaces an earlier shape-level microbenchmark with random weights. For the prompted model, the repaired stage-6 $k{=}3$ test runs take \costLlmRunMs~ms per record as elapsed wall-clock over the 8,198 records at request concurrency 32: an \emph{amortised throughput}, not a per-request latency. The matched quantity was measured separately (\texttt{artifacts/paper/review/a6\_latency.json}): with the same server and the same $k{=}3$ prompts, 200 stage-6 test records sent one at a time take \costLlmSeqS~s per request on average (median \costLlmSeqMedS~s, 90th percentile \costLlmSeqPninetyS~s; \costLlmPromptTok{} prompt and \costLlmComplTok{} completion tokens on average), and the same 200 prompts at concurrency 32 complete in \costLlmBatchS~s, i.e.\ \costLlmAmortMs~ms per record amortised. On these measurements a cold STEP request is about \onSpeedupSeq{} times faster than a sequential prompted request and about \onSpeedupAmort{} times faster than the prompted model's best amortised rate. Both keep the record on local hardware. No language-model output feeds STEP or DPCS.

\scriptsize\setlength{\tabcolsep}{3pt}
\begin{longtable}{rlrrrrrrr}
\caption{Per-protocol stage-6 results on the balanced test split (100 records for 81 protocols, 98 for one), in \%: recall (R), precision (P) and F1 of the top-1 prediction for STEP, and top-1 recall and F1 of the two prompted arms ($k{=}3$, seed 0, repaired family). The index is the row order of the confusion heatmaps (Figs.~\ref{fig:confstep}--\ref{fig:confmedgemma}); protocols are grouped by coarse family.}\label{tab:perprotocol}\\
\toprule
\# & protocol (family) & \multicolumn{3}{c}{STEP} & \multicolumn{2}{c}{Qwen3 $k{=}3$} & \multicolumn{2}{c}{MedGemma $k{=}3$}\\
\cmidrule(lr){3-5}\cmidrule(lr){6-7}\cmidrule(lr){8-9}
 & & R & P & F1 & R & F1 & R & F1\\
\midrule
\endfirsthead
\toprule
\# & protocol (family) & \multicolumn{3}{c}{STEP} & \multicolumn{2}{c}{Qwen3 $k{=}3$} & \multicolumn{2}{c}{MedGemma $k{=}3$}\\
\cmidrule(lr){3-5}\cmidrule(lr){6-7}\cmidrule(lr){8-9}
 & & R & P & F1 & R & F1 & R & F1\\
\midrule
\endhead
\midrule\multicolumn{9}{r}{\emph{continued}}\\
\endfoot
\bottomrule
\endlastfoot
1 & Airway & \textbf{12} & 33 & \textbf{18} & 1 & 2 & 1 & 1\\
2 & Airway-Obstruction/\allowbreak Foreign Body & \textbf{86} & 95 & \textbf{90} & 69 & 75 & 74 & 77\\
3 & Airway-Rapid Sequence Induction (RSI-Paralytic) & \textbf{74} & 72 & \textbf{73} & 0 & 0 & 0 & 0\\
\addlinespace[2pt]
4 & Cardiac Arrest-Asystole & 34 & 44 & \textbf{38} & \textbf{50} & 23 & 1 & 2\\
5 & Cardiac Arrest-Determination of Death /\allowbreak  Withholding Resuscitative Efforts & \textbf{63} & 50 & \textbf{56} & 0 & 0 & 0 & 0\\
6 & Cardiac Arrest-Do Not Resuscitate & \textbf{47} & 36 & \textbf{41} & 7 & 12 & 0 & 0\\
7 & Cardiac Arrest-Post Resuscitation Care & \textbf{58} & 53 & \textbf{55} & 0 & 0 & 0 & 0\\
8 & Cardiac Arrest-Pulseless Electrical Activity & \textbf{58} & 45 & \textbf{50} & 32 & 37 & 0 & 0\\
9 & Cardiac Arrest-Special Resuscitation Orders & \textbf{29} & 44 & \textbf{35} & 0 & 0 & 0 & 0\\
10 & Cardiac Arrest-Ventricular Fibrillation/\allowbreak  Pulseless Ventricular Tachycardia & \textbf{42} & 55 & \textbf{47} & 22 & 28 & 5 & 9\\
\addlinespace[2pt]
11 & Environmental-Cold Exposure & \textbf{39} & 60 & \textbf{47} & 2 & 4 & 15 & 23\\
12 & Environmental-Heat Exposure/\allowbreak Exhaustion & \textbf{78} & 64 & \textbf{70} & 71 & 60 & 72 & 63\\
13 & Environmental-Heat Stroke/\allowbreak Hyperthermia & \textbf{53} & 77 & \textbf{63} & 29 & 39 & 30 & 40\\
14 & Environmental-Hypothermia & \textbf{78} & 76 & \textbf{77} & 76 & 74 & 59 & 67\\
\addlinespace[2pt]
15 & Exposure-Airway/\allowbreak Inhalation Irritants & \textbf{76} & 88 & \textbf{82} & 33 & 47 & 23 & 35\\
16 & Exposure-Biological/\allowbreak Infectious & \textbf{53} & 51 & \textbf{52} & 5 & 9 & 1 & 2\\
\addlinespace[2pt]
17 & General-Back Pain & \textbf{84} & 61 & \textbf{71} & 81 & 56 & 64 & 58\\
18 & General-Behavioral/\allowbreak Patient Restraint & \textbf{78} & 70 & \textbf{74} & 67 & 38 & 43 & 36\\
19 & General-Cardiac Arrest & 51 & 64 & \textbf{57} & 39 & 16 & \textbf{93} & 21\\
20 & General-Community Paramedicine /\allowbreak  Mobile Integrated Healthcare & \textbf{33} & 43 & \textbf{38} & 6 & 9 & 1 & 2\\
21 & General-Dental Problems & \textbf{90} & 91 & \textbf{90} & 7 & 13 & 8 & 15\\
22 & General-Epistaxis & \textbf{86} & 83 & \textbf{84} & 82 & 64 & 71 & 59\\
23 & General-Exception Protocol & \textbf{18} & 26 & \textbf{21} & 0 & 0 & 0 & 0\\
24 & General-Extended Care Guidelines & \textbf{29} & 48 & \textbf{36} & 0 & 0 & 0 & 0\\
25 & General-Fever & \textbf{80} & 70 & \textbf{75} & 32 & 42 & 37 & 47\\
26 & General-Individualized Patient Protocol & \textbf{17} & 45 & \textbf{25} & 5 & 6 & 0 & 0\\
27 & General-Indwelling Medical Devices/\allowbreak Equipment & \textbf{29} & 48 & \textbf{36} & 6 & 11 & 6 & 11\\
28 & General-Interfacility Transfers & \textbf{63} & 30 & \textbf{41} & 43 & 28 & 58 & 26\\
29 & General-IV Access & \textbf{10} & 28 & \textbf{15} & 0 & 0 & 0 & 0\\
30 & General-Law Enforcement - Assist with Law Enforcement Activity & \textbf{39} & 38 & \textbf{39} & 0 & 0 & 0 & 0\\
31 & General-Law Enforcement - Blood for Legal Purposes & \textbf{90} & 86 & \textbf{88} & 2 & 4 & 0 & 0\\
32 & General-Medical Device Malfunction & \textbf{86} & 88 & \textbf{87} & 0 & 0 & 0 & 0\\
33 & General-Overdose/\allowbreak Poisoning/\allowbreak Toxic Ingestion & \textbf{65} & 63 & \textbf{64} & 58 & 40 & 7 & 12\\
34 & General-Pain Control & \textbf{32} & 33 & \textbf{33} & 17 & 15 & 4 & 6\\
35 & General-Refusal of Care & \textbf{30} & 34 & \textbf{32} & 0 & 0 & 0 & 0\\
36 & General-Spinal Immobilization/\allowbreak Clearance & \textbf{34} & 41 & \textbf{37} & 0 & 0 & 0 & 0\\
37 & General-Universal Patient Care/\allowbreak  Initial Patient Contact & 3 & 21 & \textbf{5} & 0 & 0 & \textbf{6} & 4\\
\addlinespace[2pt]
38 & Injury-Amputation & \textbf{70} & 69 & \textbf{69} & 0 & 0 & 3 & 6\\
39 & Injury-Bites and Envenomations-Land & \textbf{77} & 79 & \textbf{78} & 44 & 56 & 50 & 62\\
40 & Injury-Bleeding/\allowbreak  Hemorrhage Control & \textbf{51} & 54 & \textbf{52} & \textbf{51} & 40 & 44 & 35\\
41 & Injury-Burns-Thermal & \textbf{89} & 86 & \textbf{87} & 80 & 83 & 76 & 84\\
42 & Injury-Cardiac Arrest & \textbf{60} & 67 & \textbf{63} & 0 & 0 & 0 & 0\\
43 & Injury-Crush Syndrome & \textbf{75} & 59 & \textbf{66} & 0 & 0 & 0 & 0\\
44 & Injury-Extremity & \textbf{65} & 40 & \textbf{50} & 20 & 25 & 35 & 31\\
45 & Injury-Eye & \textbf{79} & 88 & \textbf{83} & 62 & 73 & 58 & 71\\
46 & Injury-Facial Trauma & 74 & 71 & \textbf{73} & 75 & 17 & \textbf{90} & 17\\
47 & Injury-General Trauma Management & \textbf{15} & 25 & \textbf{19} & 10 & 11 & 4 & 4\\
48 & Injury-Head & \textbf{66} & 51 & \textbf{57} & 18 & 25 & 1 & 2\\
49 & Injury-Multisystem & \textbf{40} & 42 & \textbf{41} & 0 & 0 & 0 & 0\\
50 & Injury-Spinal Cord & \textbf{68} & 50 & \textbf{58} & 0 & 0 & 0 & 0\\
51 & Injury-Thoracic & \textbf{71} & 65 & \textbf{68} & 6 & 11 & 2 & 4\\
\addlinespace[2pt]
52 & Medical-Abdominal Pain & 69 & 53 & \textbf{60} & \textbf{72} & 38 & 70 & 37\\
53 & Medical-Allergic Reaction/\allowbreak Anaphylaxis & 93 & 83 & \textbf{88} & \textbf{95} & 79 & 94 & 78\\
54 & Medical-Altered Mental Status & 47 & 42 & \textbf{44} & \textbf{62} & 16 & 52 & 13\\
55 & Medical-Apparent Life Threatening Event (ALTE) & \textbf{43} & 74 & \textbf{54} & 34 & 50 & 30 & 45\\
56 & Medical-Bradycardia & \textbf{73} & 70 & \textbf{71} & 44 & 57 & 49 & 57\\
57 & Medical-Cardiac Chest Pain & 74 & 59 & \textbf{66} & 90 & 47 & \textbf{91} & 29\\
58 & Medical-Diarrhea & \textbf{81} & 72 & \textbf{76} & 21 & 34 & 38 & 52\\
59 & Medical-Hyperglycemia & \textbf{84} & 69 & \textbf{76} & 80 & 68 & 38 & 49\\
60 & Medical-Hypertension & \textbf{80} & 68 & \textbf{73} & 75 & \textbf{73} & 61 & 68\\
61 & Medical-Hypoglycemia/\allowbreak Diabetic Emergency & \textbf{80} & 78 & \textbf{79} & 68 & 68 & 76 & 59\\
62 & Medical-Hypotension/\allowbreak Shock (Non-Trauma) & \textbf{62} & 57 & \textbf{59} & 44 & 48 & 29 & 37\\
63 & Medical-Influenza-Like Illness/\allowbreak  Upper Respiratory Infection & \textbf{23} & 32 & \textbf{27} & 15 & 13 & 6 & 9\\
64 & Medical-Nausea/\allowbreak Vomiting & \textbf{66} & 50 & \textbf{57} & 55 & 44 & 29 & 30\\
65 & Medical-Opioid Poisoning/\allowbreak Overdose & \textbf{84} & 71 & \textbf{77} & 39 & 45 & 80 & 43\\
66 & Medical-Pulmonary Edema/\allowbreak CHF & \textbf{68} & 68 & \textbf{68} & 63 & 52 & 47 & 54\\
67 & Medical-Respiratory Distress-Bronchiolitis & \textbf{45} & 30 & \textbf{36} & 0 & 0 & 8 & 11\\
68 & Medical-Respiratory Distress-Croup & \textbf{69} & 67 & \textbf{68} & 6 & 11 & 6 & 11\\
69 & Medical-Respiratory Distress/\allowbreak Asthma/\allowbreak COPD/\allowbreak Reactive Airway & 43 & 41 & \textbf{42} & 80 & 29 & \textbf{87} & 23\\
70 & Medical-Seizure & 95 & 74 & \textbf{83} & 94 & 66 & \textbf{96} & 62\\
71 & Medical-ST-Elevation Myocardial Infarction (STEMI) & \textbf{85} & 72 & \textbf{78} & 45 & 54 & 24 & 35\\
72 & Medical-Stroke/\allowbreak TIA & 88 & 74 & \textbf{80} & \textbf{92} & 72 & 80 & 63\\
73 & Medical-Supraventricular Tachycardia (Including Atrial Fibrillation) & 71 & 57 & \textbf{63} & \textbf{72} & 45 & 43 & 44\\
74 & Medical-Syncope & \textbf{80} & 51 & \textbf{62} & 67 & 46 & 64 & 44\\
75 & Medical-Tachycardia & \textbf{59} & 55 & \textbf{57} & 1 & 2 & 0 & 0\\
76 & Medical-Ventricular Tachycardia (With Pulse) & \textbf{61} & 76 & \textbf{68} & 25 & 38 & 16 & 27\\
\addlinespace[2pt]
77 & OB/\allowbreak GYN-Childbirth/\allowbreak Labor/\allowbreak Delivery & 59 & 71 & \textbf{64} & 60 & 53 & \textbf{87} & 53\\
78 & OB/\allowbreak GYN-Eclampsia & \textbf{71} & 89 & \textbf{79} & 5 & 7 & 0 & 0\\
79 & OB/\allowbreak GYN-Gynecologic Emergencies & \textbf{64} & 60 & \textbf{62} & 38 & 38 & 8 & 13\\
80 & OB/\allowbreak GYN-Pregnancy Related Emergencies & \textbf{62} & 53 & \textbf{57} & 6 & 10 & 7 & 11\\
\addlinespace[2pt]
81 & Not Done & \textbf{26} & 41 & \textbf{32} & 2 & 4 & 6 & 10\\
\addlinespace[2pt]
82 & Other & \textbf{31} & 57 & \textbf{40} & 8 & 6 & 0 & 0\\
\end{longtable}
\normalsize

\FloatBarrier
\section{STEP: Ensemble, Pooling, Fitting, Scope and Robustness}
\label{app:step}
This appendix follows the paragraphs of \S\ref{sec:step} for Stage-wise Ensembles for Temporal Pooling (STEP): \ref{app:stepens} the stage predictor and seed ensemble, \ref{app:steppool} temporal pooling (TFA), \ref{app:stepfit} fitting, cross-fitting and selection, \ref{app:stepscope} the scope of pooling, and \ref{app:steprobust} robustness and controls (\S\ref{sec:robust}--\ref{sec:controls}).

\subsection{Stage predictor and seed ensemble}
\label{app:stepens}
The stage models and the single-modality experts are MLPs with the recipe of Table~\ref{tab:hparams}; Table~\ref{tab:experts} lists the expert datasets (eight exist for seven expert slots because the context expert has two mutually exclusive capture-time variants). Table~\ref{tab:steptopk} gives every stage at depths 1, 3, 5 and 10 for the three single seeds, the three-seed ensemble of Equation~\ref{eq:ens} and STEP. The single seeds differ by \seedSpreadMin{} to \seedSpreadMax{} test top-1 points within a stage, the spread quoted in \S\ref{sec:step}. On test the ensemble adds \ensMinusNnOneimpressions{} to \ensMinusNnOnevitals{} top-1 points over the seed-0 NN at stages 4--6 and at most \ensMinusNnOneorgan{} before.

\begin{table}[t]
\centering\scriptsize\setlength{\tabcolsep}{3pt}
\caption{Per-stage detail on the repaired family, clean boundary, test split: the three single seeds of the stage MLP (fit-rows models) with their spread, the 3-seed ensemble and STEP at depths 1, 3, 5 and 10.}
\label{tab:steptopk}
\vspace{4pt}
\begin{tabular}{lrrrrrrrrrrrr}
\toprule
stage & NN s0 & NN s1 & NN s2 & spread & ens.\ top-1 & top-3 & top-5 & top-10 & STEP top-1 & top-3 & top-5 & top-10\\
\midrule
1 & 20.36 & 20.26 & 20.29 & 0.10 & 20.42 & 38.02 & \textbf{48.41} & 62.49 & \textbf{20.72} & \textbf{38.34} & 48.29 & \textbf{62.98}\\
2 & 24.54 & 24.25 & 24.44 & 0.29 & 24.54 & 43.79 & \textbf{54.56} & \textbf{68.08} & \textbf{24.63} & \textbf{43.91} & 54.53 & 68.04\\
3 & 27.31 & 26.99 & 27.38 & 0.39 & 27.56 & 48.16 & 59.38 & 72.79 & \textbf{27.85} & \textbf{48.39} & \textbf{59.40} & \textbf{72.90}\\
4 & 45.84 & 46.06 & 46.50 & 0.66 & 48.16 & 69.60 & 78.31 & 87.14 & \textbf{48.24} & \textbf{70.52} & \textbf{78.89} & \textbf{87.70}\\
5 & 50.34 & 50.91 & 50.15 & 0.77 & 52.94 & 74.53 & 82.06 & 89.77 & \textbf{53.71} & \textbf{75.56} & \textbf{82.98} & \textbf{90.36}\\
6 & 56.03 & 56.43 & 55.65 & 0.78 & 58.34 & 78.56 & 85.09 & 91.73 & \textbf{59.31} & \textbf{79.82} & \textbf{86.36} & \textbf{92.45}\\
\bottomrule
\end{tabular}
\end{table}

\begin{table}[t]
\centering\footnotesize
\caption{The eight block experts on the corrected family (one MLP per block, frozen recipe, trained on the fit rows), inner-validation top-1 per seed.}
\label{tab:experts}
\vspace{4pt}
\begin{tabular}{lrrrrr}
\toprule
expert & columns & seed 0 & seed 1 & seed 2 & ens.\ val top-1\\
\midrule
dispatch & 46 & 18.39 & 18.22 & 18.42 & --\\
complaint location & 10 & 5.79 & 5.73 & 5.71 & --\\
complaint organ system & 12 & 7.08 & 7.09 & 7.08 & --\\
symptoms & 1063 & 30.88 & 30.89 & 31.04 & --\\
vitals & 77 & 16.93 & 17.09 & 17.12 & --\\
impressions & 1246 & 40.12 & 40.63 & 39.97 & --\\
context full & 190 & 10.80 & 10.98 & 10.86 & --\\
context age & 10 & 3.62 & 3.46 & 3.77 & --\\
\bottomrule
\end{tabular}
\end{table}

\subsection{Temporal pooling (TFA)}
\label{app:steppool}
\paragraph{Inputs.} At stage $t$ the pool's inputs are, in order: the unpooled stage ensembles $\bar p_1,\ldots,\bar p_{t-1}$; the experts available at $t$ (dispatch and age-only context at stages 1--3, plus location from stage 2 and organ system from stage 3; dispatch, location, organ system, symptoms and full context at stage 4, plus vitals from stage 5 and impressions at stage 6); and last the joint stage ensemble $\bar p_t$, so $n=2t+1$ (Table~\ref{tab:epochsel}, second column). Pools are not chained: the earlier-stage inputs are ensembles, not earlier pools.

\paragraph{Gate.} The gate of Equation~\ref{eq:gate} maps the confidence trajectory $c(x)\in\mathbb{R}^{3n}$ (each input's entropy, maximum probability and top-1/top-2 margin) through a hidden layer of width 32 with tanh to $n$ weights added to the base weights $w_0$; its output layer is zero-initialised, so the pool starts at the identity. The weights are class-tied (one weight per input, shared by the 82 classes) and the per-class bias $b\in\mathbb{R}^{82}$ is shared by all records.

\paragraph{The pool as a logarithmic opinion pool.} Equation~\ref{eq:pool} gives $\hat p_t(y\mid x)\propto\exp(b_y)\prod_s r_s(y\mid x)^{w_s(x)}$, a weighted logarithmic opinion pool with per-record weights. At $w=\mathbf{e}_n$, $b=0$ it equals $r_n=\bar p_t$ exactly, which is the identity initialisation; with equal fixed weights $1/S$ over $S$ seeds it is the geometric-mean ensemble of Equation~\ref{eq:ens}.

\subsection{Fitting, cross-fitting and selection}
\label{app:stepfit}
\paragraph{Training and validation boundaries.} Every STEP, NN and ensemble number is produced under one protocol, read from the code (\texttt{scripts/paper/review/train\_base.py}, \texttt{scripts/paper/review2/fit\_pools.py}):
(1) the outer split (seed 42) holds out 8,198 test rows, and the inner split (seed 1234) divides the 73,780 training rows into 66,402 fit rows and 7,378 inner-validation rows;
(2) \emph{base models}: for each of the 14 base datasets (six stages, eight expert variants) and each seed, one MLP is trained on the 66,402 fit rows only;
(3) \emph{out-of-fold (OOF) posteriors}: for each dataset and seed, a stratified five-fold split of the fit rows trains five fold models, each predicting the fit rows it did not see, so every fit row receives a posterior from a model that never trained on it; nothing outside the fit rows is used;
(4) \emph{pool fitting}: the stage-$t$ pool is trained on the OOF posteriors of the fit rows (three-seed means of the log-softmax outputs), with AdamW (learning rate $10^{-3}$, weight decay 0, batch 512) for at most 100 epochs with patience 15;
(5) \emph{epoch selection}: after every epoch, and at epoch 0 (the identity), the pool is applied to the base models' posteriors on the 7,378 inner-validation rows, and the epoch with the lowest validation NLL (improvement $>10^{-5}$) is kept;
(6) \emph{test}: the frozen pool is applied once to the same base models' posteriors on the 8,198 test rows; the NN is the seed-0 base model and the ensemble the three-seed base ensemble, so NN, ensemble and STEP are compared on identical rows and base models.
The inner-validation rows are therefore out-of-sample for every base model and are used only for selection (and for the DPCS rates, App.~\ref{app:dpcs}); the test rows enter no fitting or selection step. The gate's initialisation is seeded (seed 0) immediately before the pool is constructed, and fitting runs on CPU with deterministic algorithms. (A pre-repair pipeline drew OOF posteriors from all 73,780 training rows and reported models trained on the whole training split; that protocol is retired and no number here uses it.)

\paragraph{Selected epochs.} Table~\ref{tab:epochsel} gives, per stage, the selected epoch and the validation NLL before and after fitting. Validation never selected the identity epoch; the selected epochs range from \selEpochMin{} to \selEpochMax, and the test top-1 / top-3 changes are those of Table~\ref{tab:main}.

\begin{table}[t]
\centering\footnotesize\setlength{\tabcolsep}{4pt}
\caption{Validation-NLL epoch selection of the STEP pool at every stage (repaired family, clean boundary). Epoch 0 is the identity initialisation, whose output equals the three-seed ensemble, so its validation NLL is the ensemble's; a pool that never improves on it is returned as the ensemble. Validation: the 7,378 inner-validation rows (fit-rows base models); test: the 8,198 test rows, read once after selection. Source: \texttt{review2/pools.json}.}
\label{tab:epochsel}
\vspace{4pt}
\begin{adjustbox}{max width=\linewidth}
\begin{tabular}{lcccccc}
\toprule
stage & inputs $n$ & selected epoch & val.\ NLL, epoch 0 (ens.) & val.\ NLL, selected & test $\Delta_1$ vs ens. & test $\Delta_3$ vs ens.\\
\midrule
1 & 3 & 30 & 3.287 & \textbf{3.262} & $+0.30$ & $+0.32$\\
2 & 5 & 69 & 3.084 & \textbf{3.053} & $+0.09$ & $+0.12$\\
3 & 7 & 9 & 2.916 & \textbf{2.880} & $+0.29$ & $+0.23$\\
4 & 9 & 79 & 2.325 & \textbf{1.987} & $+0.09$ & $+0.91$\\
5 & 11 & 37 & 1.889 & \textbf{1.748} & $+0.77$ & $+1.02$\\
6 & 13 & 37 & 1.747 & \textbf{1.537} & $+0.96$ & $+1.27$\\
\bottomrule
\end{tabular}
\end{adjustbox}
\end{table}

\paragraph{Reproduction.} Base models: \texttt{scripts/paper/review/train\_base.py} (modes \texttt{fitrows} and \texttt{oof\_fit}) through the queue of \texttt{scripts/paper/review2/}; pools and controls: \texttt{scripts/paper/review2/fit\_pools.py} writing \path{artifacts/paper/review2/pools.json}; \path{scripts/paper/review2/step_val_preds.py} refits every pool from the saved logits and asserts that it reproduces the saved test posteriors. Every table of this appendix is generated from the saved result files.

\subsection{Scope of pooling}
\label{app:stepscope}
\paragraph{Provenance-preserving stage-6 representation.} The stage-6 input merges the symptom and impression indicators of a shared ICD-10-CM code into one summed column (App.~\ref{app:features}), so column-set nesting does not imply information nesting at stage 6. The stage-6 joint model was therefore retrained (three seeds, five-fold OOF fits within the fit rows, unchanged recipe and boundaries) on the stage-5 matrix concatenated with the impression expert's own 1,246-column block (\provSepCols{} columns), so that the joint input contains every pool input's raw block (Table~\ref{tab:provenance}). The separated ensemble does not differ significantly from the merged one ($\provSepMinusMergedEnsOne$ top-1, $p=\provEnsPOne$; $\provSepMinusMergedEnsThree$ top-3, $p=\provEnsPThree$), and the pool on the separated joint adds $\provSepDOne$ top-1 ($p=\provSepPOne$) and $\provSepDThree$ top-3 over its ensemble, against $\provMergedDOne$ / $\provMergedDThree$ on the merged one. The stage-6 top-3 increment is therefore not an artefact of the merged representation (the top-1 increment, $p=\provSepPOne$, does not clear the corrected threshold); every other table keeps the merged representation.

\begin{table}[t]
\centering\footnotesize\setlength{\tabcolsep}{3.5pt}
\caption{Provenance-preserving stage-6 input (repaired family, clean boundary, balanced test split, $n=8{,}198$). \emph{Merged}: the paper's stage-6 matrix, where a code documented as both symptom and impression is one summed column. \emph{Separated}: the stage-5 matrix concatenated with the impression expert's own block (2{,}644 columns), retrained with the unchanged recipe (three seeds, five-fold out-of-fold fits within the fit rows) and pooled with every other input unchanged. Top-1 / top-3 in \%; $\Delta$ is STEP minus its own ensemble with the exact paired McNemar $p$. Separated minus merged ensemble: $-0.06$ top-1 ($p=0.88$), $-0.07$ top-3 ($p=0.84$). Sources: \texttt{review2/pools.json}, \texttt{review2/pools\_sep.json}; seeds 1--2 and the ensemble test from the saved per-record logits.}
\label{tab:provenance}
\vspace{4pt}
\begin{adjustbox}{max width=\linewidth}
\begin{tabular}{lccccc}
\toprule
stage-6 joint input & NN seeds 0 / 1 / 2 top-1 & ens.\ top-1 / top-3 & STEP top-1 / top-3 & $\Delta_1$ ($p$) & $\Delta_3$ ($p$)\\
\midrule
merged & 56.03 / \textbf{56.43} / 55.65 & \textbf{58.34} / \textbf{78.56} & \textbf{59.31} / \textbf{79.82} & $+0.96$ ($7.7\times10^{-4}$) & $+1.27$ ($6.8\times10^{-7}$)\\
separated & \textbf{56.11} / 56.07 / \textbf{55.83} & 58.28 / 78.48 & 59.03 / \textbf{79.82} & $+0.74$ ($0.010$) & $+1.34$ ($1.4\times10^{-7}$)\\
\bottomrule
\end{tabular}
\end{adjustbox}
\end{table}

\paragraph{Why the joint posterior is an input: additivity in log-odds.}\label{app:additivity} Let $u$ and $v$ be the raw blocks of two modalities with experts $q_u(y\mid u)$ and $q_v(y\mid v)$. With fixed, class-tied weights the pool of Equation~\ref{eq:pool} gives, for any two classes $y,y'$,
$\log\frac{\hat p(y\mid u,v)}{\hat p(y'\mid u,v)}=w_u\log\frac{q_u(y\mid u)}{q_u(y'\mid u)}+w_v\log\frac{q_v(y\mid v)}{q_v(y'\mid v)}+(b_y-b_{y'})$,
because the softmax normaliser cancels in the ratio. The pooled log-odds is therefore additive in the two experts' log-odds, a function $g(u)+h(v)$. An additive rule can favour $y$ when both experts do, but it cannot represent a non-additive interaction: if the true log-odds is $+c$ when exactly one of two binary indicators is set and $-c$ otherwise (an exclusive-or between the modalities), then $g(1)+h(0)=g(0)+h(1)=c$ and $g(0)+h(0)=g(1)+h(1)=-c$, whose two sums contradict each other. Only an input that sees both blocks at once, the joint predictor, can represent such a rule, which is why $\bar p_t$ is always an input. With the gate, $w=w(c(x))$ depends on all inputs' confidence summaries, so the pooled log-odds is no longer exactly additive; the gate can make the weight on $q_u$ depend on how confident $q_v$ is, a restricted interaction that sees $v$ only through the entropy, maximum and margin of $q_v$, with weights shared by all 82 classes. No claim rests on this argument alone; the empirical evidence follows.

\paragraph{Pooling the experts without the joint posterior.} On the repaired family, a static pool over the block experts available at the stage (no joint posterior, fitted and selected as every other pool) reaches \expOnlyStaticscene, \expOnlyStaticvitals{} and \expOnlyStaticimpressions\% test top-1 at stages 4--6, and the gated pool over the same experts \expOnlyGatedscene, \expOnlyGatedvitals{} and \expOnlyGatedimpressions\%, against the ensemble's \ensTopOnescene, \ensTopOnevitals{} and \ensTopOneimpressions\% (\texttt{review2/conjunction.json}; the conjunction test of App.~\ref{app:diag} uses the same pools). A historical validation test on the pre-repair features found the same deficit from stage 2 on (App.~\ref{app:stats}).

\subsection{Robustness and controls}
\label{app:steprobust}
\paragraph{Stage-wise results.} STEP ranks the recorded protocol first for \stepTopOnedispatch\% of test encounters from dispatch alone and for \stepTopOneimpressions\% once impressions are documented, and places it in the top three for \stepTopThreedispatch\% and \stepTopThreeimpressions\% (Table~\ref{tab:main}). The scene stage alone is worth \stepGainOneOneDecscene{} top-1 points, more than the other four transitions together, because the symptom codes and the four scene-time context fields arrive there at once; the remaining transitions add \stepGainOneOneDecorgan{} to \stepGainOneOneDecimpressions{} points. The pool adds $\dOnevitals$ and $\dOneimpressions$ top-1 and $\dThreevitals$ and $\dThreeimpressions$ top-3 points over the ensemble at stages 5--6, a top-3 gain of $\dThreescene$ at stage 4 (top-1 $\dOnescene$), and $\dOnelocation$ to $\dOnedispatch$ top-1 points at stages 1--3, none significant.

\paragraph{Replications and years.} Table~\ref{tab:robust} collects, stage by stage, the increment under the five sources of evidence drawn in Figure~\ref{fig:main}(b, c): the main run with its exact $p$, the static pool, the four end-to-end replications and the two cross-year directions. Table~\ref{tab:reps} gives the four replications, each repeating the whole pipeline (fourteen base datasets, three seeds each, five-fold cross-fitting within the fit rows, pool fitting and validation selection) with a disjoint seed triple; the cross-year transfer is in Table~\ref{tab:year} (App.~\ref{app:diag}).

\begin{table}[t]
\centering\scriptsize\setlength{\tabcolsep}{2.6pt}
\caption{Is the increment robust? Group 1: the main run (STEP minus the 3-seed ensemble, exact McNemar $p$; the corrected family). Group 2: the static log-linear pool (fixed weights, no gate) on the same inputs. Group 3: the whole pipeline repeated with four disjoint seed triples (Table~\ref{tab:reps}). Group 4: models trained on one year's balanced fit rows, applied once to the other year's balanced cohort (80,284 and 81,978 records; Table~\ref{tab:year}); every stage-4--6 cross-year increment has $p<10^{-3}$. Groups 2--4 are descriptive.}
\label{tab:robust}
\vspace{4pt}
\resizebox{\textwidth}{!}{\begin{tabular}{lllcccll}
\toprule
\multirow{2}{*}{stage} & \multicolumn{2}{c}{main run: STEP $-$ ens.\ ($p$)} & static pool & \multicolumn{2}{c}{four runs: mean [min, max]} & \multicolumn{2}{c}{cross-year: STEP $-$ ens.}\\
\cmidrule(lr){2-3}\cmidrule(lr){4-4}\cmidrule(lr){5-6}\cmidrule(lr){7-8}
 & $\Delta_1$ & $\Delta_3$ & $\Delta_1$ / $\Delta_3$ & $\Delta_1$ & $\Delta_3$ & 24$\to$23 $\Delta_1$ / $\Delta_3$ & 23$\to$24 $\Delta_1$ / $\Delta_3$\\
\midrule
1 dispatch & $+0.30$ ($0.06$) & {\bfseries\boldmath $+0.32$} ($0.13$) & {\bfseries\boldmath $+0.37$} / $+0.29$ & $+0.23$ [+0.15, +0.30] & $+0.21$ [+0.11, +0.32] & $+0.09$ / $+0.23$ & $+0.14$ / $+0.22$\\
2 location & {\bfseries\boldmath $+0.09$} ($0.59$) & $+0.12$ ($0.6$) & $+0.04$ / {\bfseries\boldmath $+0.16$} & $+0.05$ [-0.09, +0.12] & $+0.17$ [+0.11, +0.27] & $+0.12$ / $+0.20$ & $+0.02$ / $+0.23$\\
3 organ sys. & {\bfseries\boldmath $+0.29$} ($0.098$) & $+0.23$ ($0.3$) & $+0.16$ / {\bfseries\boldmath $+0.27$} & $+0.22$ [+0.06, +0.35] & $+0.31$ [+0.15, +0.45] & $-0.02$ / $-0.02$ & $+0.17$ / $-0.02$\\
4 scene & {\bfseries\boldmath $+0.09$} ($0.8$) & {\bfseries\boldmath $+0.91$} ($6.2\!\times\!10^{-4}$) & $-0.05$ / $+0.76$ & $+0.35$ [+0.09, +0.71] & $+0.92$ [+0.72, +1.12] & $+0.98$ / $+1.01$ & $+0.32$ / $+0.98$\\
5 vitals & {\bfseries\boldmath $+0.77$} ($0.0051$) & {\bfseries\boldmath $+1.02$} ($3.7\!\times\!10^{-5}$) & $+0.37$ / $+0.82$ & $+0.63$ [+0.38, +0.83] & $+1.09$ [+0.81, +1.43] & $+1.11$ / $+1.24$ & $+1.00$ / $+1.26$\\
6 impressions & {\bfseries\boldmath $+0.96$} ($7.7\!\times\!10^{-4}$) & {\bfseries\boldmath $+1.27$} ($6.8\!\times\!10^{-7}$) & $+0.57$ / $+1.18$ & $+1.06$ [+0.96, +1.17] & $+1.39$ [+1.21, +1.68] & $+1.15$ / $+1.36$ & $+1.00$ / $+1.24$\\
\bottomrule
\end{tabular}}
\end{table}

\begin{table}[t]
\centering\scriptsize\setlength{\tabcolsep}{2.5pt}
\caption{End-to-end run variability: the full pipeline (fourteen base datasets, three seeds each, five-fold cross-fitting within the fit rows, pool fitting and validation selection) repeated with four disjoint seed triples. Each cell gives the ensemble and STEP test top-1 with the paired increment (exact McNemar $p$) and the top-3 increment; the last columns average the increments over the four independent runs and give their range.}
\label{tab:reps}
\vspace{4pt}
\begin{adjustbox}{max width=\linewidth}
\begin{tabular}{lrrrrrr}
\toprule
stage & seeds 0--2 & seeds 3--5 & seeds 6--8 & seeds 9--11 & mean $\Delta_1$ [min, max] & mean $\Delta_3$ [min, max]\\
\midrule
1 & \stk{20.42 $\to$ \textbf{20.72}\\($+0.30$, 0.06)\\top-3 $+0.32$} & \stk{20.64 $\to$ \textbf{20.79}\\($+0.15$, 0.32)\\top-3 $+0.29$} & \stk{20.53 $\to$ \textbf{20.81}\\($+0.28$, 0.15)\\top-3 $+0.11$} & \stk{20.47 $\to$ \textbf{20.65}\\($+0.18$, 0.25)\\top-3 $+0.12$} & $+0.23$ [+0.15, +0.30] & $+0.21$ [+0.11, +0.32]\\
2 & \stk{24.54 $\to$ \textbf{24.63}\\($+0.09$, 0.59)\\top-3 $+0.12$} & \stk{24.47 $\to$ \textbf{24.57}\\($+0.10$, 0.58)\\top-3 $+0.18$} & \stk{24.54 $\to$ \textbf{24.66}\\($+0.12$, 0.52)\\top-3 $+0.11$} & \stk{\textbf{24.59} $\to$ 24.51\\($-0.09$, 0.64)\\top-3 $+0.27$} & $+0.05$ [-0.09, +0.12] & $+0.17$ [+0.11, +0.27]\\
3 & \stk{27.56 $\to$ \textbf{27.85}\\($+0.29$, 0.098)\\top-3 $+0.23$} & \stk{27.71 $\to$ \textbf{27.88}\\($+0.17$, 0.41)\\top-3 $+0.45$} & \stk{27.68 $\to$ \textbf{27.74}\\($+0.06$, 0.79)\\top-3 $+0.15$} & \stk{27.48 $\to$ \textbf{27.84}\\($+0.35$, 0.063)\\top-3 $+0.41$} & $+0.22$ [+0.06, +0.35] & $+0.31$ [+0.15, +0.45]\\
4 & \stk{48.16 $\to$ \textbf{48.24}\\($+0.09$, 0.8)\\top-3 $+0.91$} & \stk{47.76 $\to$ \textbf{48.46}\\($+0.71$, 0.0091)\\top-3 $+0.72$} & \stk{47.89 $\to$ \textbf{48.12}\\($+0.23$, 0.43)\\top-3 $+0.93$} & \stk{47.84 $\to$ \textbf{48.22}\\($+0.38$, 0.18)\\top-3 $+1.12$} & $+0.35$ [+0.09, +0.71] & $+0.92$ [+0.72, +1.12]\\
5 & \stk{52.94 $\to$ \textbf{53.71}\\($+0.77$, 0.0051)\\top-3 $+1.02$} & \stk{53.35 $\to$ \textbf{53.73}\\($+0.38$, 0.19)\\top-3 $+1.43$} & \stk{52.98 $\to$ \textbf{53.54}\\($+0.56$, 0.047)\\top-3 $+0.81$} & \stk{52.60 $\to$ \textbf{53.43}\\($+0.83$, 0.0036)\\top-3 $+1.12$} & $+0.63$ [+0.38, +0.83] & $+1.09$ [+0.81, +1.43]\\
6 & \stk{58.34 $\to$ \textbf{59.31}\\($+0.96$, 0.00077)\\top-3 $+1.27$} & \stk{58.40 $\to$ \textbf{59.58}\\($+1.17$, 6.3e-05)\\top-3 $+1.40$} & \stk{58.45 $\to$ \textbf{59.59}\\($+1.13$, 6.7e-05)\\top-3 $+1.21$} & \stk{58.38 $\to$ \textbf{59.36}\\($+0.98$, 0.00055)\\top-3 $+1.68$} & $+1.06$ [+0.96, +1.17] & $+1.39$ [+1.21, +1.68]\\
\bottomrule
\end{tabular}
\end{adjustbox}
\end{table}

\paragraph{Controls on identical inputs.} Table~\ref{tab:controls} fits, at every stage on the identical input set and with the identical optimiser, batch size, patience and validation-NLL selection, the alternatives a reader would ask for. \emph{Recalibration} of the ensemble alone (temperature and vector scaling) leaves top-1 within 0.1 points except vector scaling at stage 4 ($-0.61$) and lowers stage-6 NLL from \stageSixEnsNll{} to \stageSixVectorNll, about half of the pool's reduction (to \stageSixStepNll). \emph{The fixed-weight cumulative product} over the stage posteriors (PA; $w_s=s/t$, no learning) loses accuracy from the scene stage on: with information-nested stages the earlier members are far weaker than the current one and an unlearned product over-weights them. \emph{The static pool} (one weight per input plus a class bias; \citealp{wolczyk2021ztw,heskes1998selecting}) recovers most of the gain ($\staticOnevitals$ / $\staticThreevitals$ and $\staticOneimpressions$ / $\staticThreeimpressions$ at stages 5--6), and its non-negative \citep{breiman1996stacked} and ridge variants reproduce it closely (within 0.07 points at stages 5--6). \emph{The gate} adds $\gateOnevitals$ and $\gateOneimpressions$ top-1 on top at stages 5--6. \emph{The two axes}: the history-only pool (stage ensembles plus the joint posterior) and the expert-only pool (experts plus the joint posterior) each gain at stages 5--6, and their combination, STEP, beats both.

\begin{table}[t]
\centering\scriptsize\setlength{\tabcolsep}{2.5pt}
\caption{[Cells: top-1 above top-3.] Controls on the corrected family, clean boundary, test split: each row is a model fitted on the identical out-of-fold inputs and selected on validation NLL (epoch 0 = identity is a candidate), minus the 3-seed ensemble, top-1 / top-3 in points with exact McNemar $p$ (descriptive; 9 models $\times$ 6 stages). PA is the fixed-weight cumulative product of \citet{jazbec2023anytime} over the stage ensembles (no learning); the static pool is the weighted geometric mean with class bias of \citet{heskes1998selecting} and \citet{wolczyk2021ztw}; the non-negative variant is Breiman's constraint; ridge shrinks the weights toward the identity ($\lambda$ selected on validation).}
\label{tab:controls}
\vspace{4pt}
\begin{adjustbox}{max width=\linewidth}
\begin{tabular}{lrrrrrr}
\toprule
model $-$ ensemble & stage 1 & stage 2 & stage 3 & stage 4 & stage 5 & stage 6\\
\midrule
temperature scaling (ens.) & \stk{$+0.00$ (1)\\$+0.00$ (1)} & \stk{$+0.00$ (1)\\$+0.00$ (1)} & \stk{$+0.00$ (1)\\$+0.00$ (1)} & \stk{$+0.00$ (1)\\$+0.00$ (1)} & \stk{$+0.00$ (1)\\$+0.00$ (1)} & \stk{$+0.00$ (1)\\$+0.00$ (1)}\\
vector scaling (ens.) & \stk{$+0.10$ (0.28)\\$-0.01$ (1)} & \stk{$+0.06$ (0.58)\\$+0.17$ (0.23)} & \stk{$-0.09$ (0.48)\\$-0.16$ (0.32)} & \stk{$-0.61$ (0.005)\\$-0.16$ (0.44)} & \stk{$+0.00$ (1)\\$-0.33$ (0.06)} & \stk{$-0.10$ (0.54)\\$+0.32$ (0.032)}\\
PA fixed product, $w_s{=}s/t$ & -- & \stk{$-0.26$ (0.21)\\$-0.23$ (0.35)} & \stk{$-0.48$ (0.045)\\$-0.51$ (0.042)} & \stk{$-3.05$ ($3.1\!\times\!10^{-16}$)\\$-1.93$ ($1.9\!\times\!10^{-8}$)} & \stk{$-1.87$ ($5.7\!\times\!10^{-8}$)\\$-1.35$ ($1.6\!\times\!10^{-5}$)} & \stk{$-1.54$ ($7.9\!\times\!10^{-6}$)\\$-0.72$ (0.022)}\\
static log-linear pool & \stk{{\bfseries\boldmath $+0.37$} (0.027)\\$+0.29$ (0.16)} & \stk{$+0.04$ (0.86)\\$+0.16$ (0.5)} & \stk{$+0.16$ (0.38)\\$+0.27$ (0.21)} & \stk{$-0.05$ (0.9)\\$+0.76$ (0.0053)} & \stk{$+0.37$ (0.17)\\$+0.82$ ($8.2\!\times\!10^{-4}$)} & \stk{$+0.57$ (0.042)\\$+1.18$ ($3.8\!\times\!10^{-6}$)}\\
static pool, $w\ge0$ & \stk{{\bfseries\boldmath $+0.37$} (0.027)\\$+0.29$ (0.16)} & \stk{$+0.02$ (0.93)\\$+0.09$ (0.73)} & \stk{$+0.11$ (0.56)\\$+0.24$ (0.26)} & \stk{$+0.09$ (0.79)\\$+0.72$ (0.0078)} & \stk{$+0.39$ (0.13)\\$+0.76$ (0.0019)} & \stk{$+0.63$ (0.024)\\$+1.11$ ($9.2\!\times\!10^{-6}$)}\\
static pool, ridge to identity & \stk{$+0.34$ (0.037)\\$+0.29$ (0.16)} & \stk{$+0.00$ (1)\\$+0.16$ (0.5)} & \stk{$+0.11$ (0.55)\\$+0.24$ (0.26)} & \stk{$-0.04$ (0.93)\\$+0.79$ (0.0034)} & \stk{$+0.34$ (0.19)\\$+0.76$ (0.0017)} & \stk{$+0.60$ (0.03)\\$+1.15$ ($4.9\!\times\!10^{-6}$)}\\
gated pool, stages only & -- & \stk{$-0.15$ (0.23)\\$+0.18$ (0.28)} & \stk{$-0.12$ (0.35)\\$+0.17$ (0.29)} & \stk{$-0.18$ (0.44)\\$+0.28$ (0.2)} & \stk{$+0.26$ (0.27)\\$+0.72$ ($6.5\!\times\!10^{-4}$)} & \stk{$+0.35$ (0.17)\\$+0.39$ (0.094)}\\
gated pool, experts + joint & \stk{$+0.30$ (0.06)\\{\bfseries\boldmath $+0.32$} (0.13)} & \stk{{\bfseries\boldmath $+0.15$} (0.34)\\{\bfseries\boldmath $+0.20$} (0.39)} & \stk{$+0.28$ (0.13)\\{\bfseries\boldmath $+0.28$} (0.21)} & \stk{{\bfseries\boldmath $+0.11$} (0.72)\\{\bfseries\boldmath $+0.98$} ($1.1\!\times\!10^{-4}$)} & \stk{$+0.21$ (0.44)\\$+0.71$ (0.0019)} & \stk{$+0.37$ (0.2)\\$+0.88$ ($1.9\!\times\!10^{-4}$)}\\
STEP (gated, both axes) & \stk{$+0.30$ (0.06)\\{\bfseries\boldmath $+0.32$} (0.13)} & \stk{$+0.09$ (0.59)\\$+0.12$ (0.6)} & \stk{{\bfseries\boldmath $+0.29$} (0.098)\\$+0.23$ (0.3)} & \stk{$+0.09$ (0.8)\\$+0.91$ ($6.2\!\times\!10^{-4}$)} & \stk{{\bfseries\boldmath $+0.77$} (0.0051)\\{\bfseries\boldmath $+1.02$} ($3.7\!\times\!10^{-5}$)} & \stk{{\bfseries\boldmath $+0.96$} ($7.7\!\times\!10^{-4}$)\\{\bfseries\boldmath $+1.27$} ($6.8\!\times\!10^{-7}$)}\\
\midrule
ensemble top-1 / top-3 & 20.42 / 38.02 & 24.54 / 43.79 & 27.56 / 48.16 & 48.16 / 69.60 & 52.94 / 74.53 & 58.34 / 78.56\\
ensemble NLL & 3.240 & 3.035 & 2.868 & 2.266 & 1.860 & 1.696\\
STEP NLL & \textbf{3.221} & \textbf{3.006} & \textbf{2.834} & \textbf{1.943} & \textbf{1.720} & \textbf{1.498}\\
static NLL & 3.224 & 3.013 & 2.842 & 2.018 & 1.760 & 1.534\\
\bottomrule
\end{tabular}
\end{adjustbox}
\end{table}

\paragraph{Equal weighting and a direct static-versus-gated test.} Table~\ref{tab:e1pool} adds unfitted equal-weight pools over exactly STEP's inputs (arithmetic and geometric means of the renormalised posteriors), the geometric pool with only a class bias fitted, and direct paired tests of STEP against the static pool; all fitted pools share the out-of-fold rows, optimiser and validation-NLL selection. Equal weighting loses to the seed ensemble from stage \eoneEqWorseFrom{} on because the weak early views dominate an unweighted pool; fitting only the class bias does not repair it (stage 6: \eoneEqBiasOneSix\%), while the learned static weights add $\eoneStaticMinusEqBiasOneSix$ top-1 points over it. The current-stage ensemble has the lowest validation NLL of any single input at every stage, so a best-single-input selector returns the seed ensemble. STEP minus static is positive on top-1 at stages 4--6 in every seed triple (stage 6: mean $\eoneGSmeanOneSix$, range [$\eoneGSminOneSix$, $\eoneGSmaxOneSix$]) and lowers test NLL at every stage (stage 6: $\onNllSix$, interval Bonferroni-adjusted over the six stages, declared after the accuracy tests, [$\onNllSixLo$, $\onNllSixHi$], excluding zero at stages \onNllBonfStages; lower in \onNllRunsNeg{} of \onNllRunsTot{} run--stage pairs, whose seed triples share every record; a likelihood difference on these records, not a calibration claim), but none of the 12 paired accuracy tests clears $0.05/12$ (smallest $p=\eoneGSminp$), and top-3 differences change sign across seed triples at stages 5--6. These are new analyses of a test split already inspected in earlier rounds.

\begin{table}[t]
\centering
\footnotesize
\setlength{\tabcolsep}{3pt}
\caption{Averaging, learned pooling and gating on identical inputs (each stage's earlier-stage, expert and current-stage posteriors, as in STEP), balanced test split, top-1 / top-3 (\%), main seed triple. The equal pools are not fitted; the class-bias, static and STEP pools are fitted on the same out-of-fold fit rows with the same optimiser and selected on the same validation NLL (epoch 0 a candidate). Lower block: STEP $-$ static on the same records, with exact McNemar $p$ (12 tests; Bonferroni $\alpha{=}0.05/12$), the NLL difference (negative favours STEP; every 95\% bootstrap interval excludes 0), and the mean [min, max] over four disjoint seed triples. Script: \texttt{scripts/paper/revision/e1\_pooling\_controls.py}.}
\label{tab:e1pool}
\vspace{4pt}
\begin{adjustbox}{max width=\linewidth}
\begin{tabular}{lcccccc}
\toprule
pool over identical inputs & stage 1 & stage 2 & stage 3 & stage 4 & stage 5 & stage 6\\
\midrule
stage-$t$ seed ensemble & 20.42 / 38.02 & 24.54 / 43.79 & 27.56 / 48.16 & 48.16 / 69.60 & 52.94 / 74.53 & 58.34 / 78.56\\
equal arithmetic pool & 20.41 / 38.17 & 23.70 / 42.72 & 26.47 / 46.85 & 45.96 / 66.30 & 51.23 / 72.03 & 56.75 / 77.37\\
equal geometric pool & 20.76 / 38.33 & 24.27 / 43.30 & 26.98 / 47.71 & 44.55 / 66.82 & 49.96 / 72.47 & 56.22 / 77.25\\
equal geometric pool + class bias & 20.46 / 38.24 & 23.82 / 42.97 & 26.03 / 46.52 & 43.83 / 66.17 & 49.52 / 71.65 & 55.10 / 76.63\\
static pool (learned $w$, $b$) & \textbf{20.79} / 38.31 & 24.58 / \textbf{43.95} & 27.71 / \textbf{48.43} & 48.11 / 70.36 & 53.31 / 75.35 & 58.92 / 79.74\\
STEP (gated $w$, $b$) & 20.72 / \textbf{38.34} & \textbf{24.63} / 43.91 & \textbf{27.85} / 48.39 & \textbf{48.24} / \textbf{70.52} & \textbf{53.71} / \textbf{75.56} & \textbf{59.31} / \textbf{79.82}\\
\midrule
STEP $-$ static, top-1 ($p$) & $-0.06$ (0.5) & $+0.05$ (0.66) & $+0.13$ (0.31) & $+0.13$ (0.5) & $+0.40$ (0.033) & $+0.39$ (0.033)\\
STEP $-$ static, top-3 ($p$) & $+0.02$ (0.89) & $-0.04$ (0.86) & $-0.04$ (0.86) & $+0.16$ (0.43) & $+0.21$ (0.23) & $+0.09$ (0.64)\\
STEP $-$ static, NLL & $-0.003$ & $-0.007$ & $-0.009$ & $-0.075$ & $-0.040$ & $-0.035$\\
four runs, top-1 & $-0.01$ [-0.06, +0.05] & $+0.03$ [+0.00, +0.05] & $+0.10$ [-0.06, +0.17] & $+0.24$ [+0.13, +0.37] & $+0.20$ [+0.06, +0.40] & $+0.32$ [+0.26, +0.39]\\
four runs, top-3 & $+0.08$ [+0.01, +0.20] & $-0.03$ [-0.12, +0.02] & $-0.02$ [-0.10, +0.05] & $+0.30$ [+0.16, +0.52] & $+0.11$ [-0.18, +0.22] & $+0.19$ [-0.02, +0.38]\\
\bottomrule
\end{tabular}
\end{adjustbox}
\end{table}

\paragraph{Model-count parity.}\label{app:parity} STEP's stage-6 inputs come from 39 networks, the ensemble's from three, and the equal-weight controls above average weaker restricted views rather than more copies of the joint model. Table~\ref{tab:parity} therefore adds, from the saved predictions of the replication seeds, a twelve-seed ensemble of each stage's joint MLP. It attains STEP's accuracy at stages 4--6 (STEP minus it: $\parStepEnsTwelveOneSix$ [$\parStepEnsTwelveOneLoSix$, $\parStepEnsTwelveOneHiSix$] / $\parStepEnsTwelveThreeSix$ [$\parStepEnsTwelveThreeLoSix$, $\parStepEnsTwelveThreeHiSix$] at stage 6, all $p\ge\parStepEnsTwelvePmin$; differences of about half a point either way cannot be excluded), and the static pool falls below it on stage-5 top-1 ($\parStaticEnsTwelveOneFive$, $p=\parStaticEnsTwelvePOneFive$): the increment of Table~\ref{tab:main} over the three-seed ensemble is matched by nine more seeds of the joint model, at twelve rather than \onFitsPooled{} network fits. Given twelve seeds of every input as well (the four seed-triple pools averaged, 156 networks at stage 6, so seeds rather than networks are matched), STEP leads the twelve-seed ensemble by $\parStepFourThreeMin$ to $\parStepFourThreeMax$ top-3 at stages 4--6 ($p=\parStepFourThreePmin$--$\parStepFourThreePmax$, not surviving correction over these six tests) and not on top-1 ($p\ge\parStepFourOnePmin$), and its NLL is lower at every stage (stage 6: \parStepNllSix{} against \parEnsTwelveNllSix{}). Most of that NLL gap is the seed ensemble's overconfidence: with a temperature fitted on the inner-validation rows for each model (a further declared diagnostic), STEP's test NLL remains lower than the twelve-seed ensemble's by \tsNllAbsMin--\tsNllAbsMax{} nats at stages 4--6 (stage 6: $\tsStepEnsTwelveNllSix$, 95\% interval [$\tsStepEnsTwelveNllLoSix$, $\tsStepEnsTwelveNllHiSix$]), while the maximum posterior detects a wrong top-1 equally well at stage 6 (AUROC \tsStepAucSix{} against \tsEnsTwelveAucSix) and slightly better at stage 4 (\tsStepAucFour{} against \tsEnsTwelveAucFour). The static pool's NLL (\parStaticNllSix{} at stage 6, \parStaticNllFive{} at stage 5; fitted source temperature 0.96--0.97, so scaling would change it little) is no lower than the temperature-scaled twelve-seed ensemble's (\tsEnsTwelveNllSix{} and \tsEnsTwelveNllFive; an unpaired comparison). Beyond the seed count of the joint model, the likelihood improvement therefore comes from the gate, and pooling adds at most a small shortlist gain, with no demonstrated downstream use of either. This diagnostic was declared after review and uses the already inspected test split.

\begin{table}[t]
\centering\scriptsize\setlength{\tabcolsep}{3pt}
\caption{Model-count parity (test, $n=8{,}198$; top-1 / top-3 \%, NLL). \emph{ens.\ 3 / 12}: geometric mean of 3 or 12 seeds (0--11) of the stage-$t$ MLP; \emph{STEP} and \emph{static}: the paper's pools on seeds 0--2; \emph{STEP$\times$4}: geometric mean of the four seed-triple STEP pools (12 seeds of every input). From saved test predictions only, as a declared diagnostic after review; tests are exact McNemar, outside the corrected family. Script: \texttt{scripts/paper/revision/overnight/diag\_parity\_ens12.py}.}
\label{tab:parity}
\vspace{4pt}
\begin{adjustbox}{max width=\linewidth}
\begin{tabular}{lcccccll}
\toprule
stage & ens.\ 3 & ens.\ 12 & static & STEP & STEP$\times$4 & STEP $-$ ens.\ 12 top-1 ($p$) / top-3 ($p$) & STEP$\times$4 $-$ ens.\ 12 top-1 ($p$) / top-3 ($p$)\\
\midrule
4 & 48.16 / 69.60 / 2.266 & 48.63 / 70.61 / 2.186 & 48.11 / 70.36 / 2.018 & 48.24 / 70.52 / 1.943 & \textbf{48.87} / \textbf{71.24} / \textbf{1.915} & $-0.39$ (0.2) / $-0.10$ (0.76) & {\bfseries\boldmath $+0.23$} (0.41) / {\bfseries\boldmath $+0.62$} (0.015)\\
5 & 52.94 / 74.53 / 1.860 & 54.15 / 75.35 / 1.795 & 53.31 / 75.35 / 1.760 & 53.71 / 75.56 / 1.720 & \textbf{54.22} / \textbf{75.88} / \textbf{1.693} & $-0.44$ (0.13) / $+0.21$ (0.45) & {\bfseries\boldmath $+0.07$} (0.81) / {\bfseries\boldmath $+0.54$} (0.026)\\
6 & 58.34 / 78.56 / 1.696 & 59.29 / 79.76 / 1.620 & 58.92 / 79.74 / 1.534 & 59.31 / 79.82 / 1.498 & \textbf{59.64} / \textbf{80.31} / \textbf{1.474} & $+0.01$ (1) / $+0.06$ (0.84) & {\bfseries\boldmath $+0.34$} (0.21) / {\bfseries\boldmath $+0.55$} (0.018)\\
\bottomrule
\end{tabular}
\end{adjustbox}
\end{table}

\paragraph{A stronger tabular baseline: gradient-boosted trees.} Table~\ref{tab:gbm} trains XGBoost (GPU histogram method) at every stage on the identical feature matrices, records and splits, with a small grid (depth 4, 6, 8 at learning rate 0.1 and depth 6 at 0.05; subsample 0.8, column subsample 0.5, 64 bins, seed 0), early stopping and selection on inner-validation log-loss, and the selected fit-rows model evaluated once on test. The boosted model beats the single-seed MLP by \gbmMinusNnOneOneDecscene{} to \gbmMinusNnOneOneDecvitals{} top-1 points at stages 4--6 and comes within one point of the three-seed MLP ensemble, so the single network is an under-competitive comparator but the ensemble is not; STEP leads the boosted model by $\stepMinusGbmOneOneDecscene$ to $\stepMinusGbmOneOneDecimpressions$ top-1 points at stages 4--6 ($p\le\gbmPmaxOne$). Applying the unchanged pool recipe to XGBoost stage and expert posteriors (cross-fitted within the fit rows) adds $\gbmPoolOnevitals$ / $\gbmPoolThreevitals$ and $\gbmPoolOneimpressions$ / $\gbmPoolThreeimpressions$ top-1 / top-3 points over the single boosted model at stages 5--6 ($p\le\gbmPoolPmax$), so the increment appears for a second predictor family.

\begin{table}[t]
\centering\scriptsize\setlength{\tabcolsep}{3pt}
\caption{A tuned gradient-boosted baseline (XGBoost, GPU histogram method) on the repaired family, identical feature matrices, records and splits: grid over depth 4/6/8 at learning rate 0.1 and depth 6 at 0.05, early stopping and selection on inner-validation log-loss, fit-rows model evaluated once on test. Paired against the single-seed MLP, the 3-seed MLP ensemble and STEP (all fit-rows, clean boundary); descriptive, 18 comparisons.}
\label{tab:gbm}
\vspace{4pt}
\begin{adjustbox}{max width=\linewidth}
\begin{tabular}{lrrrrrr}
\toprule
stage & config & GBM top-1 / top-3 & GBM $-$ NN & GBM $-$ ens. & STEP $-$ GBM $\Delta_1$ [CI] ($p$) & $\Delta_3$ ($p$)\\
\midrule
1 & d4, $\eta$=0.1, 169 & 20.16 / 38.29 & $-0.20$ / $+0.20$ & $-0.26$ (0.35) / $+0.27$ (0.32) & $+0.56$ [0.02, 1.10] (0.043) & $+0.05$ (0.89)\\
2 & d4, $\eta$=0.1, 200 & 24.52 / 43.71 & $-0.02$ / $+0.33$ & $-0.02$ (0.96) / $-0.09$ (0.79) & $+0.11$ [-0.34, 0.57] (0.68) & $+0.21$ (0.47)\\
3 & d4, $\eta$=0.1, 224 & 27.32 / 48.29 & $+0.01$ / $+0.50$ & $-0.23$ (0.42) / $+0.13$ (0.67) & $+0.52$ [-0.01, 1.07] (0.054) & $+0.10$ (0.76)\\
4 & d6, $\eta$=0.05, 576 & 47.23 / 68.87 & $+1.39$ / $+1.55$ & $-0.93$ (0.014) / $-0.73$ (0.042) & $+1.01$ [0.29, 1.74] (0.007) & $+1.65$ (2.3e-06)\\
5 & d6, $\eta$=0.05, 563 & 52.53 / 74.16 & $+2.18$ / $+2.04$ & $-0.41$ (0.32) / $-0.37$ (0.33) & $+1.18$ [0.38, 1.93] (0.003) & $+1.39$ (8.9e-05)\\
6 & d6, $\eta$=0.05, 528 & 58.09 / 79.24 & $+2.06$ / $+3.26$ & $-0.26$ (0.54) / $+0.68$ (0.054) & $+1.22$ [0.46, 1.95] (0.0017) & $+0.59$ (0.083)\\
\bottomrule
\end{tabular}
\end{adjustbox}
\end{table}

\paragraph{Accuracy and cost of the trained systems.} Table~\ref{tab:e3cost} times the trained seed ensemble, the block-masked ensemble, the tuned XGBoost, the static pool and STEP on the same CPU thread, after checking that every timed model reproduces the saved predictions behind the paper's tables. Model compute is small for every system: a cold stage-6 STEP request takes \onLatStepCold~ms, about \onStepOverEnsCold{} times a stage-6 seed-ensemble request (\onLatEns~ms), and the six refreshes of a whole encounter, each computing only what is new, take \onSessStep~ms for STEP against \onSessEns~ms for the six stage ensembles. These are single-thread server compute times; they exclude data entry, interface and network latency and do not establish clinical usability. The material cost is training and maintenance: the pooled systems need \onFitsPooled{} base-network fits per replication (\onGpuHoursPooled{} GPU-hours on one A30) against three per stage for the ensemble and three in all for the masked network (\onGpuMinMasked{} GPU-minutes), which reaches stage-6 top-1 within half a point of STEP with the fewest harmful transitions. A twelve-seed ensemble, which attains STEP's accuracy (App.~\ref{app:parity}), is therefore the cheaper route to accuracy, and the masked network the simplest model, though it trails the three-seed ensemble at stages 4--5; STEP is worth its training cost only where its small NLL gain matters.

\begin{table}[t]
\centering\scriptsize\setlength{\tabcolsep}{3pt}
\caption{Accuracy and cost of the trained systems (test, $n=8{,}198$). Accuracy and harmful 5$\to$6 transitions (correct at stage 5, wrong at stage 6, \% of records) from the saved predictions behind Tables~\ref{tab:main}, \ref{tab:gbm} and~\ref{tab:flips}. Latency: batch-1 model compute on one CPU thread (Intel Xeon Silver 4314 CPU @ 2.40GHz, PyTorch 2.9.1, XGBoost 3.2.0), 200 warm-up and 2,000 test records, median / 95th percentile in ms. \emph{Cold}: every model the stage-6 request needs; \emph{incremental}: earlier-stage posteriors cached from the encounter's previous refreshes, so only the stage-6 joint model, the impressions expert and the pool run; \emph{encounter}: all six refreshes of one record, each computing only what is new. Every timed model reproduces the saved predictions (max $|\Delta\log p|\le 3\times10^{-5}$; nine base checkpoints overwritten by an earlier sensitivity run and the masked and XGBoost models, never saved, were retrained with the unchanged recipe and seeds and verified the same way). Feature preparation from prepared feature rows adds about \onPrepMs~ms and is identical across systems. Fits: models trained per replication; the pooled systems share one set of base fits. Script: \texttt{scripts/paper/revision/overnight/e3\_benchmark.py}.}
\label{tab:e3cost}
\vspace{4pt}
\begin{adjustbox}{max width=\linewidth}
\begin{tabular}{lcccrrrlrl}
\toprule
system & stage 4 & stage 5 & stage 6 & harm 5$\to$6 & stage-6 request & encounter & models / params & MB & fits\\
\midrule
Seed ensemble (stage MLP) & 48.16 / 69.60 & 52.94 / 74.53 & 58.34 / 78.56 & 5.07 & \textbf{0.44} / \textbf{0.46} & \textbf{1.9} & 3 / 1.72\,M & \textbf{6.9} & 3 per stage\\
Block-masked ensemble & 47.93 / 69.85 & 52.43 / 74.48 & 58.82 / 78.63 & \textbf{2.94} & 0.51 / 0.52 & 3.0 & 3 / 1.73\,M & \textbf{6.9} & 3\\
Tuned XGBoost & 47.23 / 68.87 & 52.53 / 74.16 & 58.09 / 79.24 & 4.39 & 27.90 / 28.37 & -- & 1 / 43,296 trees & 106 & 4-config grid per stage\\
Static pool & 48.11 / 70.36 & 53.31 / 75.35 & 58.92 / 79.74 & 3.23 & 4.12 / 4.18 & 4.6 & 39+pool / 7.6\,M & 30 & 252 MLP + 6 pool\\
\quad incremental (cached earlier stages) & & & & & \textbf{0.83} / \textbf{0.85} & & & & \\
STEP & \textbf{48.24} / \textbf{70.52} & \textbf{53.71} / \textbf{75.56} & \textbf{59.31} / \textbf{79.82} & 2.96 & 4.31 / 4.37 & 5.4 & 39+pool / 7.6\,M & 30 & 252 MLP + 6 pool\\
\quad incremental (cached earlier stages) & & & & & 0.96 / 0.98 & & & & \\
\bottomrule
\end{tabular}
\end{adjustbox}
\end{table}

\subsection{Vital-sign timing and reveal order}\label{app:timing}
\paragraph{Vital-sign timing.} The paper's vitals summarise every reading of the encounter, including readings taken after treatment began. NEMSIS records no protocol-decision time, so Table~\ref{tab:e4timing} uses an upper bound on it, fixed before any result: the earliest of the first medication and the first procedure given by EMS (entries marked as given before EMS arrival excluded) and the unit leaving the scene. Only readings timestamped at or before that minute are kept; untimed readings, and every reading of a record without a landmark, are dropped. \onEfourLandmarkPct\% of records have a landmark, \onEfourReadKeptPct\% of readings survive, \onEfourRecKeptPct\% of records keep at least one, and heart-rate coverage falls from \onEfourHrFull\% to \onEfourHrPre\%. The vitals are re-aggregated by the unchanged code, which reproduces the paper's columns from the unfiltered readings, and the stage-5 and stage-6 networks and the vitals expert are retrained with the main recipe and seeds before the pools are refitted; every other input is unchanged. Accuracy falls (stage-6 ensemble $\onEfourEnsDropOneSix$ / $\onEfourEnsDropThreeSix$ top-1 / top-3), but the increment persists: STEP minus ensemble is $\onEfourStepEnsOneFive$ / $\onEfourStepEnsThreeFive$ at stage 5 and $\onEfourStepEnsOneSix$ / $\onEfourStepEnsThreeSix$ at stage 6 (all $p\le\onEfourPmax$), against $\onEfourMainStepEnsOneFive$ / $\onEfourMainStepEnsThreeFive$ and $\onEfourMainStepEnsOneSix$ / $\onEfourMainStepEnsThreeSix$ with full-call vitals (the variants' difference is not tested), and the first-reading variant also gives positive increments at both stages, though its stage-5 top-1 increment is not significant ($p=\onCOnePOneFive$). The truncation is set by later events: the \onEfourNoLmPct\% of records without a landmark lose every reading, and the number of retained readings depends on how soon treatment began, so missing vitals partly reflect the rest of the call (the \onEfourLmNoN{} test records without a landmark are easier cases even with full-call vitals, stage-6 ensemble top-1 \onEfourNoLmFullEnsOneSix\% against \onEfourLmFullEnsOneSix\%, and \onEfourNoLmEnsOneSix\% against \onEfourLmEnsOneSix\% with pre-landmark vitals, so most of that gap is case mix). A declared diagnostic on the saved predictions restricts the comparison to the \onEfourLmN{} test records that have a landmark: the increments are $\onEfourLmOneFive$ / $\onEfourLmThreeFive$ at stage 5 and $\onEfourLmOneSix$ / $\onEfourLmThreeSix$ at stage 6 (all $p\le\onEfourLmPmax$), so they are not carried by the records without one. These tests are follow-ups on the already inspected test split, outside the corrected family of App.~\ref{app:stats}. The landmark bounds the decision from above for this definition only, and symptoms and impressions carry no timestamps, so stage 6 remains a retrospective evaluation rather than a verified pre-decision one.

\begin{table}[t]
\centering\scriptsize\setlength{\tabcolsep}{3pt}
\caption{Vital-sign timing (test, $n=8{,}198$; top-1 / top-3 \%). \emph{Full call}: the paper's features (every reading of the encounter). \emph{First reading}: each vital's earliest timestamped value. \emph{Pre-landmark}: only readings timestamped at or before the earliest of the first EMS medication, the first EMS procedure and the unit leaving the scene (an upper bound on the protocol decision, which NEMSIS does not record; \onEfourLandmarkPct\% of records have a landmark, \onEfourReadKeptPct\% of readings are kept, \onEfourRecKeptPct\% of records keep at least one). In each variant the stage-5 and stage-6 models and the vitals expert are retrained under the unchanged recipe and seeds and the pools refitted; every other input is the main-run model. Symptoms and impressions carry no timestamps. $\Delta$: STEP minus ensemble in the same variant (exact McNemar $p$). Scripts: \texttt{scripts/paper/revision/overnight/e4\_*.py}.}
\label{tab:e4timing}
\vspace{4pt}
\begin{adjustbox}{max width=\linewidth}
\begin{tabular}{llcccc}
\toprule
stage & vitals & ensemble & static pool & STEP & $\Delta$ STEP$-$ens.\ top-1 ($p$) / top-3 ($p$)\\
\midrule
5 & full call & 52.94 / 74.53 & 53.31 / 75.35 & \textbf{53.71} / \textbf{75.56} & $+0.77$ (0.0051) / $+1.02$ ($3.7\!\times\!10^{-5}$)\\
  & first reading & 52.13 / 73.86 & 52.55 / 74.92 & \textbf{52.63} / \textbf{75.10} & $+0.50$ (0.065) / $+1.24$ ($9.4\!\times\!10^{-7}$)\\
  & pre-landmark & 49.34 / 71.02 & 50.60 / 72.40 & \textbf{50.90} / \textbf{72.79} & $+1.56$ ($5.7\!\times\!10^{-8}$) / $+1.77$ ($5.4\!\times\!10^{-11}$)\\
\midrule
6 & full call & 58.34 / 78.56 & 58.92 / 79.74 & \textbf{59.31} / \textbf{79.82} & $+0.96$ ($7.7\!\times\!10^{-4}$) / $+1.27$ ($6.8\!\times\!10^{-7}$)\\
  & first reading & 58.22 / 78.42 & 58.78 / 79.30 & \textbf{59.03} / \textbf{79.40} & $+0.81$ (0.0047) / $+0.98$ ($8.1\!\times\!10^{-5}$)\\
  & pre-landmark & 55.72 / 76.62 & 56.71 / 77.52 & \textbf{57.14} / \textbf{77.75} & $+1.41$ ($1.1\!\times\!10^{-6}$) / $+1.13$ ($1.8\!\times\!10^{-5}$)\\
\bottomrule
\end{tabular}
\end{adjustbox}
\end{table}

\paragraph{Reveal order.}\label{app:order} The six-stage order is an evaluation assumption. Table~\ref{tab:e6order} evaluates two alternatives fixed before any result: \emph{A} reveals the complaint organ system before its location, and \emph{B}, explicitly hypothetical, reveals vital signs with the scene context before symptoms. Dispatch stays first and the final feature inventory is unchanged; each new intermediate dataset is built from the cohort and its networks retrained, and every pool whose history changes is refitted. Accuracy at a changed stage depends on the schedule (stage-2 STEP $\onEsixATwoOne$ / $\onEsixATwoThree$ top-1 / top-3 under A, stage-4 STEP $\onEsixBFourOne$ / $\onEsixBFourThree$ under B, against the paper's order), but the pooling conclusions do not: increments at stages 1--3 stay within \onEsixEarlyMaxAbs{} points, the stage-4--6 top-3 increment stays between $\onEsixThreeIncMin$ and $\onEsixThreeIncMax$ ($p\le\onEsixThreePmax$), and stage-6 STEP differs from the paper's order by at most \onEsixSixMaxAbs{} points ($p\ge\onEsixSixPmin$), as expected because only the pools' histories differ at stage 6. Averaged over the six stages, a descriptive summary rather than elapsed-time performance, STEP changes by $\onEsixAMeanOne$ / $\onEsixAMeanThree$ under A and $\onEsixBMeanOne$ / $\onEsixBMeanThree$ under B. These tests are follow-ups on the already inspected test split, outside the corrected family. This measures sensitivity to the evaluation schedule; it does not identify a better clinical acquisition order.

\begin{table}[t]
\centering\scriptsize\setlength{\tabcolsep}{3pt}
\caption{Reveal-order sensitivity (test, $n=8{,}198$; top-1 / top-3 \%). Two schedules fixed before any result: \emph{A} reveals the complaint organ system before the complaint location (stage 2$^A$); \emph{B}, explicitly hypothetical, reveals vitals before symptoms (stage 4$^B$: dispatch, location, organ system, vitals and scene context). Each new intermediate dataset is built from the cohort and its models retrained under the unchanged recipe and seeds; every pool whose history changed is refitted. Stages not listed are identical to the paper's order; stage 6 has the same features in every schedule, so only the pools' histories differ there. $\Delta$: STEP minus ensemble within the schedule; last column: STEP in the schedule minus STEP in the paper's order on the same records (exact McNemar $p$ throughout). Scripts: \texttt{scripts/paper/revision/overnight/e6\_build.py}, \texttt{fit\_pools\_variant.py}, \texttt{e46\_compare.py}.}
\label{tab:e6order}
\vspace{4pt}
\begin{adjustbox}{max width=\linewidth}
\begin{tabular}{llccccc}
\toprule
schedule & stage & ensemble & static pool & STEP & $\Delta$ STEP$-$ens.\ top-1 ($p$) / top-3 ($p$) & STEP vs paper order top-1 ($p$) / top-3 ($p$)\\
\midrule
A & 2$^A$ & 25.10 / 45.58 & \textbf{25.37} / \textbf{46.13} & 25.35 / 46.00 & $+0.24$ (0.17) / $+0.41$ (0.057) & $+0.72$ (0.038) / $+2.09$ ($2.1\!\times\!10^{-7}$)\\
A & 3 & 27.56 / 48.16 & 27.70 / 48.51 & \textbf{27.81} / \textbf{48.63} & $+0.26$ (0.14) / $+0.48$ (0.032) & $-0.04$ (0.78) / $+0.24$ (0.054)\\
A & 4 & 48.16 / 69.60 & 48.18 / 70.43 & \textbf{48.40} / \textbf{70.54} & $+0.24$ (0.42) / $+0.94$ ($4.0\!\times\!10^{-4}$) & $+0.16$ (0.066) / $+0.02$ (0.88)\\
A & 5 & 52.94 / 74.53 & 53.29 / 75.34 & \textbf{53.67} / \textbf{75.47} & $+0.73$ (0.0081) / $+0.94$ ($1.7\!\times\!10^{-4}$) & $-0.04$ (0.74) / $-0.09$ (0.28)\\
A & 6 & 58.34 / 78.56 & 58.94 / \textbf{79.78} & \textbf{59.25} / \textbf{79.78} & $+0.90$ (0.0016) / $+1.22$ ($1.6\!\times\!10^{-6}$) & $-0.06$ (0.5) / $-0.05$ (0.52)\\
\midrule
B & 4$^B$ & 43.45 / 65.82 & 43.77 / 66.39 & \textbf{43.95} / \textbf{66.63} & $+0.50$ (0.048) / $+0.81$ ($7.5\!\times\!10^{-4}$) & $-4.29$ ($6.5\!\times\!10^{-15}$) / $-3.89$ ($1.6\!\times\!10^{-14}$)\\
B & 5 & 52.94 / 74.53 & 53.26 / 75.31 & \textbf{53.40} / \textbf{75.43} & $+0.46$ (0.1) / $+0.90$ ($3.9\!\times\!10^{-4}$) & $-0.30$ (0.21) / $-0.12$ (0.61)\\
B & 6 & 58.34 / 78.56 & 59.03 / 79.69 & \textbf{59.27} / \textbf{79.78} & $+0.93$ (0.0019) / $+1.22$ ($1.9\!\times\!10^{-6}$) & $-0.04$ (0.9) / $-0.05$ (0.82)\\
\bottomrule
\end{tabular}
\end{adjustbox}
\end{table}

\section{DPCS: Comparison with Confidence Scores, All Six Stages and Class-Level Trajectories}
\label{app:dpcs}
This appendix details the formulation, empirical evaluation, and research implications of the Dispatch--Protocol Confidence Score (DPCS, \S\ref{sec:dpcs} and \S\ref{sec:dpcsres}). The empirical rates $a_{j,t}$ (stage-$t$ validation protocol recall) and $b_{i,t}$ (stage-$t$ validation top-1 accuracy conditioned on dispatch complaint code $i$) in Equation~\ref{eq:ab} are estimated strictly on the 7,378 inner-validation rows using STEP's stage-$t$ predictions from the fit-rows base models (\path{scripts/paper/review2/step_val_preds.py} refits each pool and verifies exact reproduction of the saved test posteriors). The score $\operatorname{DPCS}_e^{(t)}=b_{I_e,t}\max_{j\in\mathcal{K}_e^{(t)}}a_{j,t}$ and discrete tier $\ell_e^{(t)}$ in Equation~\ref{eq:dpcs} are then evaluated on the 8,198 held-out test records. By definition, DPCS is a stage-specific contextual reliability summary derived from validation hit rates; it is not a direct measure of whether the dispatcher's initial code was clinically correct, a calibrated probability of shortlist correctness, or a statistical lower confidence bound.

\begin{figure}[t]
\centering
\includegraphics[width=\textwidth]{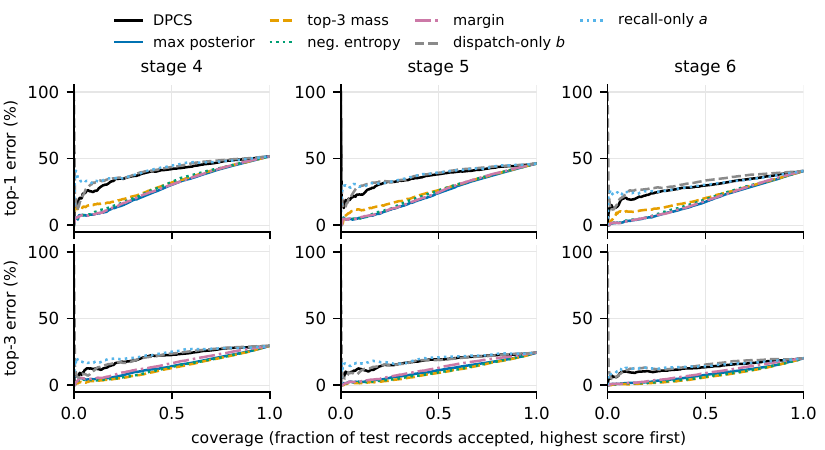}
\caption{Risk--coverage curves at stages 4--6 on the test split: the top-1 (top row) and top-3 (bottom row) error rate among the records accepted when records are admitted in decreasing order of each score. A lower curve is better; the area under it is the AURC of Table~\ref{tab:dpcsscores}.}
\label{fig:dpcsscores}
\end{figure}

\begin{figure}[h]
\centering
\includegraphics[width=0.72\textwidth]{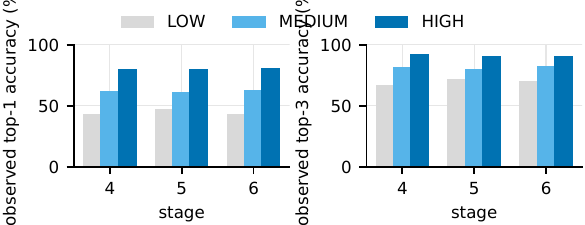}
\caption{Observed top-1 (left) and top-3 (right) accuracy of the displayed STEP shortlist within each DPCS tier at the three stages where all three tiers are populated (tier shares: Fig.~\ref{fig:dpcsmethod}b). Every bucket drawn holds at least \dpcsMinBucket{} test records; the rates the score multiplies are estimated on the inner-validation rows.}
\label{fig:dpcs}
\end{figure}

\paragraph{Empirical stratification versus posterior error-detection superiority.} Table~\ref{tab:dpcs} and Figure~\ref{fig:dpcs} show that DPCS tiers meaningfully stratify observed test accuracy as evidence accumulates. From Stage 4 onward, where all three tiers have at least \dpcsMinBucket{} test encounters, observed accuracy rises monotonically from \textsc{low} to \textsc{medium} to \textsc{high} on both top-1 (Stage 6: \dpcsLowTopOneSix\%, \dpcsMedTopOneSix\%, \dpcsHighTopOneSix\%) and top-3 (Stage 6: \dpcsLowTopThreeSix\%, \dpcsMedTopThreeSix\%, \dpcsHighTopThreeSix\%). To evaluate offline error detection directly, Table~\ref{tab:dpcsscores} compares DPCS at Stages 4--6 against four summary statistics of STEP's own posterior distribution on the identical test encounters (maximum posterior probability, top-1/top-2 margin, negative entropy, and total probability mass on the top-3 classes) as well as the two individual DPCS components ($b_{I_e,t}$ and $\max_{j\in\mathcal{K}_e^{(t)}}a_{j,t}$), using the area under the risk--coverage curve (AURC; Figure~\ref{fig:dpcsscores}) and the AUROC for detecting top-1 and top-3 shortlist errors. While multiplying $a_{j,t}$ and $b_{i,t}$ outperforms either factor alone at Stages 4--6, every posterior confidence summary achieves stronger error-detection discrimination than DPCS: at Stage 6, maximum posterior probability reaches AUROC \dpcsAurocMaxPostSix{} (AURC 0.187) versus \dpcsAurocSix{} (AURC 0.293) for DPCS on top-1 errors, and top-3 posterior mass reaches AUROC \dpcsAurocMassThreeSix{} (AURC 0.071) versus \dpcsAurocThreeSix{} (AURC 0.136) on top-3 errors.

\begin{table}[t]
\centering
\scriptsize
\setlength{\tabcolsep}{3pt}
\caption{DPCS against simple confidence scores from the same STEP posterior on the test split at stages 4--6. AURC: area under the risk--coverage curve (lower is better; records accepted in decreasing score order; Fig.~\ref{fig:dpcsscores}). AUROC: detection of a wrong top-1 (top-3) shortlist (higher is better). $\Delta$AUROC: score minus DPCS with a paired bootstrap 95\% CI over records (500 resamples); a positive value whose interval excludes zero means the simpler score is better. The paired AURC differences (in \texttt{dpcs\_r2.json}) have the same sign at every stage. Dispatch-only $b$ and recall-only $a$ are the two DPCS factors alone.}
\label{tab:dpcsscores}
\vspace{4pt}
\begin{tabular}{lrrlrrl}
\toprule
& \multicolumn{3}{c}{top-1 error} & \multicolumn{3}{c}{top-3 error}\\
\cmidrule(lr){2-4}\cmidrule(lr){5-7}
score & AURC & AUROC & $\Delta$AUROC vs DPCS [95\% CI] & AURC & AUROC & $\Delta$AUROC vs DPCS [95\% CI]\\
\midrule
\multicolumn{7}{l}{\emph{stage 4}}\\
DPCS & 0.402 & 0.633 & -- & 0.216 & 0.607 & --\\
max posterior & \textbf{0.290} & \textbf{0.789} & \textbf{+0.156 [+0.143, +0.168]} & 0.146 & 0.751 & +0.144 [+0.130, +0.157]\\
top-3 mass & 0.315 & 0.763 & +0.130 [+0.117, +0.144] & \textbf{0.134} & \textbf{0.772} & \textbf{+0.165 [+0.151, +0.178]}\\
neg. entropy & 0.303 & 0.765 & +0.132 [+0.117, +0.146] & 0.139 & 0.763 & +0.156 [+0.141, +0.170]\\
margin & 0.295 & 0.779 & +0.146 [+0.133, +0.159] & 0.163 & 0.706 & +0.099 [+0.085, +0.112]\\
dispatch-only $b$ & 0.418 & 0.608 & -0.024 [-0.032, -0.017] & 0.220 & 0.594 & -0.013 [-0.022, -0.004]\\
recall-only $a$ & 0.428 & 0.610 & -0.023 [-0.029, -0.017] & 0.239 & 0.579 & -0.028 [-0.036, -0.022]\\
\multicolumn{7}{l}{\emph{stage 5}}\\
DPCS & 0.364 & 0.614 & -- & 0.182 & 0.600 & --\\
max posterior & \textbf{0.241} & \textbf{0.790} & \textbf{+0.176 [+0.163, +0.189]} & 0.114 & 0.750 & +0.150 [+0.135, +0.165]\\
top-3 mass & 0.265 & 0.763 & +0.148 [+0.135, +0.162] & \textbf{0.101} & \textbf{0.779} & \textbf{+0.179 [+0.164, +0.193]}\\
neg. entropy & 0.251 & 0.772 & +0.157 [+0.144, +0.171] & 0.105 & 0.771 & +0.171 [+0.156, +0.186]\\
margin & 0.246 & 0.777 & +0.162 [+0.148, +0.176] & 0.128 & 0.704 & +0.103 [+0.087, +0.120]\\
dispatch-only $b$ & 0.378 & 0.595 & -0.020 [-0.029, -0.010] & 0.182 & 0.599 & -0.001 [-0.012, +0.010]\\
recall-only $a$ & 0.383 & 0.595 & -0.020 [-0.027, -0.013] & 0.201 & 0.573 & -0.027 [-0.036, -0.019]\\
\multicolumn{7}{l}{\emph{stage 6}}\\
DPCS & 0.293 & 0.650 & -- & 0.136 & 0.637 & --\\
max posterior & \textbf{0.187} & \textbf{0.809} & \textbf{+0.159 [+0.146, +0.171]} & 0.082 & 0.775 & +0.138 [+0.122, +0.156]\\
top-3 mass & 0.216 & 0.775 & +0.125 [+0.111, +0.138] & \textbf{0.071} & \textbf{0.805} & \textbf{+0.169 [+0.153, +0.186]}\\
neg. entropy & 0.197 & 0.790 & +0.139 [+0.126, +0.153] & 0.074 & 0.797 & +0.160 [+0.144, +0.178]\\
margin & 0.191 & 0.798 & +0.148 [+0.135, +0.161] & 0.093 & 0.729 & +0.092 [+0.076, +0.110]\\
dispatch-only $b$ & 0.323 & 0.600 & -0.050 [-0.058, -0.041] & 0.152 & 0.592 & -0.045 [-0.056, -0.035]\\
recall-only $a$ & 0.307 & 0.646 & -0.004 [-0.010, +0.002] & 0.149 & 0.618 & -0.019 [-0.026, -0.011]\\
\bottomrule
\end{tabular}
\end{table}

\paragraph{Conditional information analysis.} Because standalone AUROC comparisons do not test whether contextual dispatch--protocol rates provide complementary signal when combined with posterior confidence, Table~\ref{tab:e5dpcs} evaluates a leakage-controlled conditional risk model on the test split. The 7,378 inner-validation rows are partitioned into three disjoint subsets: one third ($n=2{,}459$) selects STEP's checkpoint epoch and temperature, the second third ($n=2{,}459$) estimates the DPCS rates $a_{j,t}$ and $b_{i,t}$, and the final third ($n=2{,}460$) fits logistic regression models predicting whether STEP's top-1 or top-3 shortlist misses the recorded protocol using posterior confidence alone ($C$), a four-feature posterior summary ($S$: maximum probability, margin, entropy, and top-3 mass), or each augmented with DPCS ($C{+}D$ and $S{+}D$). Across Stages 4--6 and both error targets (12 primary comparisons), adding DPCS changes held-out test log loss by at most \onEfiveMaxAbsDll{} nats; all \onEfiveNcis{} of \onEfiveNtests{} paired bootstrap 95\% confidence intervals include zero (Holm-corrected $p\ge\onEfiveMinHolm$), and test AUROC changes by at most \onEfiveMaxDauc. Because estimating $a_{j,t}$ and $b_{i,t}$ on one third of the inner-validation split introduces slightly higher sampling variance than using all 7,378 inner-validation rows, this conclusion is scoped to the evaluated unsmoothed empirical estimators and recorded-label error detection on this benchmark rather than claiming a universal impossibility result.

\begin{table}[t]
\centering\scriptsize\setlength{\tabcolsep}{3pt}
\caption{Does DPCS add information beyond the posterior? Held-out (test, $n=8{,}198$) log loss of a logistic model predicting whether STEP's top-1 (or top-3) misses the recorded protocol, from confidence alone (C: maximum posterior, or top-3 mass for top-3), a richer posterior summary (S: maximum, margin, entropy, top-3 mass), each with DPCS added (+D). Dependencies are disjoint: the inner-validation rows are split into thirds for STEP's epoch selection and temperature (P, 2,459), DPCS rate estimation (R, 2,459) and risk-model fitting (F, 2,460); test is scored once. $\Delta$: augmented minus base (negative favours DPCS), paired bootstrap 95\% interval over test records; $p_H$: Holm over the 12 primary comparisons. AUROC of the risk model. Script: \texttt{scripts/paper/revision/overnight/e5\_dpcs\_incremental.py}.}
\label{tab:e5dpcs}
\vspace{4pt}
\begin{adjustbox}{max width=\linewidth}
\begin{tabular}{llrrrrrrrr}
\toprule
stage & target & error \% & LL C & $\Delta$ C+D [CI] & LL S & $\Delta$ S+D [CI] & $p_H$ (C, S) & AUROC C / C+D & AUROC S / S+D\\
\midrule
4 & top-1 & 51.6 & \textbf{0.5539} & $+0.0004$ [$-0.0008$, $+0.0016$] & 0.5548 & $+0.0002$ [$-0.0011$, $+0.0013$] & 1.00, 1.00 & \textbf{0.789} / 0.788 & \textbf{0.789} / 0.788\\
4 & top-3 & 29.5 & \textbf{0.5047} & $+0.0001$ [$-0.0004$, $+0.0006$] & 0.5053 & $-0.0000$ [$-0.0006$, $+0.0005$] & 1.00, 1.00 & \textbf{0.772} / \textbf{0.772} & \textbf{0.772} / \textbf{0.772}\\
5 & top-1 & 46.5 & \textbf{0.5452} & $-0.0001$ [$-0.0002$, $+0.0000$] & 0.5459 & $-0.0001$ [$-0.0002$, $+0.0001$] & 1.00, 1.00 & \textbf{0.795} / \textbf{0.795} & \textbf{0.794} / \textbf{0.794}\\
5 & top-3 & 24.5 & \textbf{0.4590} & $-0.0001$ [$-0.0003$, $+0.0002$] & 0.4593 & $-0.0001$ [$-0.0003$, $+0.0001$] & 1.00, 1.00 & \textbf{0.781} / \textbf{0.781} & \textbf{0.781} / \textbf{0.781}\\
6 & top-1 & 40.8 & \textbf{0.5187} & $-0.0003$ [$-0.0015$, $+0.0009$] & \textbf{0.5187} & $-0.0004$ [$-0.0015$, $+0.0008$] & 1.00, 1.00 & \textbf{0.810} / \textbf{0.810} & \textbf{0.810} / \textbf{0.810}\\
6 & top-3 & 20.1 & \textbf{0.4019} & $-0.0004$ [$-0.0011$, $+0.0003$] & 0.4029 & $-0.0005$ [$-0.0010$, $+0.0001$] & 1.00, 0.97 & 0.803 / \textbf{0.804} & 0.803 / \textbf{0.804}\\
\bottomrule
\end{tabular}
\end{adjustbox}
\end{table}

\paragraph{Operational conservatism and stage-wise tier distributions.} Table~\ref{tab:dpcs} reports the exact test-set counts and observed accuracies across all six stages (visualized in Figure~\ref{fig:dpcsmethod}b and Figure~\ref{fig:dpcs}). Under the fixed thresholds of 0.35 and 0.60, DPCS is operationally conservative at early stages: at Stages 1--3, when dispatch and complaint metadata alone yield 20.72\%--27.85\% overall top-1 accuracy, at least \dpcsLowShareBefore\% of test encounters are assigned to the \textsc{low} tier (7,706 to 7,712 of 8,198 records) and at most \dpcsHighMaxBefore{} encounters reach \textsc{high}, although \textsc{medium} already orders above \textsc{low} (58.64\%--64.94\% vs.\ 17.95\%--25.34\% top-1). Crucially, DPCS does not label most encounters low confidence throughout the entire call: as clinical evidence accumulates at Stages 4--6, the tier distribution shifts upward so that by Stage 6, the majority of test encounters (5,303 of 8,198, or 64.7\%) reside in the \textsc{medium} tier, alongside 2,039 (24.9\%) in \textsc{low} and 856 (10.4\%) in \textsc{high}. At the same time, multiplying two empirical rates in $[0,1]$ mechanically produces smaller numerical values than single probabilities without guaranteeing better calibration, clinical safety, or error detection. Because posterior confidence thresholds can also be chosen to produce cautious early-stage displays (and posterior probabilities can be binned directly into the same three tiers), lower numerical scores alone do not establish a unique operational advantage for DPCS.

\begin{table}[t]
\centering
\footnotesize
\setlength{\tabcolsep}{3pt}
\caption{DPCS tier distribution and observed reliability of the displayed STEP top-3 list, by stage. $a_{j,t}$ and $b_{i,t}$ are estimated from STEP's predictions on the 7,378 inner-validation rows (unsmoothed empirical rates); DPCS is then computed for every test record ($K{=}3$) and bucketed at the fixed, unfitted cut points 0.35/0.60. Counts are shown with every accuracy: buckets with $n<100$ are parenthesised and marked $\S$ and are never cited in the text. At stages 1--3 at least 94\% of records are LOW and at most 108 are HIGH; from stage 4 every tier holds at least 298 records.}
\label{tab:dpcs}
\vspace{4pt}
\begin{tabular}{crrrrrrrrr}
\toprule
stage & \multicolumn{3}{c}{LOW ($<0.35$)} & \multicolumn{3}{c}{MEDIUM ($0.35$--$0.60$)} & \multicolumn{3}{c}{HIGH ($\geq0.60$)}\\ & $n$ & top-1 & top-3 & $n$ & top-1 & top-3 & $n$ & top-1 & top-3\\
\midrule
1 & 7,706 & 17.95 & 36.10 & 482 & 64.94 & 73.86 & 10 & (30.00)$^{\S}$ & (50.00)$^{\S}$\\
2 & 7,708 & 22.06 & 41.93 & 382 & 58.64 & 70.68 & 108 & 87.96 & 90.74\\
3 & 7,712 & 25.34 & 46.47 & 386 & 62.69 & 75.65 & 100 & 87.00 & 91.00\\
4 & 5,962 & 42.33 & 66.12 & 1,938 & 61.56 & 80.75 & 298 & 79.87 & 91.95\\
5 & 4,615 & 46.48 & 71.35 & 3,183 & 60.92 & 79.77 & 400 & 79.75 & 90.50\\
6 & 2,039 & 42.82 & 69.69 & 5,303 & 62.21 & 82.05 & 856 & 80.61 & 90.19\\
\bottomrule
\end{tabular}
\end{table}

\paragraph{Dispatch support, estimation granularity, and candidate hypotheses.} While the primary cohort is balanced across the 82 protocols (approximately 90 inner-validation records per protocol for $a_{j,t}$), dispatch complaint codes follow their natural empirical distribution across those encounters. In the 7,378 inner-validation rows, the median support per dispatch code is \dpcsSupportMedian{} encounters, and only \dpcsSupportLtThirty\% of test encounters carry a dispatch code with fewer than 30 inner-validation records. Consequently, while unsmoothed rate estimation in rare dispatch cells may introduce noise for a small fraction of encounters, sample scarcity alone cannot be asserted to explain the overall AUROC gap relative to posterior confidence without a dedicated support-stratified ablation. Other plausible mechanisms include: (i) coarse marginal aggregation within broad dispatch complaints (such as ``Sick Person'' or ``Breathing Problem''), which assigns the same $b_{i,t}$ to encounters with very different symptom and vital-sign clarity; (ii) the fixed multiplicative functional form $b_{I_e,t}\max_{j\in\mathcal{K}_e^{(t)}}a_{j,t}$; and (iii) the possibility that instance-level posterior entropy and margin already internalize the difficulty of the dispatch and protocol context. These explanations remain hypotheses for future investigation, and increasing validation sample size cannot be assumed to make marginal dispatch--protocol summaries outperform instance-specific posteriors.

\paragraph{Offline discrimination, human factors, and future research directions.} Offline error-detection metrics (AUROC and AURC) and human-centered clinical utility address distinct evaluation questions: offline metrics establish how accurately a score ranks error risk, whereas how practicing EMTs interpret and act upon reliability displays requires prospective human-factors evaluation. Importantly, lower displayed confidence cannot be assumed to automatically improve clinical trust or safety, as persistently cautious tiers could also induce alert fatigue or lead users to ignore the annotation. A rigorous future user study should compare four controlled interface conditions under matched simulated or silent-shadowing scenarios: (1) shortlist display without reliability annotation, (2) posterior-confidence tier annotation, (3) DPCS contextual tier annotation, and, if warranted, (4) a combined display pairing instance-level posterior confidence with contextual dispatch--protocol base rates, measuring appropriate reliance, decision time, and cognitive workload. Methodologically, concrete future extensions to DPCS include empirical Bayes or hierarchical shrinkage across related dispatch categories, support-aware credible intervals around $(a_{j,t}, b_{i,t})$, and conditioning reliability estimates on finer-grained clinical subgroups; evaluating actual dispatch code correctness (as distinct from model prediction accuracy conditioned on the recorded dispatch code) would require independent clinical adjudication.

\begin{figure}[h]
\centering
\includegraphics[width=\textwidth]{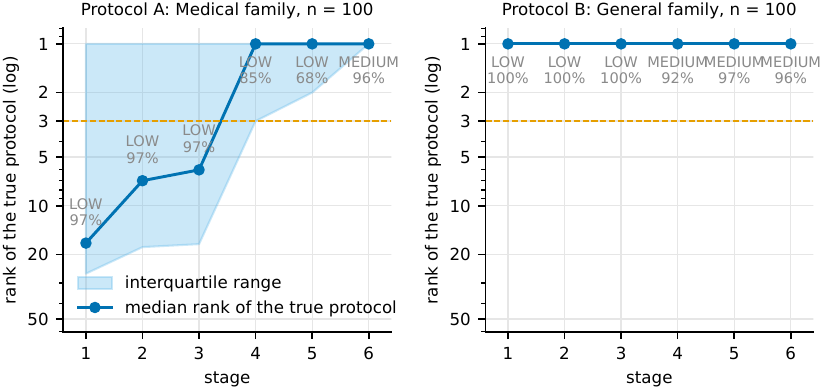}
\caption{Class-level rank trajectories with DPCS tiers for a late-converging protocol (A, Medical family) and an early-converging one (B, General family); the dashed line is the top-3 boundary and the annotation is the modal tier with its share. Aggregates over each class's test records; no individual record is drawn.}
\label{fig:traces}
\end{figure}

\paragraph{Class-level trajectories.} Figure~\ref{fig:traces} shows two class-level rank trajectories under STEP: for each protocol, the median and interquartile range of the true protocol's rank across that class's test records at each stage, annotated with the modal DPCS tier and its share. Protocol A converges late (its median rank enters the shortlist only at the scene stage) and its modal tier rises from \textsc{low} to \textsc{medium} only after impressions; Protocol B is ranked first from dispatch on and its tier rises at the scene stage, when the dispatch-level reliability $b_{I_e,t}$ and the recognisability $a_{j,t}$ both improve. No individual record is drawn.

\section{Statistical Protocol and Multiplicity}
\label{app:stats}
\paragraph{The paired test.} Every comparison in the paper is an exact, two-sided McNemar test on per-record hit vectors. For control and candidate hit vectors $h^c,h^d\in\{0,1\}^n$ on the same records, let $n_{10}=\#\{h^c=1,h^d=0\}$ and $n_{01}=\#\{h^c=0,h^d=1\}$; the $p$-value is the two-sided exact binomial probability of $n_{01}$ successes in $n_{10}+n_{01}$ trials at rate $\frac12$ (1 if $n_{10}+n_{01}=0$). It is computed twice, with the hit defined by $h_1$ and by $h_3$, on the same records; the top-3 hit set contains the top-1 hit set, so the two tests are strongly positively dependent.

\paragraph{Test families of the reported results.} Two families are tested on the test split and corrected separately. (i) The \nstepfamily{} comparisons of STEP against the three-seed ensemble (six stages, two metrics), Bonferroni level $0.05/\nstepfamily$; \nFamilyASurvive{} survive: top-3 at stages 4--6 and top-1 at stage 6 (Table~\ref{tab:main}). (ii) The \nllmfamily{} prompted-run comparisons against STEP (four prompted runs, six stages, two metrics), level $0.05/\nllmfamily$, all of which survive. Validation deltas are reported, not tested against a threshold, and the controls of Table~\ref{tab:controls} and the diagnostics of App.~\ref{app:diag} are descriptive. STEP's pooling rule is applied at all six stages regardless of any test.

\paragraph{Historical (pre-repair) acceptance test.} Before the data repairs and the revised pooling rule, a component entered the pipeline only when a paired validation test accepted it: a candidate was accepted when either metric improved at $p<0.025$ and neither was harmed at $p<0.05$ (the accept level is the Bonferroni split of $0.05$ across the two metrics; the harm level stays at $0.05$ because making harm harder to detect would be the wrong trade for a safety-critical display). A post-hoc finding also had to clear a family-wise correction over its wave (for the modality-fusion wave, $0.05/72=6.9{\times}10^{-4}$); a finding that cleared the per-candidate bar but not the family-wise one was not adopted. The rule documents how the feature set, the pool's form and SEP's configuration were chosen; it no longer decides the stages at which STEP pools. Table~\ref{tab:auditsummary} summarises every such comparison with a machine-readable record, \ngates\ in total, of which \ngatesselected\ were accepted, and Table~\ref{tab:design} lists the design decisions the tests settled. Both tables are historical: they were run on the pre-repair feature family. Under that rule the combined pool was rejected at stages 1--3 on validation, accepted at stage 4 on top-3 alone but not admitted (it failed its wave's family-wise bar, and its verdict moved with the gate's initialisation across eight seeds, whereas the stage-5 and stage-6 pools were accepted under every initialisation), and admitted at stages 5--6. Pooling the experts without the joint posterior lost to the ensemble on validation at every stage from 2 on. The revised pipeline uses the seed-0 initialisation at every stage.

\begin{table}[h]
\caption{Historical record (pre-repair feature family; App.~\ref{app:stats}). Every acceptance-test comparison in the project with a machine-readable record, by wave: 238 paired comparisons, 84 accepted. The split is the one the test was run on; the last column says whether the wave's accepted candidates entered the pipeline before this revision; the revised STEP pipeline pools at every stage regardless of these verdicts, and its own test comparisons are counted separately (App.~\ref{app:stats}). The full per-comparison ledger (deltas and $p$-values for every pair) is shipped with the code.}
\centering
\scriptsize
\setlength{\tabcolsep}{3pt}
\begin{tabular}{p{4.9cm}lrrrp{2.9cm}}
\toprule
wave & split & pairs & accepted & rejected & entered the historical system\\
\midrule
Task 2: modality fusion & inner-val & 72 & 25 & 47 & yes: pool at stages 5--6 (now all stages)\\
Temporal pool, pre-repair data & inner-val (pre-repair data) & 14 & 9 & 5 & no\\
Temporal pool, config D, stages 5--6 & inner-val & 14 & 8 & 6 & no (stage axis alone)\\
Temporal pool, config D, stages 1--4 & inner-val & 28 & 3 & 25 & no\\
Task 1: temporal pool with context, stages 2--4 & inner-val & 15 & 3 & 12 & no\\
Task 1: temporal pool with context, stages 5--6 & inner-val & 14 & 10 & 4 & no (stage axis alone)\\
Context, uniform placement & inner-val (superseded) & 6 & 6 & 0 & no\\
Task 0c: context at audited capture time & inner-val & 6 & 6 & 0 & yes: context block\\
Config-D ladder, acceptance-test re-check & inner-val & 6 & 6 & 0 & yes: feature set\\
Task 1: vitals sequence with context & inner-val & 2 & 1 & 1 & no\\
Architectures on config D, held-out split & held-out partition val & 24 & 1 & 23 & no (baselines)\\
LLM arm vs NN 3-seed ensemble, test8198 & test8198 & 24 & 0 & 24 & no (comparison)\\
LLM within-arm, zero-shot $\to$ k=3, test8198 & test8198 & 6 & 5 & 1 & no (comparison)\\
LLM $k$ selection on val820, pooled $n=14{,}760$ & val820 (pooled) & 2 & 1 & 1 & yes: $k{=}3$ for the LLM arm\\
Task 1: adopt-vs-production & inner-val & 4 & 0 & 4 & no\\
Provenance split & inner-val & 1 & 0 & 1 & no\\
\midrule \textbf{all waves} & & \textbf{238} & \textbf{84} & \textbf{154} & \\
\bottomrule
\end{tabular}

\label{tab:auditsummary}
\end{table}

\begin{table}[t]
\caption{Why each design choice was made (historical acceptance-test decisions; STEP now pools at every stage, App.~\ref{app:stats}). Effects are $\Delta$top-1 / $\Delta$top-3 in pp with the exact paired McNemar $p$; inner-validation acceptance test unless a split is stated. Where the driver is top-3 rather than top-1, the verdict says so.}
\centering
\scriptsize
\setlength{\tabcolsep}{2.5pt}
\begin{adjustbox}{max width=\linewidth}
\begin{tabular}{p{2.7cm}p{2.9cm}p{4.9cm}p{3.0cm}}
\toprule
decision & alternative(s) & effect ($p$) & verdict\\
\midrule
gated log-linear pool & ungated pool; residual MLP & gated $+0.62$ / $+0.92$ (0.020 / $8.2\!\times\!10^{-5}$); residual $-0.19$ / $-0.04$ & gated selected (top-3 leg); residual rejected 20/20 cells\\
joint posterior last (identity init) & experts only & $-3.27$ / $-2.07$ ($1.0\!\times\!10^{-13}$ / $5.1\!\times\!10^{-8}$) & experts-only rejected (harm); experts added to joint selected\\
3-seed geometric mean & single seed & val $+2.91$ / $+2.32$; test $+2.06$ / $+2.31$ ($3.1\!\times\!10^{-10}$ / $8.7\!\times\!10^{-16}$) & selected\\
combined pool at stage 4 & 3-seed ensemble & $+0.49$ / $+0.75$ (0.109 / 0.010) & historically not admitted (family-wise bar); applied under the revised policy (App.~\ref{app:step})\\
last-epoch checkpoint & best-validation epoch & top-1 $-0.26$ / $-0.53$ (stages 5 / 6; 0.445 / 0.120) & rejected\\
MLP & LSTM; transformer; two cascades (stage 6, held-out split) & $-1.35$ / $-0.79$; $-1.49$ / $-1.14$; $-3.21$; $-2.91$ (all $p<0.01$) & all rejected with harm at stages 5--6\\
merged clinical vocabulary & provenance split (3,801 features) & $-1.06$ / $-0.58$ (0.005 / 0.090) & rejected (harmed top-1)\\
config-D features: explicit missingness; extended vitals; timed context & plain \texttt{zero\_mask}; without; none & stage 5: $+2.52$ / $+1.83$; $+5.49$ / $+3.77$; stage 4 context $+5.76$ / $+4.73$ (all $p<10^{-4}$) & selected (stage-6 encoding step on top-3 only)\\
LLM $k{=}3$ & zero-shot; $k{=}5$ (val820) & $k{=}3$ {\bfseries\boldmath $+2.20$} / {\bfseries\boldmath $+1.80$}; $k{=}5$ $+1.76$ / $-1.21$ ($9.7\!\times\!10^{-5}$) & $k{=}5$ rejected: harmed top-3\\
acceptance test, $\alpha_{\text{accept}}=0.025$, $\alpha_{\text{harm}}=0.05$ & single top-1 test & reversed four decisions & design rule (harm leg screens for significant validation harm)\\
\bottomrule
\end{tabular}
\end{adjustbox}

\label{tab:design}
\end{table}

\section{What Earlier Posteriors Carry: Diagnosis, Alternatives, Distribution Shift and Generality}
\label{app:diag}
All measurements of this appendix are on the repaired family under the boundaries of App.~\ref{app:stepfit}; the diagnosis uses development data only (OOF posteriors of the 66,402 fit rows and the fit-rows models' predictions on the 7,378 inner-validation rows), and test is read only for the final, frozen comparisons in the tables.

\paragraph{Transitions.} At every transition the mean rises while a minority of records fall (Figure~\ref{fig:rankflow} shows the flows on the test records): 2.7\% to 5.8\% of the development records are ranked correctly by the earlier ensemble and incorrectly by the later one (transitions $1\to2$ to $5\to6$: 4.3, 2.7, 4.8, 5.8, 5.3\%), against 5.5\% to 25.5\% that move the other way, and the true-class probability decreases on 31\% to 43\% of records. For 71\% to 82\% of the harmful flips the earlier top-1 is still inside the later top-3, which is why the top-3 increment of the pool is the more robust one. The stage models are strongly dependent (hit correlation between adjacent stages 0.42 to 0.78), yet 10.7\% and 11.7\% of the stage-5 and stage-6 errors are records some earlier stage ranked correctly and 13.3\% and 13.7\% are records some block expert ranked correctly (oracle upper bounds, never used for selection). Missingness of the new block is not the driver: at $5\to6$ the harm rate on records with an empty impression block (5.2\%) equals the overall rate (5.3\%).

\begin{figure}[t]
\centering
\includegraphics[width=\textwidth]{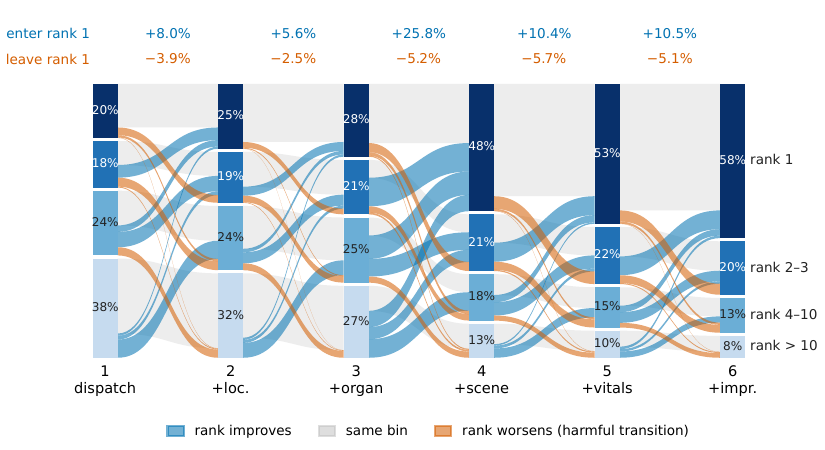}
\caption{What the earlier posteriors carry. Rank of the true protocol under the three-seed ensemble (the pool's joint input) on the 8,198 balanced test records, binned, with the flows between consecutive stages: blue flows improve the bin, vermillion flows worsen it (a harmful transition), grey flows keep it; the printed shares are the records entering and leaving rank 1 at each transition (aggregates in the repository's \texttt{figures\_3d.json}). Takeaway: at every transition a few per cent of records leave rank 1 while many more enter it, and most of those leaving stay in the top three, which is what a pool over the stage posteriors can recover and a hard reversion rule cannot separate from legitimate corrections.}
\label{fig:rankflow}
\end{figure}

\paragraph{Does the earlier posterior add information given the current one?} Fitting on the OOF posteriors and reading validation, a per-class scale and bias on the current ensemble alone lowers NLL at stage 6 from 1.747 to 1.630 and leaves top-1 unchanged (57.85 to 57.87); adding the previous stage's posterior as a second input of a static pool gives 58.02 / 1.603, all earlier stages 58.25 / 1.601, all earlier stages with per-class weights 58.54 / 1.589, the block experts 58.40 / 1.592, and both axes 58.74 / 1.573 (top-3 77.05 to 78.37). The same ordering holds at stages 4--5 and nothing beyond calibration appears at stages 2--3. About half of the NLL reduction is recalibration of the current posterior; the accuracy gain and the other half come from the earlier and block-specific posteriors given the current one (an estimation benefit, not new patient information).

\paragraph{Hard retention cannot separate harm from correction.} A logistic detector on inference-time quantities (both stages' entropy, margin and maximum posterior, their agreement, the cross-stage probabilities of each other's top-1, their KL divergence and the number of non-zero entries in the new block) recognises a harmful flip against all other records with validation AUROC 0.86 to 0.93, but restricted to disagreements it separates harmful flips from legitimate corrections with AUROC only 0.66 to 0.80, since both are disagreements with a confident later model. Reverting to the earlier prediction where the detector fires therefore blocks about as many corrections as it rescues: the net top-1 change at the detector's own threshold is $-0.34$ to $+0.70$ points across transitions $1\to2$ to $5\to6$, and $-0.05$ to $-1.19$ when the 10\% highest-scoring records are reverted (negative at every transition) and between $\hardRuleNetMin$ and $\hardRuleNetMax$ over reversion fractions of 1--20\%, against oracle gains of $+2.7$ to $+5.8$. The soft pool, which reweights rather than switches, obtains $\dOnevitals$ and $\dOneimpressions$ top-1 at stages 5--6 (Table~\ref{tab:main}); a hard, agreement-conditioned rule of the kind used for early exits \citep{kaya2019shallow} is the wrong instrument here.

\paragraph{A training-time alternative: stage-congruent training.} A training-time analogue of the pool asks the stage-$t$ network not to depart from the stage-$(t{-}1)$ posterior where that posterior was right: the frozen recipe's cross-entropy plus a focal congruence term $\lambda\,\mathbb{1}[\arg\max q = y]\,\mathrm{KL}(q \,\|\, p_\theta)$, where $q$ is the OOF three-seed posterior of stage $t{-}1$ on the same fit row and $p_\theta$ the stage-$t$ network's posterior (the focal distillation of \citealp{yan2021positive}, with a teacher that holds strictly less information than the student); $\lambda=3$ with the focal weight was selected on validation top-1 at stages 4--6 among $\lambda\in\{0.3,1,3\}$ and plain distillation. Table~\ref{tab:cong} gives the three-seed results against the plain ensemble, against a self-distillation control that uses the stage's \emph{own} OOF ensemble as teacher, and with the pool refitted on the congruent networks. The congruent single network gains 1.0 to 1.3 validation top-1 points over the plain one, but the congruent three-seed ensemble is $-0.57$, $-0.45$ and $-0.20$ test top-1 points from the plain ensemble at stages 4--6 ($p=0.10$, $0.19$, $0.58$), the self-distillation control lands at the same place, and the pool refitted on the congruent networks is no better than the pool on the plain ones. The congruence term lowers the harmful-flip rate from 5.1--5.7\% to 4.2--4.6\% but lowers the correction rate by as much, and its NLL reduction (stage 6: 1.696 to 1.577) is the smoothing any distillation from a three-seed teacher provides. What the pool recovers is not recoverable by regularising the later network toward the earlier one.

\begin{table}[t]
\centering\scriptsize\setlength{\tabcolsep}{3pt}
\caption{Stage-congruent training on the corrected family (clean boundary, test split, 3-seed ensembles). Congruent: the stage MLP trained with the frozen recipe plus a focal congruence term ($\lambda=3$) toward the stage-$(t{-}1)$ out-of-fold ensemble, weighted on records the earlier stage ranks correctly; self-distilled: the same term toward the stage's own out-of-fold ensemble (the ensemble-distillation control); STEP on congruent: the pool refitted with the congruent joint model. Paired McNemar $p$ and bootstrap CIs on the 8,198 test records; descriptive.}
\label{tab:cong}
\vspace{4pt}
\begin{adjustbox}{max width=\linewidth}
\begin{tabular}{lrrrrrrr}
\toprule
stage & plain ens. & congruent ens. & $\Delta_1$ [CI] ($p$) & $\Delta_3$ ($p$) & $\Delta$NLL & self-distilled ens.\ ($\Delta_1$, $p$) & STEP on congruent ($\Delta_1$ vs STEP on plain, $p$)\\
\midrule
2 & \textbf{24.54} / 43.79 & 24.37 / 43.61 & $-0.17$ [-0.51, 0.17] (0.38) & $-0.18$ (0.48) & $-0.001$ & -- & 24.53 / \textbf{43.94} ($-0.10$, 0.59)\\
3 & 27.56 / 48.16 & 27.57 / \textbf{48.50} & $+0.01$ [-0.35, 0.39] (1) & $+0.34$ (0.19) & $-0.017$ & -- & \textbf{27.82} / 48.34 ($-0.02$, 0.95)\\
4 & 48.16 / 69.60 & 47.58 / 69.27 & $-0.57$ [-1.22, 0.12] (0.1) & $-0.33$ (0.35) & $-0.185$ & \textbf{48.32} ($+0.16$, 0.64) & 48.18 / \textbf{70.11} ($-0.06$, 0.85)\\
5 & \textbf{52.94} / 74.53 & 52.49 / 73.88 & $-0.45$ [-1.13, 0.24] (0.19) & $-0.65$ (0.045) & $-0.032$ & 52.53 ($-0.41$, 0.2) & 52.88 / \textbf{74.95} ($-0.83$, 0.0029)\\
6 & 58.34 / 78.56 & 58.15 / 78.49 & $-0.20$ [-0.83, 0.48] (0.58) & $-0.06$ (0.87) & $-0.119$ & 58.43 ($+0.09$, 0.82) & \textbf{58.98} / \textbf{79.34} ($-0.33$, 0.25)\\
\bottomrule
\end{tabular}
\end{adjustbox}
\end{table}

\paragraph{The conjunction test.} A pool with class-tied weights is additive in the experts' log-odds (App.~\ref{app:additivity}), so a non-additive interaction between blocks is not expressible within that class of pool. Whether the jointly trained model in fact recovers records whose block signals agree but whose decision-level combination fails is an empirical question. Let $E$ be the test records an experts-only static pool (App.~\ref{app:stepscope}) ranks incorrectly, $S\subset E$ those whose true protocol is in the top-3 of at least two experts (the individual signals are present, the combination is missed) and $N\subset E$ those no expert has in its top-3. At stages 4--6 the jointly trained MLP ensemble recovers 29--33\% of $S$ against 13--15\% of $E\setminus S$ and 6--7\% of $N$; STEP recovers 24--31\% of $S$. Two decision-level combiners that are more expressive than the class-tied pool, the gated pool and a per-class-weighted pool over the same expert posteriors, recover only 7--12\% of $S$. This is an empirical contrast between the jointly trained input and three decision-level rules on the same posteriors; it does not show that no decision-level rule could recover these records.

\paragraph{One block-masked network for all stages.} If the pooling gain is finite-sample shrinkage toward lower-variance views, the same effect should be obtainable at training time by making one network predict well at every reveal stage: each fit row enters training in its six stage views (the stage-$t$ matrix scattered into the stage-6 column layout, absent columns zero, plus a six-way stage indicator), so the network cannot over-rely on a block that may be absent. One network then serves all six stages with no pool. (Zeroing the later blocks on the stage-6 matrix is not equivalent: the 1,063 shared clinical columns carry merged symptom and impression counts at stage 6, and a first pilot that used that shortcut scored six \% too high at stage 4; every stage view is taken from the stage-$t$ matrix itself.) Pre-registered hypotheses and results (Tables~\ref{tab:flips} and~\ref{tab:main}; three seeds, frozen recipe): (H1) the masked single network beats the per-stage single network at stages 4--6 by $+0.61$, $+0.29$ and $+0.81$ test top-1 \% (only the last near significance); (H2) the masked three-seed ensemble does not differ significantly from the per-stage ensemble ($-0.23$, $-0.51$, $+0.48$; all $p>0.15$), so masking does not add accuracy beyond seed ensembling; (H3) the pool refitted on the masked networks adds $+0.52$, $+1.23$ and $+0.59$ top-1 over the masked ensemble at stages 4--6 and lands where STEP on the per-stage networks lands (48.45 / 53.66 / 59.40 against 48.24 / 53.71 / 59.31), so pooling the masked networks lands where pooling the per-stage networks does, while the masked ensemble alone (58.82 at stage 6) lies between the per-stage ensemble (58.34) and STEP (59.31). What masking changes is the \emph{consistency} of the stage sequence: harmful transitions fall from 5.17 / 5.66 / 5.07\% to 4.70 / 4.24 / 2.94\% at $3\to4$, $4\to5$ and $5\to6$ while corrections fall much less (25.8 / 10.4 / 10.5

\begin{table}[t]
\centering\scriptsize\setlength{\tabcolsep}{3pt}
\caption{Harmful transitions (record ranked correctly at stage $t{-}1$ and incorrectly at $t$) and corrections (the reverse), in \% of the 8,198 test records, for each method's own stage sequence on the repaired family under the clean boundary. Every method except the last two is a per-stage 3-seed ensemble with or without a pool; the masked ensemble is one network per seed trained over all six reveal stages with the unrevealed blocks masked, serving every stage.}
\label{tab:flips}
\vspace{4pt}
\begin{adjustbox}{max width=\linewidth}
\begin{tabular}{lrrrrrr}
\toprule
method & $3\to4$ harm / corr. & $4\to5$ harm / corr. & $5\to6$ harm / corr. & top-1 at 4 / 5 / 6 & mean harm 4--6 & mean corr.\ 4--6\\
\midrule
plain ens. & 5.17 / \textbf{25.77} & 5.66 / \textbf{10.44} & 5.07 / \textbf{10.48} & 48.16 / 52.94 / 58.34 & 5.30 & \textbf{15.56}\\
PA product & \textbf{2.61} / 20.64 & \textbf{2.20} / 8.16 & \textbf{1.63} / 7.37 & 45.11 / 51.07 / 56.81 & \textbf{2.15} & 12.06\\
static pool & 4.33 / 24.73 & 4.05 / 9.25 & 3.23 / 8.84 & 48.11 / 53.31 / 58.92 & 3.87 & 14.27\\
STEP & 4.51 / 24.91 & 3.72 / 9.19 & 2.96 / 8.56 & 48.24 / \textbf{53.71} / 59.31 & 3.73 & 14.22\\
congruent ens. & 4.09 / 24.10 & 4.21 / 9.11 & 4.21 / 9.87 & 47.58 / 52.49 / 58.15 & 4.17 & 14.36\\
masked ens. (one model) & 4.70 / 25.09 & 4.24 / 8.75 & 2.94 / 9.33 & 47.93 / 52.43 / 58.82 & 3.96 & 14.39\\
STEP on masked & 4.31 / 25.18 & 3.55 / 8.76 & 2.89 / 8.64 & \textbf{48.45} / 53.66 / \textbf{59.40} & 3.58 & 14.19\\
\bottomrule
\end{tabular}
\end{adjustbox}
\end{table}

\paragraph{Year hold-out.} A NEMSIS 2023 cohort was built with the 2024 rules (single-protocol records; the protocol label is the index of the eProtocols.01 code in the protocol list, verified to agree with the 2024 mapping on 1.55 million single-protocol records; up to 1,000 records per protocol, 80,284 in all; delimiter-safe extraction; the unchanged vitals and context code) together with a naturally distributed 2023 sample of 100,000 single-protocol records, and every 2024 model was applied to both. (A first extraction that had joined no vital-sign readings was caught by the vitals coverage check and rebuilt; the numbers are from the corrected build.) Across the year the balanced-cohort ensemble drifts by $\yFwdDriftOnedispatch$, $\yFwdDriftOnelocation$, $\yFwdDriftOneorgan$ top-1 \% at stages 1--3 and $\yFwdDriftOnescene$, $\yFwdDriftOnevitals$, $\yFwdDriftOneimpressions$ at stages 4--6, i.e.\ the drift sits in the symptom, vital-sign and impression blocks, and the pooling increment transfers at full size: $\yFwdOnescene$ / $\yFwdThreescene$, $\yFwdOnevitals$ / $\yFwdThreevitals$ and $\yFwdOneimpressions$ / $\yFwdThreeimpressions$ top-1 / top-3 at stages 4--6 ($n=80{,}284$, every $p<10^{-26}$). In the reverse direction, a family trained on the 2023 balanced fit rows alone (in-year 2023 test: \yRevInYearOneimpressions\% top-1 at stage 6, \yRevInYearOnevitals\% at stage 5) drifts on the 81,978 2024 records to \yRevExtOneimpressions\% and \yRevExtOnevitals\% ($\yRevDriftOneimpressions$ and $\yRevDriftOnevitals$ \%), and again the pool adds $\yRevOneimpressions$ / $\yRevThreeimpressions$ and $\yRevOnevitals$ / $\yRevThreevitals$ top-1 / top-3 at stages 6 and 5 ($p<10^{-26}$) and $\yRevOnescene$ / $\yRevThreescene$ at stage 4 (Table~\ref{tab:year}, which also gives the static pool and the fixed product). The increment is therefore symmetric across the two years and of the same size as within a year, while the joint models themselves lose 5 to 8 \% to the year shift. On the natural 2023 sample the class bias fitted on balanced rows works against the deployment prior, as on the natural 2024 cohort below.

\begin{table}[t]
\centering\scriptsize\setlength{\tabcolsep}{2.5pt}
\caption{Year hold-out. Models trained on one year's balanced fit rows only (clean boundary, unchanged recipe and pool selection) and applied once to the other year: a balanced 2023 cohort built with the same rules (single-protocol records, up to 1,000 per protocol, delimiter-safe extraction, unchanged vitals-repair and context code), a naturally distributed 2023 sample, and, in reverse, the 2024 balanced cohort from a 2023-trained family. Micro top-1 / top-3, STEP minus the 3-seed ensemble (paired bootstrap CI, exact McNemar $p$), the static pool's increment, and macro top-1.}
\label{tab:year}
\vspace{4pt}
\begin{adjustbox}{max width=\linewidth}
\begin{tabular}{llrrrrrr}
\toprule
direction & st. & ens.\ top-1 / top-3 & STEP top-1 / top-3 & $\Delta_1$ [95\% CI] ($p$) & $\Delta_3$ ($p$) & static $\Delta_1$ / $\Delta_3$ & ens.\ / STEP macro top-1\\
\midrule
2024 $\to$ 2023 balanced (80,284) & 1 & 18.69 / 34.74 & \textbf{18.78} / \textbf{34.97} & {\bfseries\boldmath $+0.09$ [0.00, 0.18]} (0.049) & {\bfseries\boldmath $+0.23$} (0.00026) & $+0.06$ / $+0.15$ & 18.87 / \textbf{18.97}\\
 & 2 & 22.23 / 39.93 & \textbf{22.35} / \textbf{40.14} & {\bfseries\boldmath $+0.12$ [0.03, 0.22]} (0.012) & {\bfseries\boldmath $+0.20$} (0.0032) & $+0.01$ / $+0.18$ & 22.34 / \textbf{22.49}\\
 & 3 & \textbf{24.55} / \textbf{44.23} & 24.53 / 44.20 & $-0.02$ [-0.13, 0.09] (0.7) & {\bfseries\boldmath $-0.02$} (0.74) & {\bfseries\boldmath $+0.08$} / $-0.05$ & \textbf{24.68} / 24.66\\
 & 4 & 41.08 / 62.41 & \textbf{42.06} / \textbf{63.41} & {\bfseries\boldmath $+0.98$ [0.82, 1.15]} (1.3e-27) & {\bfseries\boldmath $+1.01$} (5.1e-31) & $+0.68$ / $+0.56$ & 41.20 / \textbf{42.19}\\
 & 5 & 46.07 / 67.43 & \textbf{47.19} / \textbf{68.67} & {\bfseries\boldmath $+1.11$ [0.94, 1.28]} (1.8e-35) & {\bfseries\boldmath $+1.24$} (3.8e-50) & $+0.88$ / $+0.95$ & 46.24 / \textbf{47.36}\\
 & 6 & 52.69 / 73.76 & \textbf{53.84} / \textbf{75.11} & {\bfseries\boldmath $+1.15$ [0.96, 1.34]} (3.2e-34) & {\bfseries\boldmath $+1.36$} (2.8e-59) & $+0.81$ / $+1.18$ & 52.81 / \textbf{53.95}\\
\addlinespace[2pt]
2024 $\to$ 2023 natural (100,000) & 1 & \textbf{7.89} / \textbf{13.83} & 7.50 / 13.62 & $-0.39$ [-0.46, -0.33] (1.3e-39) & {\bfseries\boldmath $-0.21$} (7.1e-11) & {\bfseries\boldmath $-0.24$} / $-0.50$ & 17.91 / \textbf{18.35}\\
 & 2 & \textbf{8.43} / \textbf{15.64} & 7.94 / 15.33 & {\bfseries\boldmath $-0.49$ [-0.56, -0.43]} (2.9e-49) & {\bfseries\boldmath $-0.31$} (1.3e-12) & $-0.60$ / $-0.34$ & 20.82 / \textbf{21.05}\\
 & 3 & \textbf{9.24} / \textbf{19.08} & 8.98 / 18.19 & $-0.26$ [-0.33, -0.19] (1e-15) & $-0.89$ (4.4e-84) & {\bfseries\boldmath $-0.21$} / {\bfseries\boldmath $-0.77$} & 22.50 / \textbf{22.69}\\
 & 4 & \textbf{18.26} / \textbf{35.25} & 17.12 / 31.84 & {\bfseries\boldmath $-1.14$ [-1.27, -1.01]} (6.9e-67) & {\bfseries\boldmath $-3.41$} (0) & $-1.40$ / $-4.83$ & 41.43 / \textbf{41.73}\\
 & 5 & \textbf{19.04} / \textbf{37.16} & 18.59 / 34.64 & {\bfseries\boldmath $-0.45$ [-0.58, -0.32]} (2.4e-12) & {\bfseries\boldmath $-2.53$} (4.2e-184) & $-0.63$ / $-3.37$ & 45.43 / \textbf{46.39}\\
 & 6 & \textbf{23.15} / \textbf{45.71} & 22.33 / 40.38 & {\bfseries\boldmath $-0.82$ [-0.96, -0.69]} (4.6e-31) & {\bfseries\boldmath $-5.34$} (0) & $-1.00$ / $-6.17$ & 52.99 / \textbf{54.80}\\
\addlinespace[2pt]
2023 $\to$ 2024 balanced (81,978) & 1 & 19.40 / 36.09 & \textbf{19.54} / \textbf{36.30} & {\bfseries\boldmath $+0.14$ [0.02, 0.25]} (0.028) & {\bfseries\boldmath $+0.22$} (9.9e-05) & {\bfseries\boldmath $+0.14$} / $+0.20$ & 19.40 / \textbf{19.53}\\
 & 2 & 22.18 / 39.94 & \textbf{22.20} / \textbf{40.17} & {\bfseries\boldmath $+0.02$ [-0.10, 0.13]} (0.77) & $+0.23$ (0.00027) & $-0.00$ / {\bfseries\boldmath $+0.30$} & 22.17 / \textbf{22.19}\\
 & 3 & 24.40 / \textbf{44.10} & \textbf{24.57} / 44.08 & {\bfseries\boldmath $+0.17$ [0.06, 0.28]} (0.0024) & {\bfseries\boldmath $-0.02$} (0.83) & $+0.07$ / $-0.15$ & 24.40 / \textbf{24.57}\\
 & 4 & 41.92 / 62.94 & \textbf{42.24} / \textbf{63.92} & {\bfseries\boldmath $+0.32$ [0.15, 0.48]} (0.00014) & {\bfseries\boldmath $+0.98$} (9.8e-34) & $+0.07$ / $+0.81$ & 41.91 / \textbf{42.23}\\
 & 5 & 46.01 / 67.78 & \textbf{47.01} / \textbf{69.04} & {\bfseries\boldmath $+1.00$ [0.81, 1.18]} (5.9e-27) & {\bfseries\boldmath $+1.26$} (1.7e-48) & $+0.91$ / $+1.17$ & 46.00 / \textbf{47.00}\\
 & 6 & 51.16 / 72.83 & \textbf{52.16} / \textbf{74.06} & {\bfseries\boldmath $+1.00$ [0.82, 1.17]} (4.4e-27) & {\bfseries\boldmath $+1.24$} (2.3e-50) & $+0.77$ / $+1.04$ & 51.15 / \textbf{52.15}\\
\addlinespace[2pt]
\bottomrule
\end{tabular}
\end{adjustbox}
\end{table}

\paragraph{Distribution comparison at every stage.} Table~\ref{tab:dist} gives, at stages 5--6, the comparison of \S\ref{sec:dist} (the balanced-test rows are the macros of Table~\ref{tab:main}), and Table~\ref{tab:distfull} extends it to all six stages and to top-1, on the four evaluation cohorts, as trained and after the $\tau{=}1$ Bayes prior approximation with the two-fold known prior; Figure~\ref{fig:dist} (\S\ref{sec:dist}) draws the two natural cohorts against the balanced test split. The correction and the prior are computed CPU-only from the saved per-record logits; on the natural 2024 cohort the $\tau{=}1$ top-1 / top-3 values reproduce the known-prior column of Table~\ref{tab:priorshift} exactly, and on the balanced cross-year cohorts, whose target prior is uniform to within a few records per protocol (the 2023 cohort has \yTwentyThreeBelowThousand{} protocols below 1,000 records, the smallest with \yTwentyThreeMinCount), the correction moves every value by less than a point. At every stage the pool's class bias costs micro accuracy as trained on both natural cohorts, and the $\tau{=}1$ correction restores STEP's lead at every depth from the scene stage on (at stages 1--3 on the 2023 cohort the top-3 and top-5 differences are within $\pm0.04$).

\begin{table}[t]
\centering\footnotesize\setlength{\tabcolsep}{2.5pt}
\caption{Balanced-cohort ensemble and STEP applied unchanged (as trained) and after the $\tau{=}1$ Bayes prior approximation (two-fold known prior; \S\ref{sec:dist}) to the real-world and multi-year samples: micro top-3 / top-5 (\%) and the paired STEP $-$ ens.\ difference at stages 5--6; all stages, top-1 and exact $p$ in Table~\ref{tab:distfull}.}
\label{tab:dist}
\vspace{4pt}
\begin{adjustbox}{max width=\linewidth}
\begin{tabular}{lclccc}
\toprule
cohort ($n$) & st. & setting & ens.\ top-3 / top-5 & STEP top-3 / top-5 & $\Delta_3$ / $\Delta_5$ (STEP $-$ ens.)\\
\midrule
balanced 2024 test (8,198) & 5 & as trained & \ensTopThreevitals / \ensTopFivevitals & \textbf{\stepTopThreevitals} / \textbf{\stepTopFivevitals} & $\dThreevitals$ / $\dFivevitals$\\
natural 2024 (90,759) & 5 & as trained & \textbf{39.40} / \textbf{55.59} & 36.52 / 50.42 & $-2.88$ / $-5.17$\\
 &  & $\tau{=}1$ & 87.36 / 93.97 & \textbf{88.29} / \textbf{94.55} & $+0.93$ / $+0.58$\\
natural 2023 (100,000) & 5 & as trained & \textbf{37.16} / \textbf{53.29} & 34.64 / 48.37 & $-2.53$ / $-4.92$\\
 &  & $\tau{=}1$ & 84.23 / 92.05 & \textbf{85.21} / \textbf{92.75} & $+0.98$ / $+0.70$\\
balanced 2023, 24$\to$23 (80,284) & 5 & as trained & 67.43 / 75.85 & \textbf{68.67} / \textbf{76.97} & $+1.24$ / $+1.13$\\
balanced 2024, 23$\to$24 (81,978) & 5 & as trained & 67.78 / 76.32 & \textbf{69.04} / \textbf{77.53} & $+1.26$ / $+1.21$\\
\midrule
balanced 2024 test (8,198) & 6 & as trained & \ensTopThreeimpressions / \ensTopFiveimpressions & \textbf{\stepTopThreeimpressions} / \textbf{\stepTopFiveimpressions} & $\dThreeimpressions$ / $\dFiveimpressions$\\
natural 2024 (90,759) & 6 & as trained & \textbf{47.52} / \textbf{65.12} & 41.92 / 57.64 & $-5.60$ / $-7.49$\\
 &  & $\tau{=}1$ & 88.14 / 94.67 & \textbf{89.45} / \textbf{95.51} & $+1.32$ / $+0.85$\\
natural 2023 (100,000) & 6 & as trained & \textbf{45.71} / \textbf{63.83} & 40.38 / 56.45 & $-5.34$ / $-7.38$\\
 &  & $\tau{=}1$ & 86.02 / 93.59 & \textbf{87.18} / \textbf{94.50} & $+1.16$ / $+0.91$\\
balanced 2023, 24$\to$23 (80,284) & 6 & as trained & 73.76 / 81.13 & \textbf{75.11} / \textbf{82.49} & $+1.36$ / $+1.36$\\
balanced 2024, 23$\to$24 (81,978) & 6 & as trained & 72.83 / 80.59 & \textbf{74.06} / \textbf{81.68} & $+1.24$ / $+1.10$\\
\bottomrule
\end{tabular}
\end{adjustbox}
\end{table}

\begin{table}[t]
\centering\scriptsize\setlength{\tabcolsep}{2pt}
\caption{Distribution comparison, all six stages: the balanced-cohort 3-seed ensemble and STEP applied unchanged (as trained) and after the $\tau{=}1$ Bayes prior approximation (two-fold known prior, as in Table~\ref{tab:dist}; $\pi_{\text{train}}$ is uniform for the 2024-trained family and the empirical 2023 balanced-cohort frequencies for the 2023-trained family) to the naturally distributed 2024 (90,759) and 2023 (100,000) samples and to the balanced cross-year hold-outs (80,284 and 81,978). Micro top-1 / top-3 / top-5 (\%), STEP minus ensemble in points with the exact two-sided McNemar $p$ for top-1 and top-3 (descriptive; the top-5 $p$ values are in the repository's \texttt{dist\_compare\_r5.json}). On the natural 2024 cohort the $\tau{=}1$ top-1 / top-3 values reproduce the known-prior column of Table~\ref{tab:priorshift} exactly. On the balanced cross-year cohorts the target prior is near-uniform, so $\tau{=}1$ is close to the identity.}
\label{tab:distfull}
\vspace{4pt}
\begin{tabular}{llcccccc}
\toprule
cohort & st. & setting & ens.\ top-1 / 3 / 5 & STEP top-1 / 3 / 5 & $\Delta_1$ ($p$) & $\Delta_3$ ($p$) & $\Delta_5$\\
\midrule
nat.\ 2024 & 1 & as trained & \textbf{9.30} / \textbf{15.51} / \textbf{19.00} & 8.91 / 15.27 / 18.67 & $-0.40$ ($5.8\!\times\!10^{-38}$) & $-0.24$ ($3.4\!\times\!10^{-12}$) & $-0.33$\\
 &  & $\tau{=}1$ & 53.38 / 78.13 / 85.74 & \textbf{53.59} / \textbf{78.23} / \textbf{85.86} & $+0.21$ ($6.6\!\times\!10^{-6}$) & $+0.10$ ($0.018$) & $+0.13$\\
 & 2 & as trained & \textbf{9.58} / \textbf{17.17} / \textbf{22.69} & 9.07 / 16.88 / 22.56 & $-0.50$ ($5.5\!\times\!10^{-45}$) & $-0.29$ ($4.1\!\times\!10^{-11}$) & $-0.14$\\
 &  & $\tau{=}1$ & 54.00 / 79.53 / 87.07 & \textbf{54.65} / \textbf{79.70} / \textbf{87.18} & $+0.65$ ($3.1\!\times\!10^{-23}$) & $+0.17$ ($1.8\!\times\!10^{-5}$) & $+0.12$\\
 & 3 & as trained & \textbf{10.06} / \textbf{19.87} / 25.38 & 9.77 / 19.12 / \textbf{26.97} & $-0.30$ ($3.0\!\times\!10^{-18}$) & $-0.75$ ($3.4\!\times\!10^{-63}$) & $+1.59$\\
 &  & $\tau{=}1$ & 54.27 / 80.44 / 88.25 & \textbf{55.03} / \textbf{80.68} / \textbf{88.33} & $+0.76$ ($1.1\!\times\!10^{-26}$) & $+0.23$ ($6.0\!\times\!10^{-7}$) & $+0.07$\\
 & 4 & as trained & \textbf{19.49} / \textbf{37.18} / \textbf{51.61} & 18.31 / 33.49 / 45.51 & $-1.18$ ($1.8\!\times\!10^{-63}$) & $-3.68$ ($<10^{-300}$) & $-6.10$\\
 &  & $\tau{=}1$ & 55.23 / 84.73 / 92.31 & \textbf{60.28} / \textbf{86.98} / \textbf{93.56} & $+5.05$ ($<10^{-300}$) & $+2.25$ ($9.7\!\times\!10^{-184}$) & $+1.25$\\
 & 5 & as trained & \textbf{20.92} / \textbf{39.40} / \textbf{55.59} & 20.23 / 36.52 / 50.42 & $-0.68$ ($1.6\!\times\!10^{-24}$) & $-2.88$ ($9.7\!\times\!10^{-219}$) & $-5.17$\\
 &  & $\tau{=}1$ & 60.16 / 87.36 / 93.97 & \textbf{61.98} / \textbf{88.29} / \textbf{94.55} & $+1.82$ ($3.7\!\times\!10^{-81}$) & $+0.93$ ($1.0\!\times\!10^{-52}$) & $+0.58$\\
 & 6 & as trained & \textbf{24.42} / \textbf{47.52} / \textbf{65.12} & 23.24 / 41.92 / 57.64 & $-1.19$ ($9.2\!\times\!10^{-57}$) & $-5.60$ ($<10^{-300}$) & $-7.49$\\
 &  & $\tau{=}1$ & 60.81 / 88.14 / 94.67 & \textbf{63.45} / \textbf{89.45} / \textbf{95.51} & $+2.64$ ($2.3\!\times\!10^{-137}$) & $+1.32$ ($4.0\!\times\!10^{-90}$) & $+0.85$\\
\addlinespace[2pt]
nat.\ 2023 & 1 & as trained & \textbf{7.89} / \textbf{13.83} / \textbf{17.59} & 7.50 / 13.62 / 17.19 & $-0.39$ ($1.3\!\times\!10^{-39}$) & $-0.21$ ($7.1\!\times\!10^{-11}$) & $-0.40$\\
 &  & $\tau{=}1$ & 51.77 / \textbf{74.83} / 83.40 & \textbf{52.26} / \textbf{74.83} / \textbf{83.49} & $+0.49$ ($4.3\!\times\!10^{-27}$) & $-0.00$ ($0.98$) & $+0.09$\\
 & 2 & as trained & \textbf{8.43} / \textbf{15.64} / \textbf{21.96} & 7.94 / 15.33 / 21.89 & $-0.49$ ($2.9\!\times\!10^{-49}$) & $-0.31$ ($1.3\!\times\!10^{-12}$) & $-0.08$\\
 &  & $\tau{=}1$ & 52.52 / \textbf{75.76} / 85.25 & \textbf{53.35} / 75.75 / \textbf{85.39} & $+0.83$ ($4.2\!\times\!10^{-44}$) & $-0.01$ ($0.8$) & $+0.14$\\
 & 3 & as trained & \textbf{9.24} / \textbf{19.08} / 24.70 & 8.98 / 18.19 / \textbf{27.21} & $-0.26$ ($1.0\!\times\!10^{-15}$) & $-0.89$ ($4.4\!\times\!10^{-84}$) & $+2.50$\\
 &  & $\tau{=}1$ & 53.02 / 77.22 / \textbf{86.75} & \textbf{54.00} / \textbf{77.30} / 86.70 & $+0.99$ ($1.5\!\times\!10^{-48}$) & $+0.08$ ($0.11$) & $-0.04$\\
 & 4 & as trained & \textbf{18.26} / \textbf{35.25} / \textbf{49.37} & 17.12 / 31.84 / 43.47 & $-1.14$ ($6.9\!\times\!10^{-67}$) & $-3.41$ ($<10^{-300}$) & $-5.90$\\
 &  & $\tau{=}1$ & 52.76 / 81.71 / 90.24 & \textbf{57.30} / \textbf{83.59} / \textbf{91.61} & $+4.53$ ($<10^{-300}$) & $+1.88$ ($1.7\!\times\!10^{-143}$) & $+1.37$\\
 & 5 & as trained & \textbf{19.04} / \textbf{37.16} / \textbf{53.29} & 18.59 / 34.64 / 48.37 & $-0.45$ ($2.4\!\times\!10^{-12}$) & $-2.53$ ($4.2\!\times\!10^{-184}$) & $-4.92$\\
 &  & $\tau{=}1$ & 57.04 / 84.23 / 92.05 & \textbf{58.76} / \textbf{85.21} / \textbf{92.75} & $+1.72$ ($1.5\!\times\!10^{-86}$) & $+0.98$ ($2.2\!\times\!10^{-56}$) & $+0.70$\\
 & 6 & as trained & \textbf{23.15} / \textbf{45.71} / \textbf{63.83} & 22.33 / 40.38 / 56.45 & $-0.82$ ($4.6\!\times\!10^{-31}$) & $-5.34$ ($<10^{-300}$) & $-7.38$\\
 &  & $\tau{=}1$ & 57.84 / 86.02 / 93.59 & \textbf{60.36} / \textbf{87.18} / \textbf{94.50} & $+2.53$ ($1.9\!\times\!10^{-144}$) & $+1.16$ ($2.6\!\times\!10^{-67}$) & $+0.91$\\
\addlinespace[2pt]
bal.\ 2023 (24$\to$23) & 1 & as trained & 18.69 / 34.74 / 44.54 & \textbf{18.78} / \textbf{34.97} / \textbf{44.57} & $+0.09$ ($0.049$) & $+0.23$ ($2.6\!\times\!10^{-4}$) & $+0.03$\\
 &  & $\tau{=}1$ & 18.81 / 34.93 / 44.72 & \textbf{18.84} / \textbf{35.12} / \textbf{44.83} & $+0.03$ ($0.6$) & $+0.20$ ($0.0013$) & $+0.11$\\
 & 2 & as trained & 22.23 / 39.93 / 50.09 & \textbf{22.35} / \textbf{40.14} / \textbf{50.59} & $+0.12$ ($0.012$) & $+0.20$ ($0.0032$) & $+0.50$\\
 &  & $\tau{=}1$ & \textbf{22.32} / 40.10 / 50.25 & 22.26 / \textbf{40.20} / \textbf{50.68} & $-0.06$ ($0.23$) & $+0.09$ ($0.17$) & $+0.43$\\
 & 3 & as trained & \textbf{24.55} / \textbf{44.23} / 54.61 & 24.53 / 44.20 / \textbf{54.72} & $-0.02$ ($0.7$) & $-0.02$ ($0.74$) & $+0.11$\\
 &  & $\tau{=}1$ & \textbf{24.58} / 44.24 / 54.86 & 24.54 / \textbf{44.31} / \textbf{54.93} & $-0.04$ ($0.52$) & $+0.07$ ($0.33$) & $+0.07$\\
 & 4 & as trained & 41.08 / 62.41 / 71.21 & \textbf{42.06} / \textbf{63.41} / \textbf{72.33} & $+0.98$ ($1.3\!\times\!10^{-27}$) & $+1.01$ ($5.1\!\times\!10^{-31}$) & $+1.12$\\
 &  & $\tau{=}1$ & 41.12 / 62.47 / 71.27 & \textbf{42.02} / \textbf{63.50} / \textbf{72.40} & $+0.90$ ($1.5\!\times\!10^{-23}$) & $+1.03$ ($9.6\!\times\!10^{-33}$) & $+1.13$\\
 & 5 & as trained & 46.07 / 67.43 / 75.85 & \textbf{47.19} / \textbf{68.67} / \textbf{76.97} & $+1.11$ ($1.8\!\times\!10^{-35}$) & $+1.24$ ($3.8\!\times\!10^{-50}$) & $+1.13$\\
 &  & $\tau{=}1$ & 46.04 / 67.47 / 75.87 & \textbf{47.19} / \textbf{68.73} / \textbf{76.96} & $+1.15$ ($1.7\!\times\!10^{-37}$) & $+1.26$ ($1.5\!\times\!10^{-51}$) & $+1.09$\\
 & 6 & as trained & 52.69 / 73.76 / 81.13 & \textbf{53.84} / \textbf{75.11} / \textbf{82.49} & $+1.15$ ($3.2\!\times\!10^{-34}$) & $+1.36$ ($2.8\!\times\!10^{-59}$) & $+1.36$\\
 &  & $\tau{=}1$ & 52.72 / 73.79 / 81.15 & \textbf{53.92} / \textbf{75.21} / \textbf{82.55} & $+1.20$ ($1.8\!\times\!10^{-37}$) & $+1.42$ ($1.7\!\times\!10^{-64}$) & $+1.41$\\
\addlinespace[2pt]
bal.\ 2024 (23$\to$24) & 1 & as trained & 19.40 / 36.09 / 45.69 & \textbf{19.54} / \textbf{36.30} / \textbf{45.97} & $+0.14$ ($0.028$) & $+0.22$ ($9.9\!\times\!10^{-5}$) & $+0.28$\\
 &  & $\tau{=}1$ & 19.40 / 36.30 / 45.84 & \textbf{19.50} / \textbf{36.40} / \textbf{46.01} & $+0.09$ ($0.08$) & $+0.10$ ($0.1$) & $+0.17$\\
 & 2 & as trained & 22.18 / 39.94 / 50.53 & \textbf{22.20} / \textbf{40.17} / \textbf{50.82} & $+0.02$ ($0.77$) & $+0.23$ ($2.7\!\times\!10^{-4}$) & $+0.29$\\
 &  & $\tau{=}1$ & 22.50 / 40.08 / 50.50 & \textbf{22.55} / \textbf{40.25} / \textbf{50.78} & $+0.05$ ($0.4$) & $+0.17$ ($0.0064$) & $+0.28$\\
 & 3 & as trained & 24.40 / \textbf{44.10} / 54.51 & \textbf{24.57} / 44.08 / \textbf{54.64} & $+0.17$ ($0.0024$) & $-0.02$ ($0.83$) & $+0.13$\\
 &  & $\tau{=}1$ & 24.77 / 44.17 / 54.59 & \textbf{24.85} / \textbf{44.18} / \textbf{54.75} & $+0.08$ ($0.17$) & $+0.01$ ($0.94$) & $+0.16$\\
 & 4 & as trained & 41.92 / 62.94 / 72.19 & \textbf{42.24} / \textbf{63.92} / \textbf{73.09} & $+0.32$ ($1.4\!\times\!10^{-4}$) & $+0.98$ ($9.8\!\times\!10^{-34}$) & $+0.90$\\
 &  & $\tau{=}1$ & 41.99 / 62.99 / 72.19 & \textbf{42.33} / \textbf{63.92} / \textbf{73.05} & $+0.34$ ($3.6\!\times\!10^{-5}$) & $+0.92$ ($5.1\!\times\!10^{-30}$) & $+0.86$\\
 & 5 & as trained & 46.01 / 67.78 / 76.32 & \textbf{47.01} / \textbf{69.04} / \textbf{77.53} & $+1.00$ ($5.9\!\times\!10^{-27}$) & $+1.26$ ($1.7\!\times\!10^{-48}$) & $+1.21$\\
 &  & $\tau{=}1$ & 46.10 / 67.81 / 76.37 & \textbf{47.16} / \textbf{69.07} / \textbf{77.58} & $+1.06$ ($1.3\!\times\!10^{-29}$) & $+1.26$ ($2.4\!\times\!10^{-48}$) & $+1.21$\\
 & 6 & as trained & 51.16 / 72.83 / 80.59 & \textbf{52.16} / \textbf{74.06} / \textbf{81.68} & $+1.00$ ($4.4\!\times\!10^{-27}$) & $+1.24$ ($2.3\!\times\!10^{-50}$) & $+1.10$\\
 &  & $\tau{=}1$ & 51.21 / 72.87 / 80.60 & \textbf{52.30} / \textbf{74.12} / \textbf{81.72} & $+1.09$ ($8.3\!\times\!10^{-32}$) & $+1.25$ ($2.1\!\times\!10^{-51}$) & $+1.12$\\
\addlinespace[2pt]
\bottomrule
\end{tabular}
\end{table}

\paragraph{The prior-shift trade-off.} Table~\ref{tab:priorshift} reports the remaining views of the naturally distributed 2024 cohort at stages 5--6 with the models of Table~\ref{tab:main} applied unchanged. As trained, the pool \emph{lowers} micro accuracy: at stage 6 micro top-1 falls from \extEnsMicroOne\% to \extStepMicroOne\% and top-3 from \extEnsMicroThree\% to \extStepMicroThree\% ($p<10^{-50}$), because the pool's per-class bias, fitted on balanced rows, moves mass away from the majority protocol. What transfers is the protocol-level effect: macro top-1 rises from \extEnsMacroOne\% to \extStepMacroOne\% at stage 6, and once the deployment prior is supplied ($\log\pi_{\text{ext}}/\pi_{\text{train}}$ added to every method's log-posterior, the $\tau{=}1$ correction of Table~\ref{tab:dist}) micro top-1 rises from \extEnsKnownPrior\% to \extStepKnownPrior\%; the transductive label-free EM estimate of the prior \citep{saerens2002adjusting} keeps STEP ahead of the ensemble on micro accuracy but not ahead of the static pool on top-1 (Table~\ref{tab:e2em} repeats it with disjoint adaptation and evaluation halves and on the 2023 cohort). What is \emph{exploratory} is the remedy, a refit with the class bias frozen at zero after these results had been inspected. On the balanced test split it keeps $\extNoBiasBalOne$ / $\extNoBiasBalThree$ of the stage-6 increment, and on the natural cohort it is ahead of the ensemble on micro top-1 (\extNoBiasMicroOne\% against \extEnsMicroOne\%) and macro F1 (\extNoBiasMacroF{} against \extEnsMacroF). Because the choice was made after the external cohort had been seen, those numbers are not a held-out test of it. A bias-free pool with an EM-estimated prior is a candidate for held-out evaluation, not a recommendation; the EM correction trades macro recall for micro accuracy, and the paper reports both variants.

\begin{table}[t]
\centering\scriptsize\setlength{\tabcolsep}{2.4pt}
\caption{The prior-shift trade-off on the naturally distributed NEMSIS 2024 cohort (90,759 records, one protocol is 54\% of it; models and pools exactly as in Table~\ref{tab:main}). As trained, the pool's fitted class bias lowers micro accuracy under the deployment prior while macro top-1 (per-protocol mean) rises. ``Known prior'': $\log\pi_{\text{ext}}/\pi_{\text{train}}$ added to the log-posterior, prior estimated on one random half and scored on the other. ``EM'': prior estimated transductively from the unlabelled posteriors \citep{saerens2002adjusting}; it raises micro accuracy for every model and lowers macro F1 for the bias-free variant, because it moves rare-protocol predictions to the majority protocol. The indented row is \emph{exploratory}: the bias was frozen after the external results had been inspected, so its external numbers are not a held-out test of that choice.}
\label{tab:priorshift}
\vspace{4pt}
\resizebox{\textwidth}{!}{\begin{tabular}{llcccccccc}
\toprule
\multirow{2}{*}{model} & \multirow{2}{*}{st.} & balanced test & \multicolumn{3}{c}{natural 2024 cohort, as trained} & known prior & \multicolumn{2}{c}{EM-estimated prior}\\
\cmidrule(lr){3-3}\cmidrule(lr){4-6}\cmidrule(lr){7-7}\cmidrule(lr){8-9}
 & & $\Delta_1$ / $\Delta_3$ vs ens. & micro top-1 / top-3 & macro top-1 & macro F1 (top-1 / top-3) & micro top-1 / top-3 & micro top-1 / top-3 & macro F1\\
\midrule
3-seed ensemble & 5 & $+0.00$ / $+0.00$ & 20.92 / 39.40 & 55.36 & 17.30 / 13.79 & 60.16 / 87.36 & 53.06 / 84.84 & \textbf{24.55}\\
static pool & 5 & $+0.37$ / $+0.82$ & 20.11 / 35.86 & 56.55 & 17.12 / 13.47 & 61.64 / 88.02 & 57.54 / 85.95 & 23.57\\
STEP & 5 & {\bfseries\boldmath $+0.77$} / {\bfseries\boldmath $+1.02$} & 20.23 / 36.52 & \textbf{56.62} & 17.09 / 13.48 & \textbf{61.98} / \textbf{88.29} & 54.24 / 86.17 & 23.90\\
\quad STEP, class bias frozen at 0 & 5 & $+0.38$ / $+0.84$ & \textbf{21.65} / \textbf{40.97} & 56.13 & \textbf{17.76} / \textbf{14.00} & -- & \textbf{62.06} / \textbf{86.23} & 17.85\\
\addlinespace[1.5pt]
3-seed ensemble & 6 & $+0.00$ / $+0.00$ & 24.42 / 47.52 & 58.18 & 20.04 / 15.60 & 60.81 / 88.14 & 55.38 / 85.45 & \textbf{28.48}\\
static pool & 6 & $+0.57$ / $+1.18$ & 23.20 / 41.11 & 58.92 & 19.69 / 15.03 & 62.96 / 89.17 & 61.06 / 88.32 & 27.90\\
STEP & 6 & {\bfseries\boldmath $+0.96$} / $+1.27$ & 23.24 / 41.92 & \textbf{60.05} & 19.58 / 15.01 & \textbf{63.45} / \textbf{89.45} & 60.41 / \textbf{88.41} & 26.98\\
\quad STEP, class bias frozen at 0 & 6 & $+0.73$ / {\bfseries\boldmath $+1.43$} & \textbf{25.22} / \textbf{47.65} & 59.66 & \textbf{20.38} / \textbf{15.68} & -- & \textbf{62.78} / 87.71 & 17.99\\
\bottomrule
\end{tabular}}
\end{table}

\paragraph{EM with disjoint adaptation and evaluation, and its sensitivity to calibration.} The EM rows of Table~\ref{tab:priorshift} estimate each method's prior from the unlabelled posteriors of the same records they score (transductive adaptation; no labels are used). Table~\ref{tab:e2em} instead estimates the prior on one random half of each natural cohort (\etwoNatHalf{} records for 2024), freezes it, scores the other half, and swaps the halves; the halves are those of the known-prior correction and the EM settings are unchanged. Disjoint and transductive estimates give nearly the same accuracy. Without calibration, STEP minus the adapted ensemble at stage 6 is $\etwoNatStepEnsOneSix$ top-1 (95\% interval with EM refitted in every bootstrap replicate: [$\etwoNatStepEnsOneEeLo$, $\etwoNatStepEnsOneEeHi$]) and $\etwoNatStepEnsThreeSix$ top-3 on 2024, and $\etwoOldStepEnsOneSix$ and $\etwoOldStepEnsThreeSix$ on 2023. These differences are not robust to calibration. On the source validation rows the seed ensemble is overconfident (temperature \etwoEnsTempMin--\etwoEnsTempMax) while the pools are not (\etwoPoolTempMin--\etwoPoolTempMax), and EM is sensitive to calibration \citep{alexandari2020maximum}. With each model temperature-scaled on source validation data before EM (pre-declared as a secondary sensitivity), STEP minus ensemble at stage 6 becomes $\etwoNatStepEnsTsOneSix$ / $\etwoNatStepEnsTsThreeSix$ (top-1 / top-3) on 2024 and $\etwoOldStepEnsTsOneSix$ / $\etwoOldStepEnsTsThreeSix$ on 2023, a top-3 margin that comes from the ensemble's shortlist degrading (top-3 \onRawNatEnsThreeSix\% to \onTsNatEnsThreeSix\% on 2024, \onRawOldEnsThreeSix\% to \onTsOldEnsThreeSix\% on 2023) while STEP's is unchanged (\onRawNatStepThreeSix\% to \onTsNatStepThreeSix\% on 2024); with bias-corrected temperature scaling (added afterwards, exploratory) it is $\etwoNatStepEnsBctsOneSix$ / $\etwoNatStepEnsBctsThreeSix$ and $\etwoOldStepEnsBctsOneSix$ / $\etwoOldStepEnsBctsThreeSix$. Without calibration the static pool leads STEP on top-1 at stages 5--6 ($\etwoNatStepStaticOneFive$ and $\etwoNatStepStaticOneSix$ for STEP minus static on 2024). Ranking every record by labelled prevalence alone reaches \etwoNatPrevOne\% / \etwoNatPrevThree\% on 2024 and \etwoOldPrevOne\% / \etwoOldPrevThree\% on 2023; uncalibrated EM leaves the ensemble below this baseline at stages 4--5, within two points of it at stage 6 on 2024 and below it there on 2023 (\etwoOldEnsDisjOneSix\%), label-free pools also fall below it at stage 4 in both cohorts (STEP: \etwoNatStepDisjOneFour\% and \etwoOldStepDisjOneFour\% uncalibrated), and on 2023 also later (STEP at stage 5: \etwoOldStepDisjOneFive\% uncalibrated, \onTsOldStepOneFive\% temperature-scaled; the temperature-scaled static pool at stage 6: \onTsOldStaticOneSix\%). Label-free adaptation therefore does not establish a robust micro-accuracy advantage for the pools. Both halves come from the same population and period, so this is within-cohort adaptation, not evidence about a later deployment period.

\begin{table}[t]
\centering
\footnotesize
\setlength{\tabcolsep}{3pt}
\caption{Label-free EM prior adaptation with disjoint adaptation and evaluation halves, natural cohorts, micro top-1 / top-3 (\%). Each cohort is split into two random halves (those of the known-prior correction). EM estimates each method's class prior from the unlabelled posteriors of one half; the estimate is frozen, applied to the other half, and the roles are swapped. \emph{Transductive}: prior estimated on the scored records themselves (as in Table~\ref{tab:priorshift}). \emph{TS}: each model temperature-scaled on the 7,378 source inner-validation rows before EM (pre-declared sensitivity); \emph{BCTS}$^\dagger$: temperature plus class bias \citep{alexandari2020maximum} (added after the uncalibrated results were seen; exploratory). Macro recall / macro F1 of the top-1 prediction under disjoint EM. \emph{Known prior}: labelled class frequencies of the other half, a labelled reference, not label-free. $^\ddagger$Ranking every record by the other half's labelled class frequencies. Scripts: \texttt{scripts/paper/revision/e2\_disjoint\_em.py}, \texttt{e2\_calibrated.py}.}
\label{tab:e2em}
\vspace{4pt}
\begin{adjustbox}{max width=\linewidth}
\begin{tabular}{llccccccc}
\toprule
cohort, stage & method & as trained & EM transd. & EM disjoint & macro R / F1 & EM disj., TS & EM disj., BCTS$^\dagger$ & known prior\\
\midrule
2024, stage 4 & ensemble & \textbf{19.49} / \textbf{37.18} & 37.85 / 75.50 & 37.81 / 75.48 & \textbf{39.26} / \textbf{21.56} & \textbf{56.50} / 78.09 & 56.22 / 84.40 & 55.23 / 84.73\\
 & static pool & 18.22 / 32.50 & 46.80 / 82.27 & 46.78 / 82.21 & 25.56 / 19.88 & 42.84 / 80.73 & 56.82 / 84.81 & 59.64 / 86.49\\
 & STEP & 18.31 / 33.49 & \textbf{48.27} / \textbf{83.69} & 48.25 / \textbf{83.72} & 21.77 / 18.39 & 47.18 / \textbf{83.44} & \textbf{57.53} / \textbf{85.58} & \textbf{60.28} / \textbf{86.98}\\
 & prevalence only$^\ddagger$ & -- & -- & \textbf{53.82} / 68.35 & -- & -- & -- & --\\
\addlinespace[2pt]
2024, stage 5 & ensemble & \textbf{20.92} / \textbf{39.40} & 53.06 / 84.84 & 53.06 / 84.88 & \textbf{39.20} / \textbf{25.37} & \textbf{60.71} / 84.74 & 60.34 / 86.29 & 60.16 / 87.36\\
 & static pool & 20.11 / 35.86 & \textbf{57.54} / 85.95 & \textbf{57.55} / 85.94 & 28.12 / 24.05 & 54.41 / 85.94 & 60.16 / 86.71 & 61.64 / 88.02\\
 & STEP & 20.23 / 36.52 & 54.24 / \textbf{86.17} & 54.22 / \textbf{86.13} & 27.72 / 24.06 & 54.41 / \textbf{86.13} & \textbf{60.49} / \textbf{87.20} & \textbf{61.98} / \textbf{88.29}\\
 & prevalence only$^\ddagger$ & -- & -- & 53.82 / 68.35 & -- & -- & -- & --\\
\addlinespace[2pt]
2024, stage 6 & ensemble & \textbf{24.42} / \textbf{47.52} & 55.38 / 85.45 & 55.40 / 85.46 & \textbf{45.86} / \textbf{29.61} & \textbf{60.93} / 83.55 & 60.91 / 87.52 & 60.81 / 88.14\\
 & static pool & 23.20 / 41.11 & \textbf{61.06} / 88.32 & \textbf{61.12} / 88.31 & 31.77 / 28.29 & 59.00 / 88.09 & 61.30 / 88.15 & 62.96 / 89.17\\
 & STEP & 23.24 / 41.92 & 60.41 / \textbf{88.41} & 60.41 / \textbf{88.37} & 26.82 / 26.23 & 59.59 / \textbf{88.33} & \textbf{61.39} / \textbf{88.34} & \textbf{63.45} / \textbf{89.45}\\
 & prevalence only$^\ddagger$ & -- & -- & 53.82 / 68.35 & -- & -- & -- & --\\
\addlinespace[2pt]
2023, stage 4 & ensemble & \textbf{18.26} / \textbf{35.25} & 37.94 / 72.61 & 37.91 / 72.60 & \textbf{32.32} / \textbf{19.19} & \textbf{53.04} / 72.97 & 51.97 / 75.95 & 52.76 / 81.71\\
 & static pool & 16.85 / 30.41 & 35.75 / 71.78 & 35.75 / 71.74 & 18.13 / 15.00 & 32.04 / 69.75 & 52.80 / 76.29 & 56.60 / 83.02\\
 & STEP & 17.12 / 31.84 & \textbf{41.30} / \textbf{74.88} & 41.25 / \textbf{74.82} & 15.95 / 14.02 & 40.44 / \textbf{74.64} & \textbf{54.31} / \textbf{77.87} & \textbf{57.30} / \textbf{83.59}\\
 & prevalence only$^\ddagger$ & -- & -- & \textbf{52.95} / 66.83 & -- & -- & -- & --\\
\addlinespace[2pt]
2023, stage 5 & ensemble & \textbf{19.04} / \textbf{37.16} & 50.58 / \textbf{78.24} & 50.56 / \textbf{78.23} & \textbf{31.61} / \textbf{21.57} & \textbf{55.63} / 75.20 & 54.94 / 77.20 & 57.04 / 84.23\\
 & static pool & 18.41 / 33.80 & \textbf{51.04} / 76.48 & 51.02 / 76.41 & 20.57 / 17.87 & 48.12 / 77.09 & 55.84 / 78.38 & 58.27 / 84.85\\
 & STEP & 18.59 / 34.64 & 49.08 / 77.88 & 49.08 / 77.85 & 21.38 / 17.88 & 49.22 / \textbf{77.82} & \textbf{56.65} / \textbf{79.23} & \textbf{58.76} / \textbf{85.21}\\
 & prevalence only$^\ddagger$ & -- & -- & \textbf{52.95} / 66.83 & -- & -- & -- & --\\
\addlinespace[2pt]
2023, stage 6 & ensemble & \textbf{23.15} / \textbf{45.71} & 52.33 / \textbf{80.90} & 52.33 / \textbf{80.93} & \textbf{36.58} / \textbf{23.46} & \textbf{55.91} / 75.66 & 54.11 / 79.45 & 57.84 / 86.02\\
 & static pool & 22.15 / 39.55 & 54.05 / 79.37 & 54.04 / 79.37 & 22.81 / 19.58 & 52.77 / 79.81 & 55.99 / 80.75 & 59.86 / 86.84\\
 & STEP & 22.33 / 40.38 & \textbf{54.18} / 80.25 & \textbf{54.15} / 80.22 & 20.89 / 18.78 & 53.65 / \textbf{80.35} & \textbf{56.58} / \textbf{81.67} & \textbf{60.36} / \textbf{87.18}\\
 & prevalence only$^\ddagger$ & -- & -- & 52.95 / 66.83 & -- & -- & -- & --\\
\bottomrule
\end{tabular}
\end{adjustbox}
\end{table}

\paragraph{Beyond top-$k$: precision, recall and F1.} Tables~\ref{tab:prf} and~\ref{tab:extprf} give, for every method, macro precision, recall and F1 of the top-1 prediction and under the paper's top-3 convention (predicted positive $=$ in the shortlist), on the balanced test split and on the natural 2024 cohort. On the balanced split the pool's gain is a recall and F1 gain at top-1 but not a precision gain under the top-3 convention: STEP's shortlists contain more spurious entries per protocol, so its top-3 macro F1 is below the ensemble's even though its top-3 accuracy is higher, and the bias-free pool has the best macro F1 under both conventions. On the natural cohort the label-free prior correction raises micro accuracy at the cost of macro recall, because it moves rare-protocol predictions to the majority protocol. Which variant to deploy therefore depends on the objective.

\begin{table}[t]
\centering\footnotesize\setlength{\tabcolsep}{3.2pt}
\caption{Macro-averaged precision (P), recall (R) and F1 over the 82 protocols on the 8,198 test records of the balanced cohort, in \%, with ``predicted positive'' $=$ ``appears in the top $k$'' (at $k{=}1$ this is ordinary multi-class macro P/R/F1; at $k{=}3$ every record contributes three predicted positives, so precision falls; micro recall equals top-3 accuracy, and macro recall approximates it because supports are nearly equal). Because the cohort is balanced (100 test records for 81 protocols and 98 for one), macro and support-weighted averages coincide to the first decimal. STEP and the stage MLP are the repaired-family, clean-boundary models of Table~\ref{tab:main}; the prompted arms ran on the same repaired family (Qwen3: mean over demonstration seeds 0--2; MedGemma: seed 0).}
\label{tab:prf}
\vspace{4pt}
\begin{tabular}{clrrrrrrrrrrrr}
\toprule
& & \multicolumn{3}{c}{STEP} & \multicolumn{3}{c}{stage MLP (seed 0)} & \multicolumn{3}{c}{Qwen3-32B, $k{=}3$} & \multicolumn{3}{c}{MedGemma-27B, $k{=}3$}\\
\cmidrule(lr){3-5} \cmidrule(lr){6-8} \cmidrule(lr){9-11} \cmidrule(lr){12-14}
$k$ & stage & P & R & F1 & P & R & F1 & P & R & F1 & P & R & F1\\
\midrule
\multirow{6}{*}{1} & 1 & \textbf{19.1} & \textbf{20.7} & 16.0 & 19.0 & 20.4 & \textbf{16.3} & 12.4 & 15.0 & 10.5 & 11.4 & 15.6 & 10.4\\
 & 2 & \textbf{23.2} & \textbf{24.6} & 20.9 & 22.7 & 24.5 & \textbf{21.3} & 12.5 & 15.9 & 11.1 & 12.0 & 16.2 & 10.9\\
 & 3 & \textbf{26.9} & \textbf{27.8} & \textbf{24.5} & 25.3 & 27.3 & 23.8 & 13.2 & 16.3 & 11.1 & 13.2 & 16.4 & 11.3\\
 & 4 & \textbf{47.7} & \textbf{48.2} & \textbf{46.6} & 45.2 & 45.8 & 44.7 & 32.3 & 25.7 & 21.0 & 29.2 & 22.7 & 18.0\\
 & 5 & \textbf{53.5} & \textbf{53.7} & \textbf{52.2} & 50.0 & 50.3 & 49.0 & 32.2 & 27.1 & 23.0 & 29.5 & 24.8 & 20.1\\
 & 6 & \textbf{58.7} & \textbf{59.3} & \textbf{58.0} & 56.1 & 56.0 & 55.2 & 36.2 & 33.3 & 28.9 & 35.9 & 28.8 & 24.4\\
\midrule
\multirow{6}{*}{3} & 1 & 15.2 & \textbf{38.3} & 19.7 & 15.2 & 38.1 & \textbf{19.8} & 13.1 & 22.8 & 12.3 & \textbf{15.6} & 24.9 & 13.8\\
 & 2 & \textbf{15.9} & \textbf{43.9} & \textbf{22.1} & 15.8 & 43.4 & \textbf{22.1} & 12.5 & 24.6 & 12.6 & 15.4 & 26.1 & 14.6\\
 & 3 & \textbf{17.0} & \textbf{48.4} & \textbf{24.3} & 16.8 & 47.8 & 24.2 & 12.8 & 26.3 & 12.9 & 16.6 & 26.3 & 14.8\\
 & 4 & 25.4 & \textbf{70.5} & \textbf{36.8} & 25.2 & 67.3 & 36.0 & 26.6 & 36.0 & 21.3 & \textbf{27.3} & 36.9 & 21.6\\
 & 5 & \textbf{27.3} & \textbf{75.6} & \textbf{39.6} & 26.6 & 72.1 & 38.2 & 26.9 & 36.7 & 23.1 & 26.5 & 40.1 & 23.3\\
 & 6 & 30.1 & \textbf{79.8} & \textbf{42.9} & 30.2 & 76.0 & 42.1 & 30.0 & 42.7 & 27.7 & \textbf{30.8} & 45.3 & 26.9\\
\bottomrule
\end{tabular}
\end{table}

\begin{table}[t]
\centering\scriptsize\setlength{\tabcolsep}{2.5pt}
\caption{External cohort (90,759 naturally distributed records), stages 4--6: micro accuracy, macro F1 (top-1 and top-3 conventions), and the same after a transductive label-free EM estimate of the prior from the unlabelled posteriors of the same records that are scored \citep{saerens2002adjusting}. ``No class bias'': the pool refitted with its per-class bias frozen at zero.}
\label{tab:extprf}
\vspace{4pt}
\begin{tabular}{llrrrrrrr}
\toprule
st. & method & micro top-1 & micro top-3 & macro F1$_1$ & macro F1$_3$ & EM top-1 & EM top-3 & EM macro F1$_1$\\
\midrule
4 & ensemble & 19.49 & 37.18 & 15.62 & \textbf{12.69} & 37.85 & 75.50 & 21.01\\
 & static pool & 18.22 & 32.50 & 15.28 & 12.25 & 46.80 & 82.27 & 18.91\\
 & STEP & 18.31 & 33.49 & 15.07 & 12.18 & 48.27 & 83.69 & 18.17\\
 & static pool, no class bias & 19.48 & 37.11 & 15.62 & 12.66 & 58.83 & 82.83 & 14.34\\
 & STEP, no class bias & \textbf{19.76} & \textbf{37.28} & \textbf{15.63} & 12.60 & \textbf{59.50} & \textbf{83.81} & 13.63\\
 & masked ens. & 18.87 & 35.98 & 15.27 & 12.56 & 43.50 & 79.95 & \textbf{21.49}\\
\addlinespace[1pt]
5 & ensemble & 20.92 & 39.40 & 17.30 & 13.79 & 53.06 & 84.84 & 24.55\\
 & static pool & 20.11 & 35.86 & 17.12 & 13.47 & 57.54 & 85.95 & 23.57\\
 & STEP & 20.23 & 36.52 & 17.09 & 13.48 & 54.24 & 86.17 & 23.90\\
 & static pool, no class bias & 21.63 & 40.86 & 17.66 & 13.96 & 60.78 & 85.39 & 18.01\\
 & STEP, no class bias & \textbf{21.65} & \textbf{40.97} & \textbf{17.76} & \textbf{14.00} & \textbf{62.06} & \textbf{86.23} & 17.85\\
 & masked ens. & 20.61 & 38.67 & 17.30 & 13.61 & 49.76 & 83.22 & \textbf{24.60}\\
\addlinespace[1pt]
6 & ensemble & 24.42 & 47.52 & 20.04 & 15.60 & 55.38 & 85.45 & \textbf{28.48}\\
 & static pool & 23.20 & 41.11 & 19.69 & 15.03 & 61.06 & 88.32 & 27.90\\
 & STEP & 23.24 & 41.92 & 19.58 & 15.01 & 60.41 & \textbf{88.41} & 26.98\\
 & static pool, no class bias & \textbf{25.46} & \textbf{48.06} & \textbf{20.60} & \textbf{15.76} & 61.69 & 85.26 & 21.44\\
 & STEP, no class bias & 25.22 & 47.65 & 20.38 & 15.68 & \textbf{62.78} & 87.71 & 17.99\\
 & masked ens. & 23.66 & 43.46 & 19.46 & 14.91 & 50.05 & 84.50 & 27.84\\
\addlinespace[1pt]
\bottomrule
\end{tabular}
\end{table}

\paragraph{When does pooling earlier posteriors help? A synthetic study.} Table~\ref{tab:synthetic} simulates two nested blocks: a 20-dimensional informative first block and a 300-dimensional second block of which a fraction $f_2$ is informative, revealed in that order to two per-stage MLPs trained and pooled under the same boundaries, over sample sizes 2,000 to 32,000, with and without redundancy of the earlier block, 50\% missingness of the new block and a cross-block interaction in which the second block is informative only jointly with the first. Two regimes appear. When the added block is uninformative or weak and the sample is small, the later stage loses 4 to 24 \% to the added dimensions (the peaking phenomenon), 5 to 29\% of records flip from correct to incorrect, and the pool recovers most of the loss (static $+4$ to $+11$ at $n=2{,}000$ and still $+4$ at $n=32{,}000$ when the block is pure noise; the gated pool up to $+24$), with the validation-fitted weight on the earlier stage rising as the sample grows and the new block stays noisy (0.27 at $n=2{,}000$ to 0.96 at $n=32{,}000$). When the added block is strongly informative the later stage is near-perfect, harmful flips vanish, validation returns weights close to the identity and the pool is inert ($\pm0.6$ \%), while at $n=2{,}000$ the fixed products lose 1 to 2 \%. Redundancy of the earlier block, missingness of the new one and a cross-block interaction change the size of the loss and the recovery but not the regime (Table~\ref{tab:synthetic}). The study is consistent with finite-sample variance control on the new block as the mechanism; recovery of accuracy alone does not separate variance from optimisation, regularisation or representation effects, which the regularised-joint control below addresses in part. It is not clinical evidence.

\begin{table}[t]
\centering\scriptsize\setlength{\tabcolsep}{3pt}
\caption{Synthetic progressive reveal (two blocks, 20 classes; block 2 has 300 dimensions of which a fraction $f_2$ is informative; mean over two generator seeds; test top-1 in \%). Harm and gain: records ranked correctly by stage 1 and not by stage 2, and the reverse. $\Delta$: pooled model minus the stage-2 three-seed ensemble under the same clean protocol (static and gated pools fitted on out-of-fold posteriors and selected on validation; PA the fixed product with $w=(1/2,1)$). Static $w$: the fitted stage weights. The full grid with redundancy and missingness is in the repository (\texttt{SYNTHETIC.md}).}
\label{tab:synthetic}
\vspace{4pt}
\begin{tabular}{rrcrrrrrrrr}
\toprule
$n$ & $f_2$ & interaction & acc.\ stage 1 & acc.\ stage 2 & harm & gain & static $\Delta$ & gated $\Delta$ & PA $\Delta$ & static $w$\\
\midrule
2,000 & 0.00 & no & 69.8 & 45.5 & 28.6 & 4.4 & $+11.12$ & {\bfseries\boldmath $+24.12$} & $+14.25$ & (0.27, 0.72)\\
2,000 & 0.05 & no & 69.8 & 65.0 & 17.4 & 12.6 & $+6.50$ & {\bfseries\boldmath $+12.12$} & $+9.00$ & (0.26, 0.91)\\
2,000 & 0.30 & no & 69.8 & 99.4 & 0.2 & 29.9 & {\bfseries\boldmath $-0.62$} & $-0.88$ & $-0.88$ & (0.21, 1.22)\\
2,000 & 0.30 & yes & 69.8 & 98.0 & 0.9 & 29.1 & $+0.62$ & {\bfseries\boldmath $+0.75$} & $+0.50$ & (0.21, 1.22)\\
8,000 & 0.00 & no & 70.8 & 61.3 & 14.9 & 5.4 & $+9.03$ & {\bfseries\boldmath $+9.31$} & $+4.44$ & (0.75, 0.20)\\
8,000 & 0.05 & no & 70.8 & 79.8 & 7.2 & 16.2 & {\bfseries\boldmath $+2.00$} & $+1.69$ & $+1.69$ & (0.47, 0.52)\\
8,000 & 0.30 & no & 70.8 & 99.5 & 0.2 & 28.8 & $-0.03$ & $+0.00$ & {\bfseries\boldmath $+0.03$} & (0.18, 1.24)\\
8,000 & 0.30 & yes & 70.8 & 99.6 & 0.3 & 29.0 & {\bfseries\boldmath $+0.06$} & {\bfseries\boldmath $+0.06$} & {\bfseries\boldmath $+0.06$} & (0.17, 1.14)\\
32,000 & 0.00 & no & 69.7 & 65.6 & 8.6 & 4.4 & $+4.23$ & {\bfseries\boldmath $+4.28$} & $+1.94$ & (0.96, 0.04)\\
32,000 & 0.05 & no & 69.7 & 82.0 & 5.0 & 17.3 & $+0.63$ & {\bfseries\boldmath $+0.71$} & {\bfseries\boldmath $+0.71$} & (0.25, 0.52)\\
32,000 & 0.30 & no & 69.7 & 99.7 & 0.1 & 30.1 & {\bfseries\boldmath $-0.02$} & {\bfseries\boldmath $-0.02$} & $-0.04$ & (0.16, 0.91)\\
32,000 & 0.30 & yes & 69.7 & 99.6 & 0.2 & 30.1 & {\bfseries\boldmath $+0.02$} & {\bfseries\boldmath $+0.02$} & {\bfseries\boldmath $+0.02$} & (0.09, 0.94)\\
\bottomrule
\end{tabular}
\end{table}

\paragraph{Does regularising the joint predictor remove the gain?} Table~\ref{tab:synreg} repeats the synthetic study with the stage-2 joint model tuned on the validation split over weight decay $\{0,10^{-4},10^{-3},10^{-2}\}$, dropout $\{0.2,0.5\}$ and early stopping on validation NLL, the selected configuration ensembled over three seeds and the static pool over the stage-1 and tuned stage-2 ensembles fitted and selected as before. Tuning raises the joint model by up to three \% at $n=2{,}000$ and by nothing at $n\ge8{,}000$, and the pool's gain over the tuned ensemble remains (pure-noise block: $\synRegTwoKNoise$, $\synRegEightKNoise$ and $\synRegThirtyTwoKNoise$ top-1 \% at $n=2{,}000$, $8{,}000$ and $32{,}000$ against $\synPlainTwoKNoise$, $\synPlainEightKNoise$ and $\synPlainThirtyTwoKNoise$ untuned); where the block is strongly informative the pool is inert with or without tuning. Regularising the joint predictor is therefore not a substitute for pooling in this generator, which narrows, without closing, the alternative that the increment is an artefact of an under-tuned joint model.

\begin{table}[t]
\centering\scriptsize\setlength{\tabcolsep}{4pt}
\caption{Does regularising the joint predictor remove the pooling gain? Same generator and protocol as Table~\ref{tab:synthetic} (two blocks, 20 classes; block 2 has 300 dimensions of which a fraction $f_2$ is informative). For the stage-2 joint model a grid over weight decay $\{0,10^{-4},10^{-3},10^{-2}\}$, dropout $\{0.2,0.5\}$ and early stopping on validation NLL is tuned on the validation split by top-1; the selected configuration (listed per repetition) is ensembled over three seeds and the static pool over the stage-1 and stage-2 ensembles is fitted and selected as before. Test top-1, mean over repetitions.}
\label{tab:synreg}
\vspace{4pt}
\begin{tabular}{rrrrrrrr}
\toprule
$n$ & $f_2$ & stage 1 & stage 2 plain & stage 2 tuned & tuned (wd / dropout) & pool $-$ plain ens. & pool $-$ tuned ens.\\
\midrule
2,000 & 0.0 & 70.2 & 43.4 & \textbf{46.5} & 0/0.5, 0.001/0.5 & $+14.37$ & $+11.38$\\
2,000 & 0.05 & 70.2 & 63.5 & \textbf{65.5} & 0.01/0.2, 0/0.5 & $+7.25$ & $+6.38$\\
2,000 & 0.3 & 70.2 & \textbf{99.4} & 99.2 & 0/0.2, 0/0.5 & $-0.75$ & $-0.50$\\
8,000 & 0.0 & 70.9 & 61.8 & \textbf{62.0} & 0.0001/0.5, 0.0001/0.5 & $+8.44$ & $+8.16$\\
8,000 & 0.05 & 70.9 & \textbf{80.3} & 80.1 & 0.01/0.5, 0.001/0.5 & $+1.88$ & $+1.69$\\
8,000 & 0.3 & 70.9 & \textbf{99.4} & \textbf{99.4} & 0/0.2, 0/0.5 & $+0.06$ & $+0.00$\\
32,000 & 0.0 & 69.9 & 66.3 & \textbf{66.4} & 0.0001/0.2, 0/0.2 & $+3.55$ & $+3.46$\\
32,000 & 0.05 & 69.9 & \textbf{82.9} & \textbf{82.9} & 0.001/0.2, 0.01/0.5 & $+0.44$ & $+0.13$\\
32,000 & 0.3 & 69.9 & 99.7 & \textbf{99.8} & 0.001/0.5, 0.001/0.2 & $-0.01$ & $+0.00$\\
\bottomrule
\end{tabular}
\end{table}

\begin{table}[t]
\centering\scriptsize\setlength{\tabcolsep}{3pt}
\caption{A second, unrelated staged-feature task: UCI Covertype (7 cover types) with an artificial, documented reveal order of its natural feature groups (topography; distances and hillshade; wilderness area; soil type), two subsample sizes, the paper's protocol (per-stage MLP $\times$ 3 seeds, block experts, out-of-fold pool fitting inside the fit rows, validation selection, one read of a 30 percent test split). $\Delta$: pool minus the 3-seed ensemble, exact McNemar $p$ (descriptive).}
\label{tab:covtype}
\vspace{4pt}
\begin{tabular}{rlrrrrr}
\toprule
$n$ & stage & single & ensemble top-1 / top-3 & static $\Delta_1$ ($p$) & gated $\Delta_1$ ($p$) & gated $\Delta_3$ ($p$)\\
\midrule
2,000 & 1 topography & \textbf{66.00} & 65.00 / 99.17 & {\bfseries\boldmath $+0.00$} (1) & {\bfseries\boldmath $+0.00$} (1) & $+0.00$ (1)\\
2,000 & 2 distances/hillshade & 68.50 & \textbf{69.17} / 99.67 & {\bfseries\boldmath $+0.00$} (1) & $-0.17$ (1) & $-0.17$ (1)\\
2,000 & 3 wilderness & 70.33 & \textbf{70.67} / 99.00 & {\bfseries\boldmath $+0.00$} (1) & $-0.67$ (0.39) & $+0.17$ (1)\\
2,000 & 4 soil & 68.33 & \textbf{69.00} / 99.17 & {\bfseries\boldmath $+0.00$} (1) & {\bfseries\boldmath $+0.00$} (1) & $+0.33$ (0.62)\\
\addlinespace[2pt]
5,000 & 1 topography & 66.73 & \textbf{67.13} / 99.07 & {\bfseries\boldmath $+0.13$} (0.69) & $+0.07$ (1) & $+0.00$ (1)\\
5,000 & 2 distances/hillshade & \textbf{72.47} & 72.00 / 99.53 & {\bfseries\boldmath $+0.13$} (0.69) & $+0.00$ (1) & $+0.07$ (1)\\
5,000 & 3 wilderness & \textbf{73.67} & 73.47 / 99.67 & $-0.13$ (0.8) & {\bfseries\boldmath $+0.13$} (0.69) & $+0.00$ (1)\\
5,000 & 4 soil & \textbf{74.67} & 74.13 / 99.40 & $+0.40$ (0.33) & {\bfseries\boldmath $+0.47$} (0.34) & $+0.00$ (1)\\
\addlinespace[2pt]
20,000 & 1 topography & \textbf{68.77} & 68.73 / 98.93 & $+0.00$ (1) & {\bfseries\boldmath $+0.07$} (0.6) & $+0.00$ (1)\\
20,000 & 2 distances/hillshade & 73.38 & \textbf{73.88} / 99.47 & $-0.27$ (0.18) & {\bfseries\boldmath $+0.07$} (0.78) & $+0.05$ (0.38)\\
20,000 & 3 wilderness & \textbf{76.62} & 76.55 / 99.62 & $+0.18$ (0.17) & {\bfseries\boldmath $+0.25$} (0.2) & $+0.07$ (0.22)\\
20,000 & 4 soil & 79.62 & \textbf{79.77} / 99.73 & {\bfseries\boldmath $+0.35$} (0.11) & $+0.25$ (0.26) & $+0.02$ (1)\\
\addlinespace[2pt]
2,000 (+500 noise cols) & 1 topography & \textbf{66.00} & 65.00 / 99.17 & {\bfseries\boldmath $+0.00$} (1) & {\bfseries\boldmath $+0.00$} (1) & $+0.00$ (1)\\
2,000 (+500 noise cols) & 2 distances/hillshade & 68.50 & \textbf{69.17} / 99.67 & {\bfseries\boldmath $+0.00$} (1) & $-0.17$ (1) & $-0.17$ (1)\\
2,000 (+500 noise cols) & 3 wilderness & 70.33 & \textbf{70.67} / 99.00 & {\bfseries\boldmath $+0.00$} (1) & $-0.67$ (0.39) & $+0.17$ (1)\\
2,000 (+500 noise cols) & 4 soil & 69.83 & \textbf{71.33} / 98.00 & $+1.83$ (0.14) & {\bfseries\boldmath $+2.83$} (0.068) & $+1.50$ (0.022)\\
\addlinespace[2pt]
5,000 (+500 noise cols) & 1 topography & 66.73 & \textbf{67.13} / 99.07 & {\bfseries\boldmath $+0.13$} (0.69) & $+0.07$ (1) & $+0.00$ (1)\\
5,000 (+500 noise cols) & 2 distances/hillshade & \textbf{72.47} & 72.00 / 99.53 & {\bfseries\boldmath $+0.13$} (0.69) & $+0.00$ (1) & $+0.07$ (1)\\
5,000 (+500 noise cols) & 3 wilderness & \textbf{73.67} & 73.47 / 99.67 & $-0.13$ (0.8) & {\bfseries\boldmath $+0.13$} (0.69) & $+0.00$ (1)\\
5,000 (+500 noise cols) & 4 soil & \textbf{76.33} & 75.87 / 98.33 & $+0.47$ (0.61) & {\bfseries\boldmath $+3.93$} (3.3e-06) & $+1.13$ (1.5e-05)\\
\addlinespace[2pt]
20,000 (+500 noise cols) & 1 topography & \textbf{68.77} & 68.73 / 98.93 & $+0.00$ (1) & {\bfseries\boldmath $+0.07$} (0.6) & $+0.00$ (1)\\
20,000 (+500 noise cols) & 2 distances/hillshade & 73.38 & \textbf{73.88} / 99.47 & $-0.27$ (0.18) & {\bfseries\boldmath $+0.07$} (0.78) & $+0.05$ (0.38)\\
20,000 (+500 noise cols) & 3 wilderness & \textbf{76.62} & 76.55 / 99.62 & $+0.18$ (0.17) & {\bfseries\boldmath $+0.25$} (0.2) & $+0.07$ (0.22)\\
20,000 (+500 noise cols) & 4 soil & 78.83 & \textbf{79.98} / 98.90 & {\bfseries\boldmath $+4.08$} (2.6e-25) & $+3.85$ (2e-22) & $+0.75$ (1.2e-08)\\
\addlinespace[2pt]
\bottomrule
\end{tabular}
\end{table}

\paragraph{A second, unrelated staged-feature task.} Table~\ref{tab:covtype} applies the protocol to UCI Covertype (581,012 records, 7 cover types, 54 features) with an artificial, documented reveal order of its natural feature groups: topography (elevation, aspect, slope), then distances to hydrology, roads and fire points with the three hillshades, then the four wilderness indicators, then the 40 soil-type indicators; subsamples of 2,000, 5,000 and 20,000 records (seed 0) are split 70/30 with a 15\% inner validation, continuous columns are standardised on the fit rows, and the rest follows the paper (caption of Table~\ref{tab:covtype}). On the unmodified task the pool is inert at every size (increments within $\pm\covUnmodMaxAbs$, all $p>\covUnmodMinP$; top-3 is above 98\% from stage 1, so the task leaves no room there): the blocks are low-dimensional and the joint model is not in the finite-sample regime. To test the prediction rather than the dataset, the last block was padded with 500 sparse binary columns whose activation rate (1--5\%) depends weakly on the class (a log-normal tilt with standard deviation 0.3), i.e.\ a high-dimensional, weakly informative block of the kind the synthetic study identifies. The pool then recovers more than the added columns cost: $\covPadOneFiveK$ / $\covPadThreeFiveK$ top-1 / top-3 at $n=5{,}000$ ($p=\covPadPOneFiveK$) and $\covPadOneTwentyK$ / $\covPadThreeTwentyK$ at $n=20{,}000$ ($p=\covPadPOneTwentyK$; static pool $\covPadStaticOneTwentyK$), and the harmful-transition rate at $3\to4$ falls from \covHarmEnsFiveK\% to \covHarmGatedFiveK\% at $n=5{,}000$. This is one dataset with one artificial reveal order and one synthetic padding, compared against three-seed ensembles only, so it shows the increment is not specific to NEMSIS and appears in the predicted regime; it does not show that natural staged blocks elsewhere sit in that regime.

\end{document}